\documentclass{article}

\usepackage[main,final]{neurips_2026}

\usepackage[utf8]{inputenc} 
\usepackage[T1]{fontenc}    
\usepackage{hyperref}       
\usepackage{url}            
\usepackage{booktabs}       
\usepackage{amsfonts}       
\usepackage{nicefrac}       
\usepackage{microtype}      
\usepackage{xcolor}         
\usepackage{amsmath}
\usepackage{amsthm}
\usepackage[textsize=tiny]{todonotes}
\usepackage{graphicx}
\usepackage{subcaption}  
\usepackage{tikz}
\usetikzlibrary{positioning}
\usepackage{algorithm}
\usepackage{algorithmic}
\usepackage{wrapfig}

\usepackage[table]{xcolor}

\usepackage[normalem]{ulem}

\theoremstyle{plain}
\newtheorem{theorem}{Theorem}[section]
\newtheorem{proposition}[theorem]{Proposition}
\newtheorem{lemma}[theorem]{Lemma}
\newtheorem{assumption}[theorem]{Assumption}

\usepackage{titletoc}

\newcommand{\red}[1]{{\color{red} #1}}

\title{Behavior-Discriminative Reward Shaping for Reward-Robust Reinforcement Learning}

\author{%
  Zixuan Liu \\
  Department of Computer Science \\
  Tulane University \\
  New Orleans, LA 70118, USA \\
  \texttt{zliu41@tulane.edu} \\
  \And
  Fangzheng Wu \\
  Department of Computer Science \\
  Tulane University \\
  New Orleans, LA 70118, USA \\
  \texttt{fwu6@tulane.edu} \\
  \And
  Brian Summa \\
  Department of Computer Science \\
  Tulane University \\
  New Orleans, LA 70118, USA \\
  \texttt{bsumma@tulane.edu} \\
  \And
  Zizhan Zheng \\
  Department of Computer Science \\
  Tulane University \\
  New Orleans, LA 70118, USA \\
  \texttt{zzheng3@tulane.edu} \\
}

\begin{document}

\maketitle

\begin{abstract}
  Reward-robust RL typically models misspecification by specifying an uncertainty set that constrains the discrepancy of plausible rewards from a base reward. However, such reward space discrepancies can be behaviorally irrelevant: under potential-based reward shaping (PBRS), many distinct rewards preserve policy ordering. This can introduce substantial redundancy into standard uncertainty sets and degrade optimization performance. We propose \textbf{Shaping-Aware Reward-Robust RL}, which constructs uncertainty sets over PBRS equivalence classes by projecting each reward to a canonical representative, ensuring that the resulting set contains only rewards that induce behaviorally distinct policy rankings. We prove that this projection preserves the optimal robust value while shrinking the uncertainty set and improving empirical performance. Using the connection between robustness and regularization, we obtain a practical algorithm to solve the shaping-aware reward-robust RL problem and enjoy convergence guarantees under standard assumptions. Experiments on benchmarks spanning diverse task domains and levels of complexity show consistent improvements over representative robust RL baselines and exhibit improved robustness to reward perturbations. The code is available at \url{https://github.com/ZixuanLiu4869/reward_shaping}.
\end{abstract}

\section{Introduction}
Despite the impressive success of reinforcement learning (RL) across a wide range of domains~\cite{kiran2021deep,christiano2017deep,kober2013reinforcement,liu2024enhancing,liu2026risk,liu2026ua,liu2026targeting}, deploying RL in real-world systems remains challenging. A central bottleneck is reward misspecification: the reward function used for training is often an imperfect proxy for the desired objective due to limited data, human biases, modeling errors, or distribution shift~\cite{mannor2007bias,jeon2020reward,enders2024risk}. Therefore, in safety and reliability critical settings, we would like policies that continue to perform well under plausible reward perturbations. This motivates reward-robust RL, which models misspecification by assuming all plausible rewards lie within a given uncertainty set and solves a max-min objective that maximizes worst-case performance. Equivalently, it can also be viewed as a dynamic zero-sum game: an adversary selects the worst-case reward within the uncertainty set, while the agent seeks a policy that maximizes the resulting worst-case return~\cite{morimoto2005robust,lim2013reinforcement,pinto2017robust,anonymous2025robust}. 

Existing reward-robust RL frameworks typically construct uncertainty sets by measuring the expected divergence of plausible rewards from a base reward~\cite{gadot2024solving,eysenbach2021maximum}. However, a key weakness that has been largely overlooked in the robust RL literature is that controlling such divergences in reward space does not necessarily correspond to differences in decision-making. In particular, potential-based reward shaping (PBRS), which encompasses a general and widely adopted family of shaping transformations~\cite{ng1999policy,muller2025improving,lidayan2024bamdp}, shows that certain shaping rewards leave the optimal policy invariant and, more generally, preserve policy ordering. As a consequence, a divergence-based uncertainty set can contain many behaviorally equivalent rewards: they differ under the chosen divergence measure but induce identical policy rankings. This redundancy can be harmful in robust optimization. In particular, it can waste adversary iterations on rewards that provide no new behavioral distinctions rather than finding genuinely more adverse rewards (as theoretically justified and empirically shown in Section~\ref{sec:motivation} and Figure~\ref{fig:grid-world}), ultimately weakening the adversary and degrading robust training.

In this work, we make the first step towards behavior-discriminative robust RL by proposing \textbf{Shaping-Aware Reward-Robust RL}. 
Our framework addresses the above issues by constructing a shaping-aware projected uncertainty set in which each PBRS-equivalent reward class is represented by a canonical reward.
Intuitively, this removes rewards that provably do not affect policy ranking and yields an uncertainty set that better reflects behaviorally distinct objectives. We show that this reduction does not change the optimal robust value of the original reward-robust RL problem, while substantially simplifying the adversary's search space. Empirically, we find that optimization on the projected uncertainty set makes adversary updates more identifiable and improves both performance and convergence.

However, 
directly solving the max-min robust RL problem with projected rewards requires projecting every candidate reward considered by the adversary to its canonical representative, which is computationally expensive. To address this, we first show that optimizing a suitably regularized objective with a projected base reward provides a conservative lower bound for the shaping-aware robust objective. This gives a tractable surrogate that only requires projecting the base reward, thereby avoiding repeated projections during the inner max-min optimization. We further show that, for certain uncertainty sets, such as entropy- and KL-induced uncertainty sets, this lower bound becomes exact in our projected setting. 
We further choose the canonical representative by minimizing the discrepancy induced by the uncertainty set within each PBRS equivalence class. This projection selects a representative that is consistent across equivalence classes and conservative with respect to the robust objective. Putting these together, we derive a practical algorithm that alternates between learning the canonical form of the base reward and optimizing the corresponding regularized RL objective. Since our approach only introduces a lightweight projection-learning step prior to standard policy updates, it can be integrated with a wide range of existing RL optimizers. Moreover, under standard assumptions, we establish convergence guarantees for the overall algorithm. 

We empirically evaluate our method on benchmarks spanning diverse task domains and levels of complexity, comparing against representative robust RL baselines. Across all environments, our method consistently achieves higher returns and reaches strong performance in fewer training steps. We further assess robustness under reward perturbations and observe that our method exhibits smaller performance degradation than standard baselines when rewards are shifted. Overall, these results support the view that removing PBRS-induced redundancy leads to more effective robust policy learning under reward uncertainty. 

To summarize, the key contribution of this work is: 1) To the best of our knowledge, we are the first to incorporate PBRS equivalence into the reward-robust RL and to construct uncertainty sets based on behavioral distinctions, rather than raw reward discrepancies, yielding stronger robust learning performance. 2) We derive a practical shaping-aware reward-robust RL algorithm that trains policies against this behaviorally distinct uncertainty set, and we provide convergence guarantees under standard assumptions. 3) We empirically validate the method across diverse environments, showing higher returns, faster convergence, and improved robustness to reward perturbations compared to representative baselines.
\vspace{-2ex}
\section{Preliminary}
\vspace{-2ex}
\label{sec:preliminary}
\textbf{Notation.} 
For a finite set \(\mathcal{Z}\), we write \(\mathbb{R}^{\mathcal{Z}}\) for the set of real-valued functions on \(\mathcal{Z}\), and \(\Delta_{\mathcal{Z}}\) for the probability simplex over \(\mathcal{Z}\). More generally, given any set \(\mathcal{X}\), \(\Delta_{\mathcal{Z}}^{\mathcal{X}}\)
denotes the collection of mappings from \(\mathcal{X}\) into
\(\Delta_{\mathcal{Z}}\). For \(a,b \in \mathbb{R}^{\mathcal{Z}}\), we define the standard inner product as \(\langle a,b\rangle := \sum_{z\in\mathcal{Z}} a(z)\,b(z)\).
Let \(C \subset \mathbb{R}^{\mathcal{Z}}\) be a convex set and let \(\Omega : C \to \mathbb{R}\) be strongly convex. Throughout the paper, \(\Omega\) serves as a policy regularizer. We define \(
\Omega^{*}(y)
= \max_{a\in C} \bigl\{\langle a,y\rangle - \Omega(a)\bigr\}, \forall\, y \in \mathbb{R}^{\mathcal{Z}} 
\), which corresponds to the Legendre-Fenchel transform (also known as the convex conjugate) on \(C\). 

\textbf{Standard RL.}
Consider an infinite-horizon Markov decision process (MDP) \((\mathcal{S},\mathcal{A},\mu_0,\gamma,P,r)\), where \(\mathcal{S}\) and \(\mathcal{A}\) are finite state and action spaces, \(0 < \mu_0 \in \Delta_{\mathcal{S}}\) is the initial state distribution and
\(\gamma \in (0,1)\) is the discount factor. Define the state action space \(\mathcal{X} := \mathcal{S}\times\mathcal{A}\) and the transition space \(\mathcal{X'} := \mathcal{S}\times\mathcal{A}\times\mathcal{S}\). The dynamics are given by a transition kernel \(P \in \Delta_{\mathcal{S}}^{\mathcal{X}}\), which assigns to each \((s,a)\in \mathcal{X}\) a distribution over next states in \(\mathcal{S}\). The reward is a function
\(r \in \mathbb{R}^{\mathcal{X'}}\) defined over \((s,a,s')\in \mathcal{X'}\), where \(s'\) generated from \(P(\cdot|s,a)\) is the state reached after executing action \(a\) in state \(s\). A policy \(\pi \in \Delta_{\mathcal{A}}^{\mathcal{S}}\) specifies an action distribution for each state \(s \in \mathcal{S}\). Given \(\mu_0,\pi\) and \(P\), the discounted return under reward \(r\) is: \(
J(\pi,r)
:= \mathbb{E}\!\left[\sum_{t=0}^{\infty} \gamma^{t} r(s_t,a_t,s_{t+1})
    \,\Big|\, \mu_0,\pi,P \right].
\) The standard RL objective is to find a policy \(\pi\) that maximizes the discounted return.

\textbf{Potential-Based Reward Shaping (PBRS).} 
PBRS~\cite{ng1999policy} augments the environment reward via a potential function \(\Phi\in \mathbb{R}^{\mathcal{S}}\), which assigns a scalar to each state. Given \(\Phi\), PBRS defines a shaping function \(F\) on transition tuples \((s,a,s')\in\mathcal{X'}\) by \(
F(s,a,s') = \gamma \Phi(s') - \Phi(s),
\) and the shaping reward \(r'\in \mathbb{R}^{\mathcal{X'}}\) as \(
r'(s,a,s') = r(s,a,s') + F(s,a,s').
\) A key property of PBRS is policy invariance: replacing \(r\) with \(r'\) 
preserves the return ordering among all policies (Theorem~1 of~\cite{ng1999policy}).

\textbf{Reward-Robust RL.}
A reward-robust MDP is given by
\((\mathcal{S},\mathcal{A},\mu_0,\gamma,P,\mathcal{U})\), where the true reward is unknown but assumed to lie in a prescribed uncertainty set \(\mathcal{U}\subseteq \mathbb{R}^{\mathcal{X'}}\).
Typically, \(\mathcal{U}\) is defined around a base reward \(r\in\mathbb{R}^{\mathcal{X'}}\) through a discrepancy budget: \(
\mathcal{U}=\{\tilde r\in \mathbb{R}^{\mathcal{X'}} : \Omega^*(r-\tilde r)\le \epsilon\},
\)
where \(\Omega^*\) is a dissimilarity measure and \(\epsilon>0\) controls the robustness level. The reward-robust RL objective seeks a policy \(\pi \) that maximizes worst-case return:
\begin{equation}
\label{eq:robust_rl_objective}
\max_{\pi \in \Delta_{\mathcal{A}}^{\mathcal{S}}}\,\min_{\tilde r\in\mathcal{U}} J(\pi,\tilde r).
\end{equation}
The uncertainty sets studied in the literature typically arise from specific choices of \(\Omega^*\). For example, define the reward difference \(u(s,a,s')=r(s,a,s')-\tilde r(s,a,s')\), the simplest and common used is $l_p$-constrained uncertainty set, where \(\Omega^*(u)=\|u\|_{p}\)~\cite{kumar2023policy,ho2021partial,kumar2025dual}. In addition,~\cite{eysenbach2021maximum} consider an \(\Omega^*\) of the form \(
\Omega^*(u)= \mathbb{E}\left[
            \sum_{t} \gamma^{t}
            \log \sum_{a }
\exp(u(s_t,a,s_{t+1}))\Big| \mu_0, \pi, P \right]\), which induces a log-sum-exp discrepancy over reward differences along trajectories. In addition,~\cite{derman2023robustness} define a reference policy weighted variant: \(
\Omega^*(u) = \mathbb{E}\left[
            \sum_{t} \gamma^{t}
            \log \sum_{a\in\mathcal{A}}\pi_{\mathrm{ref}}(a|s_t)
           \exp(u(s_t, a,s_{t+1}))\Big| \mu_0, \pi, P\right] 
\), where \(\pi_{\mathrm{ref}}\) is a reference policy. Notice that most \(\Omega^*\) considered in prior work can be derived from a regularizer \(\Omega\) (detailed below). Our work mainly focuses on the uncertainty set induced by such \(\Omega^*\).

\textbf{Regularized RL.}
A regularized MDP is a tuple
\((\mathcal{S},\mathcal{A},\mu_0,\gamma,P,r,\Omega)\),
where \(\Omega\) denotes an additional regularization term. The regularized RL objective takes the form
\begin{equation}
\label{eq:regularized_rl}
\max_{\pi \in \Delta_{\mathcal{A}}^{\mathcal{S}}} \; J(\pi,r)+\Omega(\pi),
\end{equation}
A widely used choice is policy-entropy regularization~\cite{haarnoja2018soft,eysenbach2021maximum}: \(
\Omega_{\mathrm{Ent}}(\pi)= - \mathbb{E}\!\left[\sum_{t}\gamma^t
\sum_{a}\pi(a\mid s_t)\log\pi(a\mid s_t)
\Big| \mu_0,\pi,P\right].
\)
Another common regularizer is the KL divergence to a reference policy~\cite{derman2023robustness,filippi2010optimism}: \(
\Omega_{\mathrm{KL}}(\pi)= - \mathbb{E}\left[\sum_{t}\gamma^t
\sum_{a}\pi(a\mid s_t)\log
\frac{\pi(a\mid s_t)}{\pi_{\mathrm{ref}}(a\mid s_t)}
\Big|\mu_0,\pi,P\right].
\) Prior work has established a connection between regularization and robustness in RL for specific choices of $\Omega$ and $\Omega^*$~\cite{eysenbach2021maximum,derman2023robustness,husain2021regularized}. In particular, entropy-regularized and KL-regularized RL objectives can be interpreted as reward-robust RL problems, with corresponding discrepancies given by the log-sum-exp penalty~\cite{eysenbach2021maximum} and a reference policy weighted variant~\cite{derman2023robustness}. For simplicity, in the remainder of the paper, we refer to these reward-robust formulations by the name of their corresponding regularizer (e.g., “entropy” or “KL”) whenever the meaning is clear. A detailed related work on PBRS, reward-robust RL, and regularized RL is in Appendix~\ref{appendix:related-work}.

\section{Shaping-Aware Reward-Robust RL}

Solving the inner minimization in the reward-robust RL objective~\eqref{eq:robust_rl_objective} requires an adversary to search for the worst-case reward within the uncertainty set. However, when the uncertainty set is defined directly in reward space, it can contain many PBRS-equivalent rewards that induce identical policy behavior, leading to redundant adversary search. In Section~\ref{sec:motivation}, we illustrate this issue on a toy MDP and discuss its impact on optimization behavior. In Section~\ref{sec:project}, we introduce \textbf{Shaping-Aware Reward-Robust RL}, which removes this redundancy by projecting the uncertainty set onto PBRS equivalence classes so that each element corresponds to a behaviorally distinct objective. We show that this projection preserves the optimal robust value while yielding a smaller and more identifiable uncertainty set.

\subsection{Why PBRS Redundancy Harm Optimization?}
\label{sec:motivation}

\begin{figure}[ht]
    \centering

    \begin{subfigure}[b]{0.22\textwidth}
        \centering
        \begin{tikzpicture}[>=stealth, node distance=3cm, scale=0.8]
            \node[draw,circle,minimum size=18pt] (s1) {$s_1$};
            \node[draw,circle,minimum size=18pt, right=2.0cm of s1] (s2) {$s_2$};

            \draw[->] (s1) edge[bend left=20] node[above]{$a_1$} (s2);
            \draw[->] (s1) edge[bend right=20] node[below]{$a_2$} (s2);
        \end{tikzpicture}
        \caption{}
        \label{fig:toy-mdp}
    \end{subfigure}
    \hfill
    \begin{subfigure}[b]{0.75\textwidth}
        \centering
        \includegraphics[width=\linewidth]{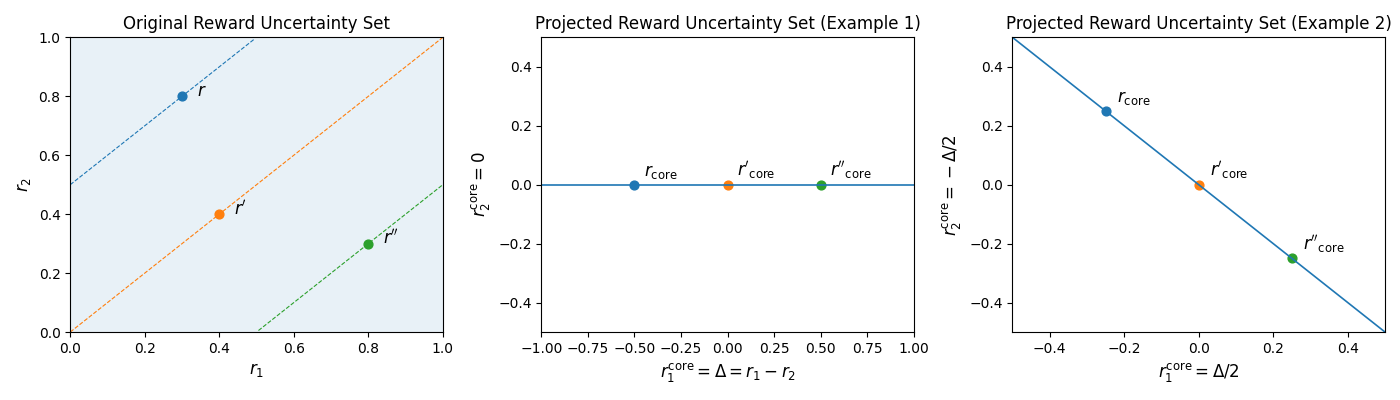}
        \caption{}
        \label{fig:dimension-shrink}
    \end{subfigure}

    \caption{(a) A two-state toy MDP. The agent starts in \(s_1\) and can choose between two actions, \(a_1\) and \(a_2\). Regardless of the action taken in \(s_1\), the transition to \(s_2\) is deterministic. (b) The original reward uncertainty set is the square $[0,1]^2$ over the $(r_1, r_2)$ plane. All points on the same dashed diagonal induce identical policy rankings under PBRS. Projecting each PBRS equivalence class to a canonical representative collapses the uncertainty set to a one-dimensional set. The left figure highlights example rewards $r,r',r''$. The middle and right figure show their corresponding representatives $r_{\text{core}},r'_{\text{core}},r''_{\text{core}}$ under two different projection choices.}
    \label{fig:toy-mdp-dim}
    \vspace{-4ex}
\end{figure}

In this section, we illustrate PBRS-induced redundancy and its optimization implications via the toy MDP in Figure~\ref{fig:toy-mdp}. This MDP has two states \(s_1,s_2\) and two actions \(a_1,a_2\). From \(s_1\), either action deterministically transitions to \(s_2\). Let \(r_1 := r(s_1,a_1,s_2)\), \(r_2 := r(s_1,a_2,s_2)\), and consider the rectangular uncertainty set in Figure~\ref{fig:dimension-shrink} (left), where \(
\mathcal{U}:=\{(r_1,r_2): r_1\in[0,1],\, r_2\in[0,1]\}.
\) We first characterize the PBRS-equivalent rewards within \(\mathcal{U}\). As discussed in Section~\ref{sec:preliminary}, a potential function takes the form \(F(s,a,s')=\gamma\Phi(s')-\Phi(s)\). In this toy MDP, both actions share the same transition \(s_1\!\to\! s_2\), hence the shaping term applied to each action is identical, i.e., \(
F_1 = F_2 = \gamma\Phi(s_2)-\Phi(s_1)=:F.
\)
Consequently, adding the same shift to both rewards, \(
(\tilde r_1,\tilde r_2)=(r_1,r_2)+F\cdot(1,1),
\)
does not change policy ranking. The only behaviorally relevant quantity is the reward difference \(\Delta := r_1-r_2\). Geometrically, PBRS equivalence classes correspond to the slope-\(1\) diagonals in the \((r_1,r_2)\)-plane. As illustrated by the dashed lines in Figure~\ref{fig:dimension-shrink} (left), all rewards lying on the same dashed diagonal are behaviorally equivalent. As a result, \(\mathcal{U}\) contains entire families of distinct rewards that are PBRS equivalent, introducing substantial redundancy. To remove the redundancy, a natural way is to select a canonical representative $r^{\text{core}}$ from each equivalence class, thereby projecting the original reward space onto a lower-dimensional, behaviorally distinct space. Many canonicalization rules are available. For instance, we may fix \(r_2^{\text{core}}=0\) and set \(r_1^{\text{core}}=\Delta\), yielding a one-dimensional set \(\{(\Delta,0):\Delta\in[-1,1]\}\) (Figure~\ref{fig:dimension-shrink} (middle)). Alternatively, we may center rewards by enforcing zero mean, e.g.,
\(r_1^{\text{core}}=\tfrac{\Delta}{2}\) and \(r_2^{\text{core}}=-\tfrac{\Delta}{2}\), which corresponds to the diagonal line through the origin (Figure~\ref{fig:dimension-shrink} (right)). 

The PBRS-induced redundancy can lead to inefficient and sometimes misleading worst-case optimization. Consider the adversary's update over \(\tilde r \in \mathcal{U}\). If the adversary optimizes directly in the original reward space, its iterates can move along PBRS-invariant directions. Though in theory, the adversary would not systematically move in these directions, since the expected gradient along PBRS-invariant directions is zero. In practice, with finite rollouts and approximate return estimates, gradient estimates can induce non-zero noise even along behaviorally irrelevant directions. This noise can cause a random walk along PBRS-invariant directions, wasting adversary updates and potentially leading to premature stabilization at PBRS-equivalent rewards that provide little genuine progress in lowering the worst-case value. By contrast, optimizing over the projected reward space, where PBRS-invariant directions are removed, ensures that each adversary update is identifiable in the sense that it can meaningfully affect the worst-case objective. We further show this in Section~\ref{sec:experiment} (MiniGrid experiment) and Appendix~\ref{appendix:non-convex} in non-convex uncertainty sets.

\subsection{Robust Objective Invariance Under Shaping}
\label{sec:project}

In this section, we make the redundancy phenomenon precise and introduce our shaping-aware reward-robust RL framework to address it. We first define behaviorally equivalent rewards under PBRS: rewards that differ only by a PBRS term shift the return of every policy by the same constant, and therefore induce identical behavior in the sense of decision-making (i.e., leave policy ranking unchanged)~\cite{ng1999policy}. 

\begin{lemma}
\label{lemma:shaping}
Let \(\Phi\in\mathbb{R}^{\mathcal{S}}\) be any bounded potential function and let \(F(s,a,s')=\gamma\Phi(s')-\Phi(s)\) be the corresponding shaping term. Define the shaped reward \(r'(s,a,s')=r(s,a,s')+F(s,a,s')\). Then, for any policy \(\pi\), \(
J(\pi,r') = J(\pi,r) + C(\Phi,\mu_0),
\) where \(C(\Phi,\mu_0)=-\mathbb{E}_{s_0\sim\mu_0}[\Phi(s_0)]\) depends only on \(\Phi\) and the initial distribution \(\mu_0\), and is independent of \(\pi\). We say \(r'\) is PBRS-equivalent to \(r\).
\end{lemma}
Lemma~\ref{lemma:shaping} is a well-known property of PBRS, and we include the proof in Appendix~\ref{appendix:lemma-pbrs} for completeness. Notice that the additive term \(C(\Phi,\mu_0)\) can always be eliminated by introducing a common start state $s_0$ and setting $\Phi(s_0)=0$. Thus, without loss of generality, we adopt the convention \(C=0\) throughout the paper. In practice, we enforce this condition by adding an extra loss term detailed in Appendix~\ref{appendix:implementation-mujuco}. Next, we show that the standard uncertainty sets considered in the literature include rewards that are behaviorally indistinguishable under PBRS.
\begin{proposition}
\label{prop:affine_redundancy}
Let \(\mathcal{U}\) be an uncertainty set as defined in Section~\ref{sec:preliminary}, of the form \( \mathcal{U}=\{\tilde r: \Omega^*(r-\tilde r) \le \epsilon\} \). Assume \(\mathcal{U}\) contains a neighborhood of some \(\tilde r\) (e.g., \(\mathcal{U}\) has nonempty interior around \(\tilde r\)).
Then there exists a nontrivial shaping direction \(F\neq 0\) (generated by some \(\Phi\in\mathbb{R}^{\mathcal{S}}\)) and a scalar \(\delta>0\) such that \(
\tilde r + \alpha F \in \mathcal{U},\text{ for all }|\alpha|\le \delta.
\)
All rewards in \(\{\tilde r+\alpha F:|\alpha|\le\delta\}\) are PBRS-equivalent and therefore induce identical policy rankings.
\end{proposition}

The proof is provided in Appendix~\ref{appendix:propo-redudant}. Proposition~\ref{prop:affine_redundancy} highlights that, because \(\mathcal{U}\) is defined over raw rewards, it can contain entire neighborhoods of PBRS-equivalent rewards. This redundancy motivates constructing an uncertainty set over PBRS equivalence classes, so that each element represents a unique, behaviorally distinct objective. To formalize this idea, let \(\mathcal{R}\subseteq \mathbb{R}^{\mathcal{X'}}\) denote the reward space of interest. Define the shaping subspace \(
\mathcal{S}_{\text{shape}} := \{F_{\Phi} : \Phi \in \mathbb{R}^{\mathcal{S}}\},
\) and the corresponding quotient space of PBRS equivalence classes \(
\mathcal{R}_{\text{core}} := \mathcal{R} / \mathcal{S}_{\text{shape}},
\)
whose elements are classes
\([r]=\{r+F_\Phi: F_\Phi \in \mathcal{S}_{\text{shape}}\}\). Assume we fix a canonical representative for each equivalence class via a projection map
\(\text{Proj} :\mathcal{R} \to \mathcal{R} \) such that for any \(r\in\mathcal{R}\),
\(\text{Proj}(r)\in[r]\) and \(\text{Proj}(r+F_\Phi)=\text{Proj}(r)\) for all \(F_\Phi\in\mathcal{S}_{\text{shape}}\). Using the above definition, given any uncertainty set \(\mathcal{U}\subseteq\mathcal{R}\), we then define its projected uncertainty set as \(
\mathcal{U}_{\text{core}} := \text{Proj}(\mathcal{U})
= \{\,\text{Proj}(r) : r \in \mathcal{U}\,\}\subseteq \mathcal{R},
\) which collapses PBRS-equivalent rewards in \(\mathcal{U}\) to a single representative defined by \(\text{Proj}\). We now establish that working with \(\mathcal{U}_{\text{core}}\) is not only conceptually cleaner, but also preserves the robust objective. In particular, we have:

\begin{theorem}
\label{theorem:core-eq}
Let \(\mathcal{U}_{\text{core}}\) be the projected uncertainty set of \(\mathcal{U}\) as defined above, we have
\begin{equation}
\label{eq:core_robust_rl}
 \min_{\tilde r \in \mathcal{U}} J(\pi, \tilde r)
\;=\;
\min_{\tilde r_{\text{core}} \in \mathcal{U}_{\text{core}}}
J(\pi, \tilde r_{\text{core}}), \qquad \forall \pi.
\end{equation}
Consequently, the standard reward-robust RL problem can be solved over the shaping-aware projected uncertainty set \(\mathcal{U}_{\text{core}}\) without changing the robust optimal value. 
\end{theorem}

The proof of Theorem~\ref{theorem:core-eq} is provided in Appendix~\ref{appendix:theorem-core-eq}. It follows directly from the definition of $\mathcal{U}_{\text{core}}$ together with Lemma~\ref{lemma:shaping}. Theorem~\ref{theorem:core-eq} implies that one can directly solve the robust objective over \(\mathcal U_{\mathrm{core}}\) instead of \(\mathcal U\). We instantiate this direct approach in the MiniGrid experiment in Section~\ref{sec:experiment}, where the projected uncertainty set is explicit and can be optimized over efficiently. We quantify how projection reduces the adversary's effective search space in Appendix~\ref{appendix:dimension-drop}, where the extent of the reduction scales with the number of independent shaping directions present in \(\mathcal{U}\). Additionally, Appendix~\ref{appendix:property-proj} shows that under mild assumptions, $\mathcal{U}_{\text{core}}$ inherits key geometric properties of $\mathcal{U}$, such as compactness and convexity. These properties ensure the optimization on $\mathcal{U}_{\text{core}}$ is well-posed.

\section{Shaping-Aware Reward-Robust Algorithm}

A direct optimization of~\eqref{eq:core_robust_rl} requires solving the inner minimization over \(\mathcal{U}_{\mathrm{core}}\). In practice, this means maintaining feasibility after each adversary update and thus projecting intermediate rewards back onto \(\mathcal{U}_{\mathrm{core}}\) at every inner step. However, computing this projection is itself a nontrivial optimization, adding substantial per-step overhead. This overhead can be prohibitive in large-scale RL, where the inner minimization is already the main computational bottleneck. While alternative implementations may reduce this cost, designing an efficient algorithm is nontrivial. In Section~\ref{sec:shaping-aware-proj}, we address the issue by leveraging the connection between robustness and regularization. And then we specify a concrete canonicalization rule. We present the overall algorithm in Section~\ref{sec:policy-gradient-algorithm}.

\subsection{Robustness-Regularization Duality Under Shaping}
\label{sec:shaping-aware-proj}

We first show that the regularized objective is a conservative surrogate to~\eqref{eq:core_robust_rl}. 

\begin{theorem}
\label{theorem:core-regular-eq-robust}
Fix \(\epsilon\ge 0\) and \(\lambda>0\). Let \(\Omega\) be a proper, closed, and strongly convex regularization term, and let \(\Omega^*\) denote its convex conjugate. Define the uncertainty set \(
\mathcal{U}
:=\Bigl\{\tilde r:
\Omega^*\left(\frac{r-\tilde r}{\lambda}\right)\le \epsilon
\Bigr\}
\) and the projected uncertainty set \(\mathcal{U}_{\mathrm{core}}=\text{Proj}(\mathcal U)\). Let $r_{\mathrm{core}}:=\text{Proj}(r)$ be the projection of base reward $r$. Then the shaping-aware reward-robust objective satisfies 
\[
\min_{\tilde r_{\mathrm{core}}\in\mathcal U_{\mathrm{core}}}
J(\pi,\tilde r_{\mathrm{core}})
\ge
J(\pi,r_{\mathrm{core}})-\lambda\Omega(\pi)
-\lambda\epsilon, \qquad \forall \pi .
\]
\end{theorem}
The proof is deferred to Appendix~\ref{appendix:core-regular-eq-robust}, where we first prove the inequality for the original uncertainty set and then show that the same inequality carries over to the projected uncertainty set using Lemma~\ref{lemma:shaping} and Theorem~\ref{theorem:core-eq}, as projection preserves the robust objective. Theorem~\ref{theorem:core-regular-eq-robust} shows that the regularized objective with the projected base reward \(r_{\mathrm{core}}\) provides a conservative lower bound for~\eqref{eq:core_robust_rl} and maximizing this lower bound gives a tractable surrogate. Instead of repeatedly solving the inner minimization over projected rewards, it suffices to compute \(r_{\mathrm{core}}\) and optimize the corresponding regularized objective, which substantially reduces computational overhead. 
We next show that, for the entropy- and KL-induced uncertainty sets, this lower-bound relation becomes exact.

\begin{theorem}
\label{theorem:entropy-kl-exact-duality}
Fix \(\epsilon\ge 0\) and \(\lambda>0\). Define the uncertainty set \(\mathcal U=\left\{
\tilde r: \Omega^*\!\left(\frac{r-\tilde r}{\lambda}\right)
\le \frac{\epsilon}{\lambda}
\right\}\) and the projected uncertainty set \(\mathcal U_{\mathrm{core}}=\mathrm{Proj}(\mathcal U)\).
Let \(r_{\mathrm{core}}:=\mathrm{Proj}(r)\). For the entropy-based and KL-based choices of \(\Omega^*\) introduced in
Section~\ref{sec:preliminary}, the following exact correspondence holds:
\[
\min_{\tilde r_{\mathrm{core}}\in\mathcal U_{\mathrm{core}}}
J(\pi,\tilde r_{\mathrm{core}})
=J(\pi,r_{\mathrm{core}})
+
\lambda\Omega(\pi)
-
\epsilon,\qquad \forall \pi,
\]
where \(\Omega=\Omega_{\mathrm{Ent}}\) for the entropy-based and \(\Omega=\Omega_{\mathrm{KL}}\) for the KL-based uncertainty set. 
\end{theorem}
The proof is deferred to Appendix~\ref{appendix:entropy-kl-duality-proof}. We first establish the exact correspondence for the original uncertainty sets, following prior work~\cite{ashlag2025state}, and then extend it to the projected uncertainty set similar to Theorem~\ref{theorem:core-regular-eq-robust}. Theorem~\ref{theorem:entropy-kl-exact-duality} shows that, for these uncertainty sets, maximizing the regularized objective with \(r_{\mathrm{core}}\) is equivalent to optimizing the corresponding shaping-aware robust objective.

It remains to specify the projection operator \(\mathrm{Proj}\), i.e., how we choose a canonical representative from each PBRS equivalence class. In principle, many canonicalizations are possible. Recall that our robust objective measures reward uncertainty using the dissimilarity induced by \(\Omega^*\). It is therefore natural to define the projection in a way that is consistent with this geometry. We adopt a conservative canonicalization: within each equivalence class \([r]\), we select the representative that is smallest under the same \(\Omega^*\)-induced geometry: \(
\operatorname{Proj}(r)=\arg\min_{r'\in[r]} \Omega^*(r').
\) Intuitively, for each PBRS-equivalence class, our method selects the representative with the smallest $\Omega^*$ value, yielding a consistent canonical reward that is aligned with the discrepancy geometry induced by the uncertainty set. Moreover, this choice avoids inadvertently interpreting PBRS-equivalent rewards outside the original budgeted set as admissible, which would correspond to certifying robustness with respect to a larger uncertainty set than the one we specify. 
Additional properties of this projection are detailed in Appendix~\ref{appendix:property-projection-omega}. In particular, we show that the projection is well-defined by establishing existence and uniqueness. We also show that, from a max-min viewpoint, this projection can tighten convergence bounds for certain optimization methods.

To implement \(\operatorname{Proj}(r)\), notice that \(r'\in[r]\) means that \(r\) differs from \(r'\) by a potential-based shaping term, i.e.,
\(r = r'+F_{\Phi}\) with \(F_{\Phi}(s,a,s') = \gamma \Phi(s') - \Phi(s)\). In practice, we approximate \(\Phi\) using a neural network \(\Phi_{\varphi}\) with parameters \(\varphi\) as a flexible and general function approximation. Then the projection reduces to optimizing over \(\varphi\): \(
\varphi^{*}
= \arg\min_{\varphi}
\Omega^*\Bigl(
  r - F_{\Phi_{\varphi}}
\Bigr).
\)
Given the learned parameters \(\varphi^{*}\), we compute the shaping term for each reward as \(
F_{\Phi_{\varphi^*}},
\) and obtain the corresponding projected reward by subtracting this shaping component, \(
r' = r - F_{\Phi_{\varphi^*}}.
\) 

\subsection{Shaping-Aware Reward-Robust Policy Gradient}
\label{sec:policy-gradient-algorithm}

We summarize the overall training procedure to solve our shaping-aware reward-robust RL problem in Algorithm~\ref{alg:projected-ppo}. At each iteration, we first learn the projected reward by estimating the projection loss \(l(\varphi_{i,k}, \theta_i; \mathcal{D}_{\pi_{\theta_i}})\), where \(\Omega^*_{\mathcal{D}_{\pi_{\theta_i}}}\) denotes the estimate of \(\Omega^*\) computed from the rollout data \(\mathcal{D}_{\pi_{\theta_i}}\), and updating \(\varphi\) via gradient descent (lines 6-8). We then replace the rewards in the same batch with the resulting projected reward and perform a standard policy update on the regularized objective (lines 10-11). We justify in Appendix~\ref{appendix:algorithm} that such
on-policy projection-learning design is data-efficient and supports online adaptation. Although Algorithm~\ref{alg:projected-ppo} updates the policy via policy gradients, in principle, our projection simply replaces the environment reward with its canonicalized version, so it is compatible with any RL algorithm that optimizes expected return (e.g., SAC~\cite{haarnoja2018soft} or PPO~\cite{schulman2017proximal}).  In Appendix~\ref{appendix:algorithm}, we prove convergence under standard conditions by first analyzing the projection-learning updates and then applying a two-time scale argument~\cite{bertsekas2009convex,gadot2024solving} to establish convergence of the overall algorithm. Notice that our convergence guarantee relies on a strong convexity assumption for the projection loss (Assumption~\ref{ass:strong-convexity}), which does not directly apply to the neural-network setting. We discuss this further in Appendix~\ref{appendix:limitations}. Algorithm~\ref{alg:projected-ppo} learns the projector approximately from a finite rollout batch, so the learned projection may deviate from the exact projector. We show that such projection error does not affect the structure of Theorem~\ref{theorem:core-regular-eq-robust} and~\ref{theorem:entropy-kl-exact-duality} (Appendix~\ref{appendix:algorithm}, Theorem~\ref{theorem:approx-lower-bound} and~\ref{theorem:approx-entropy-kl-duality}). It only introduces an approximation error that scales with \(
\mathcal O\left(
\frac{1}{1-\gamma}
\left(
\frac{\log N+\log(1/\delta)}{N}
\right)^{1/4}
\right)
\), which vanishes as the number of rollout samples \(N\) increases. Overall, our algorithm should be interpreted as an approximate implementation inspired by theory.

\begin{algorithm}[ht]
\caption{Shaping-Aware Policy Gradient}
\label{alg:projected-ppo}
\small
\begin{algorithmic}[1]
\STATE \textbf{Hyperparameters:} Iteration number $I$, $K$, Step size $\{\beta_i\}_{i=0}^{I-1}$ for $\theta$, Step size $\{\alpha_{i,k}\}_{i=0,\dots,I-1;\,k=0,\dots,K-1}$ for $\varphi$.
\STATE \textbf{Initialize: } $\theta_0 \in \Theta, \varphi_0 \in \Psi$
\FOR{iteration $i = 0, \dots, I-1$}
    \STATE Collect a batch dataset $\mathcal{D}_{\pi_{\theta_i}}$ from the current policy $\pi_{\theta_i}$.

    \STATE Initialize $\varphi_{i,0}=\varphi_i$
    \FOR{iteration $k = 0, \dots, K$}
    
    \STATE Update $\varphi$:
    \(
    \varphi_{i,k+1}=\varphi_{i,k}-\alpha_{i,k}\nabla_\varphi l(\varphi_{i,k},\theta_i;\mathcal{D}_{\pi_{\theta_i}})
    \)
    where \(l(\varphi_{i,k},\theta_i;\mathcal{D}_{\pi_{\theta_i}})= \Omega^*_{\mathcal{D}_{\pi_{\theta_i}}}\bigl(r - F_{\Phi_{\varphi_{i,k}}}\bigr)
    \)
    \ENDFOR
    \STATE Set $\varphi_{i+1}:=\varphi_{i,K}$
    \STATE Transform $\mathcal{D}_{\pi_{\theta_i}}$ to $\mathcal{D}'_{\pi_{\theta_i}}$ by replacing the rewards with  \(
        r'
        = r-F_{\Phi_{\varphi_{i+1}}}.
    \)
    \STATE Update $\theta$ via \(
    \theta_{i+1}=\theta_i+\beta_i \nabla_\theta\bar J(\theta_i,\varphi_{i+1};\mathcal{D}'_{\pi_{\theta_i}}) 
    \)
    where \(
    \bar J(\theta_i,\varphi_{i+1};\mathcal{D}'_{\pi_{\theta_i}}) = J_{\mathcal{D}'_{\pi_{\theta_i}}}(\pi_{\theta_i},r_{\varphi_{i+1}})-\Omega_{\mathcal{D}'_{\pi_{\theta_i}}}(\pi_{\theta_i})
    \)
\ENDFOR
\end{algorithmic}
\end{algorithm}

\vspace{-2ex}
\section{Experiment}
\label{sec:experiment}

We evaluate the performance and robustness of our shaping-aware reward-robust RL framework across benchmarks that span different task domains, 
and levels of complexity. Our goal is to test whether our method improves (i) nominal performance during robust training and (ii) robustness to reward misspecification, relative to standard reward-robust RL baselines. 

\textbf{Experiment setup.} We use a MiniGrid environment~\cite{chevalier2018minimalistic} as a controlled discrete benchmark for directly studying the max-min robust objective. In this setting, we adopt an \(l_2\)-constrained uncertainty set and compare policies trained by directly optimizing the standard max-min objective versus the shaping-aware max-min objective. For all remaining tasks, we focus on both entropy- and KL-based objectives discussed in Section~\ref{sec:preliminary}. In each case, we compare a standard PPO baseline against our approach, which uses the same PPO but augments training with the proposed projection step. We evaluate continuous-control performance on five OpenAI Gym MuJoCo environments~\cite{todorov2012mujoco,brockman2016openai}: Hopper-v4, Walker2d-v4, HalfCheetah-v4, Humanoid-v4, and Reacher-v4. These environments cover a meaningful range of state-action dimensions, where Hopper and Reacher represent smaller-scale settings, Walker2d and HalfCheetah represent mid-scale locomotion tasks, and Humanoid is substantially higher-dimensional. To evaluate generalization beyond continuous-control domains, we extend the evaluation to Atari~\cite{bellemare2013arcade}. Specifically, we consider Seaquest and Pong, which have discrete action spaces and high-dimensional visual observations of size \(210\times160\times3\). These experiments test whether the benefits of our method persist in visual decision-making settings. Finally, we include Adroit Pen from Gymnasium-Robotics~\cite{rajeswaran2017learning} as a more challenging simulated robotics benchmark. This task has a high-dimensional action space (23), allowing us to assess whether our method remains effective in higher-dimensional robotic manipulation. Additional implementation details are provided in Appendix~\ref{appendix:implementation}. Additional results, including learning curves showing faster convergence of our method, computational overhead, an ablation study on the sensitivity to data coverage for the projection learning, are presented in Appendix~\ref{appendix:experiment-result}.

\begin{figure}[ht]
    \centering
    \begin{subfigure}[b]{0.2\textwidth}
        \centering
        \includegraphics[width=\linewidth]{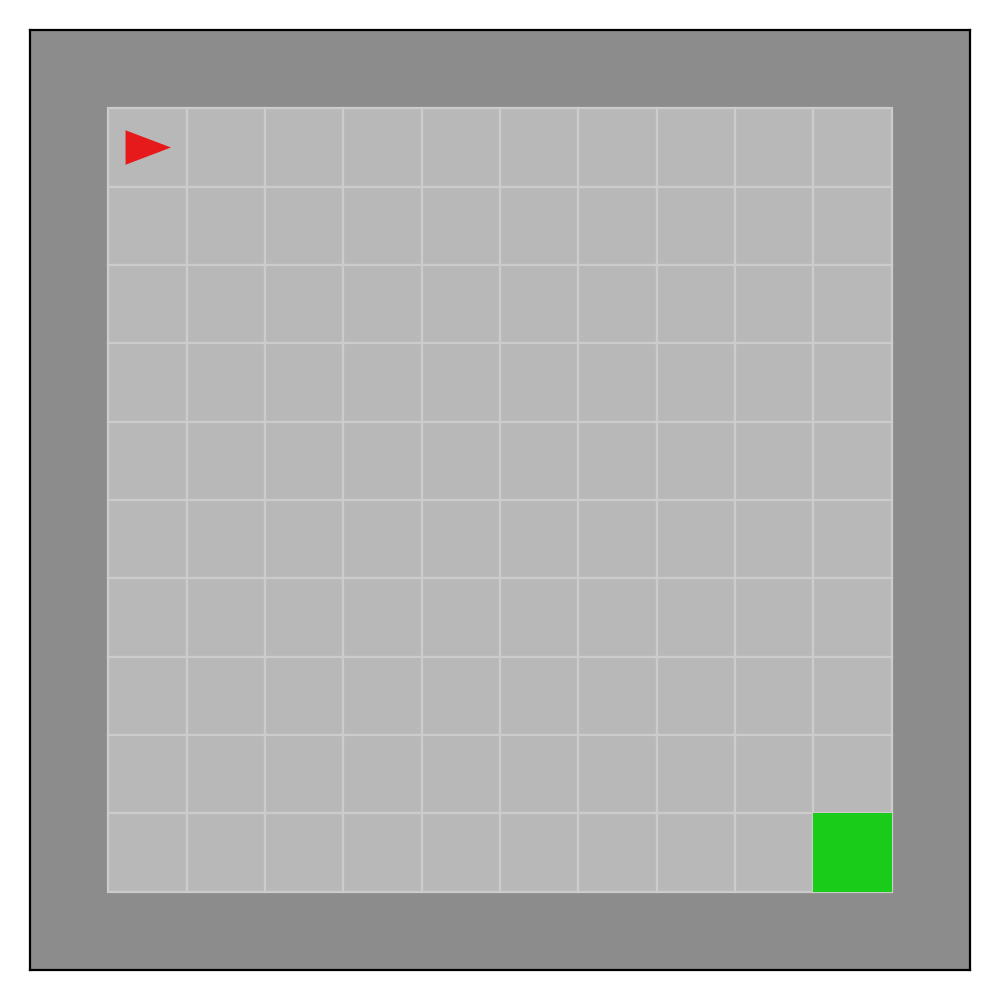}
        \caption{}
    \end{subfigure}
    \begin{subfigure}[b]{0.35\textwidth}
        \centering
        \includegraphics[width=\linewidth]{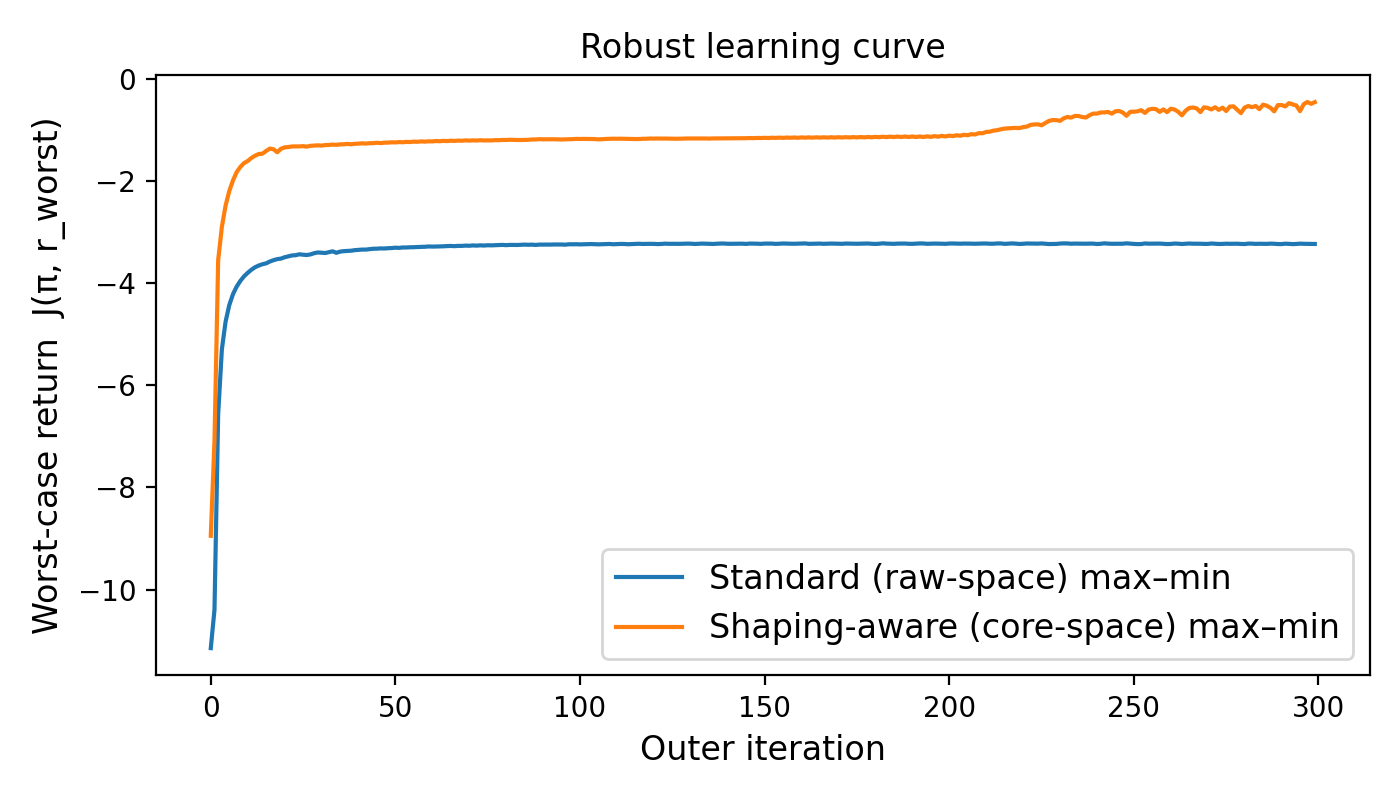}
        \caption{}
        \label{fig:worst-case-return}
    \end{subfigure}
    \begin{subfigure}[b]{0.35\textwidth}
        \centering
        \includegraphics[width=\linewidth]{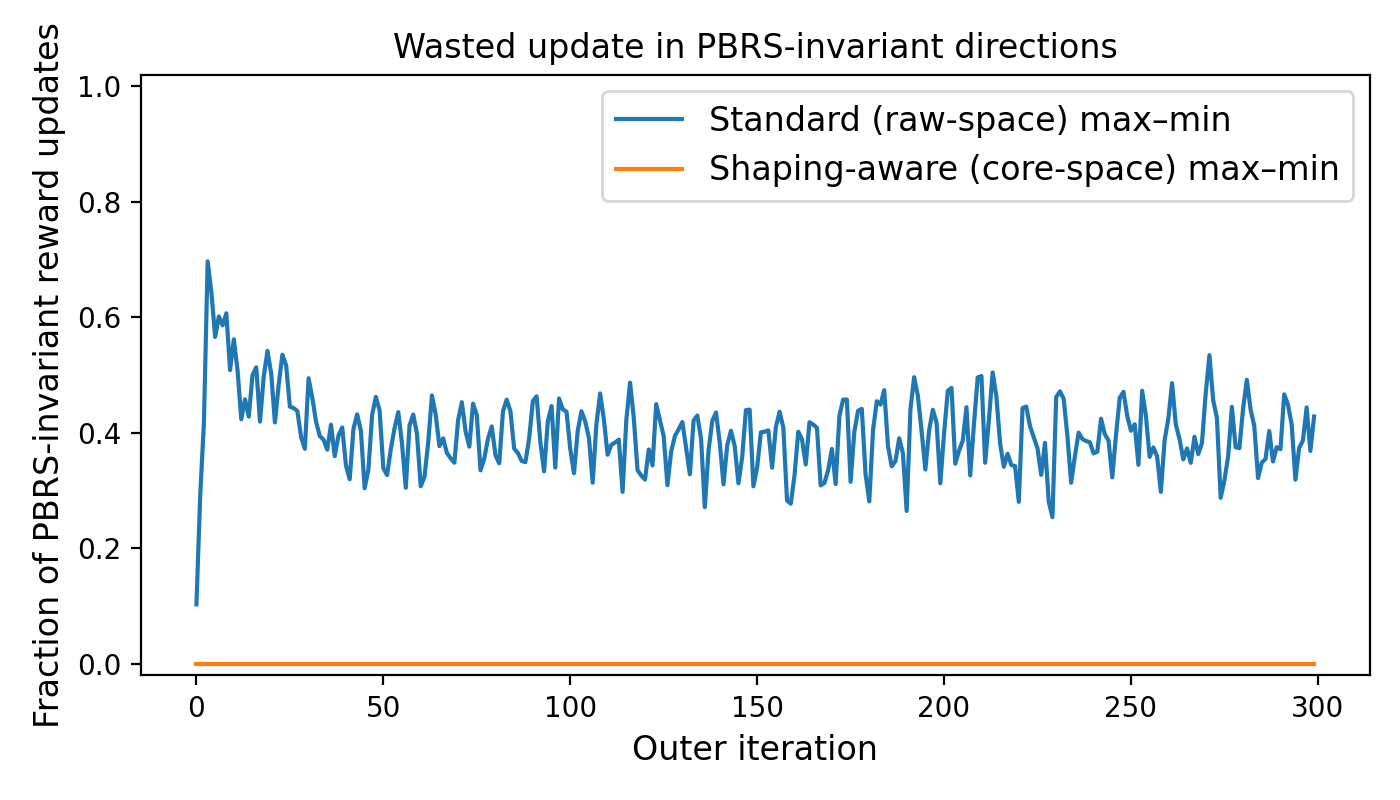}
        \caption{}
        \label{fig:waste}
    
    \end{subfigure}
    \caption{MiniGrid evaluation comparing the standard max-min baseline and our shaping-aware method. (a) Grid-world task: the agent starts in the top-left cell and must reach the goal in the bottom-right cell. (b) Worst-case return as a function of training iteration. (c) Fraction of each adversary update that lies in PBRS-invariant directions.}
    \label{fig:grid-world}
    \vspace{-4ex}
\end{figure}

\begin{table}[ht]
    \centering
    \caption{Average cumulative returns and percentage improvement for different methods on various environments.
    Each policy is evaluated for 100 episodes. The return is in terms of the environment's original reward. We report the mean $\pm$ 95\% confidence interval
    over 5 random seeds. $\Delta_{\text{entropy}}$ and $\Delta_{\text{KL}}$ denote the improvement of our method over the corresponding base method.}
    \label{tab:nominal_result}
    \resizebox{\textwidth}{!}{%
    \begin{tabular}{lcccc|cccc}
        \toprule
        Environment    & Robust RL (entropy)                    & \cellcolor[rgb]{ .863,  .902,  .941}Shaping-aware (entropy)            & \cellcolor[rgb]{ .863,  .902,  .941}$\Delta_{\text{entropy}}$ & &Robust RL (KL)                     & \cellcolor[rgb]{ .863,  .902,  .941}Shaping-aware (KL) &      \cellcolor[rgb]{ .863,  .902,  .941}$\Delta_{\text{KL}}$       \\
        \midrule
        Hopper-v4      & $2905.12\pm133.86$      & \cellcolor[rgb]{ .863,  .902,  .941}$3214.68\pm115.60$        & \cellcolor[rgb]{ .863,  .902,  .941}\red{$+309.56$} & $\uparrow 10.7\%$
                       & $2882.08\pm186.12$      & \cellcolor[rgb]{ .863,  .902,  .941}$3249.23\pm131.87$        & \cellcolor[rgb]{ .863,  .902,  .941}\red{$+367.15$} & $\uparrow 12.7\%$ \\
        Walker2d-v4    & $2861.81\pm316.32$      & \cellcolor[rgb]{ .863,  .902,  .941}$3385.88\pm281.79$        & \cellcolor[rgb]{ .863,  .902,  .941}\red{$+524.07$} & $\uparrow 18.3\%$
                       & $2676.94\pm263.54$      & \cellcolor[rgb]{ .863,  .902,  .941}$3577.35\pm352.55$        & \cellcolor[rgb]{ .863,  .902,  .941}\red{$+900.41$} & $\uparrow 33.6\%$ \\
        HalfCheetah-v4 & $1386.73\pm21.84$       & \cellcolor[rgb]{ .863,  .902,  .941}$1625.94\pm20.58$         & \cellcolor[rgb]{ .863,  .902,  .941}\red{$+239.21$} & $\uparrow 17.2\%$
                       & $1583.30\pm35.38$       & \cellcolor[rgb]{ .863,  .902,  .941}$3288.30\pm60.48$         & \cellcolor[rgb]{ .863,  .902,  .941}\red{$+1705.00$} & $\uparrow 107.7\%$ \\
        Humanoid-v4    & $463.15\pm25.82$        & \cellcolor[rgb]{ .863,  .902,  .941}$505.54\pm15.78$          & \cellcolor[rgb]{ .863,  .902,  .941}\red{$+42.39$} & $\uparrow 9.2\%$
                       & $372.60\pm28.52$        & \cellcolor[rgb]{ .863,  .902,  .941}$418.10\pm13.90$          & \cellcolor[rgb]{ .863,  .902,  .941}\red{$+45.50$} & $\uparrow 12.2\%$\\
        Reacher-v4     & $-5.98\pm0.27$          & \cellcolor[rgb]{ .863,  .902,  .941}$-4.88\pm0.49$            & \cellcolor[rgb]{ .863,  .902,  .941}\red{$+1.10$} & $\uparrow 18.4\%$
                       & $-5.88\pm0.44$          & \cellcolor[rgb]{ .863,  .902,  .941}$-5.17\pm0.32$            & \cellcolor[rgb]{ .863,  .902,  .941}\red{$+0.71$} & $\uparrow 12.1\%$ \\
         Seaquest       & $604.00\pm68.59$        & \cellcolor[rgb]{ .863,  .902,  .941}$832.00\pm48.99$          & \cellcolor[rgb]{ .863,  .902,  .941}\red{$+228.00$} & $\uparrow 37.7\%$
                       & $616.00\pm39.19$        & \cellcolor[rgb]{ .863,  .902,  .941}$840.00\pm29.39$          & \cellcolor[rgb]{ .863,  .902,  .941}\red{$+224.00$} & $\uparrow 36.4\%$ \\
        Pong           & $14.38\pm2.15$          & \cellcolor[rgb]{ .863,  .902,  .941}$16.13\pm1.36$            & \cellcolor[rgb]{ .863,  .902,  .941}\red{$+1.75$} & $\uparrow 12.2\%$
                       & $15.24\pm1.35$          & \cellcolor[rgb]{ .863,  .902,  .941}$17.88\pm1.74$            & \cellcolor[rgb]{ .863,  .902,  .941}\red{$+2.64$} & $\uparrow 17.3\%$ \\
        Adroit Pen     & $4239.94\pm114.38$      & \cellcolor[rgb]{ .863,  .902,  .941}$5085.26\pm112.20$        & \cellcolor[rgb]{ .863,  .902,  .941}\red{$+845.32$} & $\uparrow 19.9\%$
                       & $4434.22\pm68.89$       & \cellcolor[rgb]{ .863,  .902,  .941}$6087.30\pm63.68$         & \cellcolor[rgb]{ .863,  .902,  .941}\red{$+1653.08$} & $\uparrow 37.3\%$ \\
        \midrule
        Average        & $1558.64\pm85.40$       & \cellcolor[rgb]{ .863,  .902,  .941}$1832.57\pm74.60$         & \cellcolor[rgb]{ .863,  .902,  .941}\red{$+273.93$} & $\uparrow 17.6\%$
                       & $1571.81\pm77.93$       & \cellcolor[rgb]{ .863,  .902,  .941}$2184.12\pm81.74$         & \cellcolor[rgb]{ .863,  .902,  .941}\red{$+612.31$} & $\uparrow 39.0\%$ \\
        \bottomrule
    \end{tabular}}
    \vspace{-4ex}
\end{table}

\begin{figure}[ht]
    \centering
    \includegraphics[width=\linewidth]{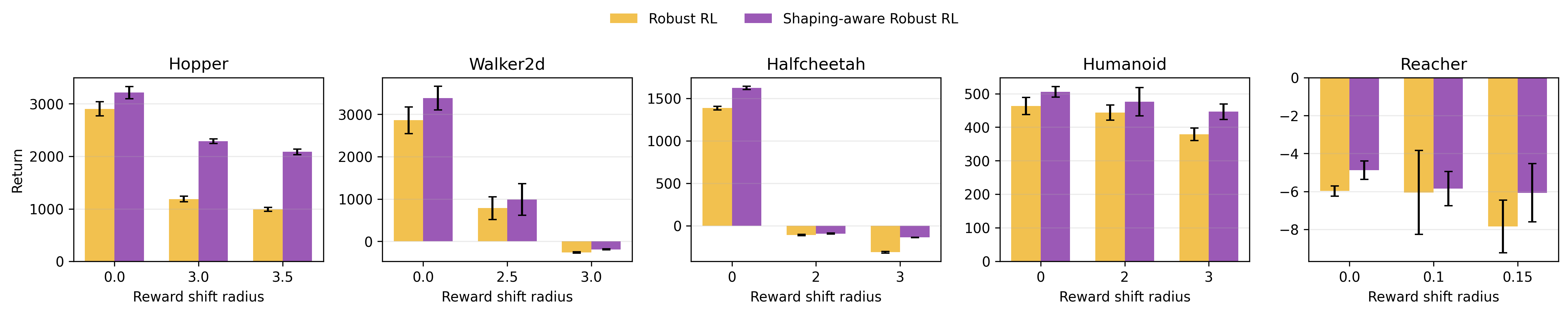}
    \caption{Mean $\pm$ 95\% confidence interval return of the evaluation policy under reward perturbation over 100 episodes and 5 seeds for the entropy-based problem. }
    \vspace{-2ex}
\end{figure}

\textbf{MiniGrid max-min robust training.} The evaluation result in the MiniGrid environment can be seen in Figure~\ref{fig:grid-world}. Specifically, Figure~\ref{fig:worst-case-return} reports the worst-case return \(J(\pi, r_{\text{worst}})\) across outer iterations for the standard max-min baseline and our shaping-aware method. Both methods improve rapidly at the beginning, but the shaping-aware variant continues to make steady progress and converges to a substantially higher worst-case return. This indicates that optimizing over the PBRS-canonicalized reward space leads to a stronger robust solution under the same uncertainty budget and comparable optimization settings. Figure~\ref{fig:waste} quantifies, at each outer iteration, what fraction of the adversary's update lies in PBRS-invariant directions. For the standard baseline, a persistent and 
significant fraction of adversary updates moves along shaping directions that do not alter policy orderings. This arises because the inner minimization is solved only approximately (via limited gradient descent), and the stochastic estimation noise (soft-max policies) together with inner-loop hyperparameter choices (stepsize) can induce drift along PBRS-invariant directions. Such updates are effectively wasted from the perspective of identifying more adverse behaviors and lead to premature stabilization at some PBRS-equivalent rewards that provide little genuine progress in lowering the worst-case value. In contrast, our method operates directly in the projected reward space, where PBRS-invariant directions are removed. As a result, the PBRS-invariant component of the adversary update is near zero, and each update is identifiable in the sense that it can meaningfully affect the worst-case objective. Together, these results support our motivating claim: removing PBRS redundancy makes adversary updates behaviorally identifiable, which translates into more effective robust training and better final worst-case performance.

\textbf{Comparison on MuJoCo, Atari, and Gymnasium-Robotics tasks.} To evaluate how well each method performs under the nominal environments, we report the results in Table~\ref{tab:nominal_result}. Across all cases, our shaping-aware variant consistently outperforms the corresponding robust RL baseline. In the MuJoCo environment, the gains are especially pronounced in mid-scale environments, e.g., 107.7\% on HalfCheetah in the KL setting. For the relatively smaller and larger environments, the improvements are moderate but still consistent. Notably, even on the highest-complexity Humanoid task, our method still achieves a 12.2\% improvement in the KL setting, suggesting that the benefit of our method persists beyond low-dimensional problems. The additional results from Atari and Gymnasium-Robotics further support this trend. On the visual Atari tasks, our method improves both Seaquest and Pong, while on the more challenging Adroit Pen manipulation task, it achieves substantial gains of 37.3\% in the KL setting, indicating that the benefits extend to high-dimensional visual and robotic manipulation domains. Overall, these results show that our method consistently improves robust policy learning and yields higher worst-case returns, thereby supporting the framework's broader applicability across different dimensions and task domains.

\begin{wraptable}{r}{0.55\textwidth}
\centering
\vspace{-10pt}
\caption{Effect of the shaping coefficient $\alpha$ on policy performance.}
\label{tab:shaping_coefficient}
\small
\begin{tabular}{@{}c l c@{}}
\toprule
$\alpha$ & Reward Representation & Return \\
\midrule
$0$    & $r_{\mathrm{core}}$ 
       & $-188.14 \pm 9.26$ \\
$0.25$ & $r_{\mathrm{core}} + 0.25F_{\Phi^\star}$ 
       & $-197.40 \pm 15.33$ \\
$0.5$  & $r_{\mathrm{core}} + 0.5F_{\Phi^\star}$ 
       & $-206.74 \pm 12.21$ \\
$0.75$ & $r_{\mathrm{core}} + 0.75F_{\Phi^\star}$ 
       & $-251.62 \pm 19.00$ \\
$1$    & $r_{\mathrm{core}} + F_{\Phi^\star} = r$ 
       & $-259.59 \pm 16.07$ \\
\bottomrule
\end{tabular}
\vspace{-10pt}
\end{wraptable}

\textbf{Testing for reward robustness.} To assess robustness to reward uncertainty, we evaluate all methods under structured reward perturbations that change the task incentives. We focus the evaluation on MuJoCo as its reward components are easily perturbable, and such structured reward-shift protocols have been widely used in prior work~\cite{husain2021regularized,ashlag2025state}. For the locomotion tasks (HalfCheetah, Hopper, Walker2d, and Humanoid), the nominal reward primarily encourages forward progress, and moving backward typically yields a non-positive reward. Following the protocol of~\cite{husain2021regularized}, during training, we flip this incentive by adding random reward shifts, so that backward motion can also be beneficial under the perturbed rewards. We report performance under the original reward during evaluation. For Reacher, where the objective is to reach a target position, we adopt the evaluation protocol of~\cite{ashlag2025state}. All policies are trained on the nominal environment. At test time, we relocate the target to a random position at varying distances from the nominal location and evaluate performance with respect to the reward induced by this relocated target. 
Implementation details for constructing the perturbed rewards are provided in Appendix~\ref{appendix:reward-perturb}. Overall, these perturbations introduce systematic and interpretable reward misspecification across domains, allowing us to compare how well each method maintains performance when the reward deviates from the nominal specification.

Figure~\ref{fig:reward_shift} reports robustness under reward perturbations for the entropy-based problem, and we include the results for KL-based problem in Appendix~\ref{appendix:reward-robust}. Overall, the shaping-aware method exhibits smaller performance degradation than the robust RL baseline as the reward shift radius increases, indicating improved stability to reward misspecification. The perturbations affect HalfCheetah and Walker2d most strongly, where returns drop sharply for both methods as the shift radius grows, suggesting these locomotion tasks are particularly sensitive to reward changes. In contrast, Hopper, Humanoid, and Reacher show more moderate changes with increasing shift radius, and the shaping-aware method consistently maintains higher returns than the baseline across perturbation levels. 

To provide a more direct evaluation of whether the observed improvement specifically arises from removing PBRS redundancy rather than from generic reward shaping, we define the actual PBRS component learned by our method as $ F_{\Phi^\star}=r- r_{\mathrm{core}}$. Then we construct $r_{\alpha}=r_{\mathrm{core}}+\alpha F_{\Phi^\star} $, and let $\alpha=0,0.25,0.5, 0.75, 1$, so that $\alpha=0$ corresponds to our canonical reward, $\alpha=1$ recovers the original reward, and the intermediate values progressively add back exactly the PBRS component removed by our projection. We evaluate these variants on Walker2d under reward perturbations in the entropy-based setting with a reward-shift radius of 3, and report their performance under the true reward. Table~\ref{tab:shaping_coefficient} shows that, under the same PPO training budget, the canonical reward with $\alpha=0$ achieves the best performance, and the return progressively deteriorates as $\alpha$ increases and more of the removed PBRS component is added back. This controlled experiment directly supports our explanation that the performance gain comes from removing the PBRS-redundant component rather than from an arbitrary reward transformation.

\begin{wraptable}{r}{0.52\textwidth}
\vspace{-1.5ex}
\centering
\caption{Ablation on canonicalization choices under reward perturbations for Walker2d with the entropy-based robust objective (reward shift radius $=3$). Results are reported as mean $\pm$ 95\% confidence interval over 5 random seeds.}
\label{tab:ablation_canon_walker2d_entropy}
\small
\setlength{\tabcolsep}{4pt}
\renewcommand{\arraystretch}{1.1}
\resizebox{0.52\textwidth}{!}{%
\begin{tabular}{lc}
\toprule
\textbf{Method} & \textbf{Walker2d} \\
\midrule
Reward-Robust (baseline)                       & $-259.59\pm16.07$ \\
Shaping-aware (ours, $\Omega^*$-projection)    & $-188.14\pm9.26$ \\
Shaping-aware (random $\Phi$ canonicalization) & $-226.17\pm36.72$ \\
Shaping-aware ($L^2$ projection)               & $-199.04\pm9.37$ \\
\bottomrule
\end{tabular}%
}
\vspace{-2ex}
\end{wraptable}

\textbf{Ablation on projection choices.}
To isolate the contribution of our \(\Omega^*\)-aligned canonicalization, Table~\ref{tab:ablation_canon_walker2d_entropy} compares several projection choices under reward perturbations. Implementation details are provided in Appendix~\ref{appendix:implement-ablation}, and full results for all environments are reported in Appendix~\ref{appendix:ablation}. First, we include a random $\Phi$ variant, which injects an arbitrary PBRS shaping term into the reward. This variant does not yield meaningful robustness improvements, indicating that the gains of our method do not come from simply adding an extra shaping-like component, but from learning a canonicalization that removes PBRS redundancy in a principled way. We also consider an $L^2$ projection that performs orthogonal projection regardless of the uncertainty-set geometry. This variant still removes PBRS-equivalent directions and thus improves robustness relative to the baseline, confirming that redundancy removal alone is beneficial. However, the $\Omega^*$-aligned projection performs better, suggesting that matching the canonicalization to the robustness geometry provides additional gains under reward perturbations.

\section{Conclusion}

In this paper, we propose \textbf{Shaping-Aware Reward-Robust RL}, which constructs behavior discriminative uncertainty sets over PBRS equivalence classes and enables practical training via a projection-learning module compatible with standard RL optimizers. Across diverse tasks, our method consistently improves robust performance over representative baselines. These results support our central claim that eliminating behaviorally equivalent rewards leads to more effective robust policy learning under reward uncertainty. We discuss limitations and future directions in Appendix~\ref{appendix:limitations}.

{
\small
\bibliographystyle{unsrt}
\bibliography{example_paper}

@article{mannor2007bias,
  title={Bias and variance approximation in value function estimates},
  author={Mannor, Shie and Simester, Duncan and Sun, Peng and Tsitsiklis, John N},
  journal={Management Science},
  volume={53},
  number={2},
  pages={308--322},
  year={2007},
  publisher={INFORMS}
}

@article{iyengar2005robust,
  title={Robust dynamic programming},
  author={Iyengar, Garud N},
  journal={Mathematics of Operations Research},
  volume={30},
  number={2},
  pages={257--280},
  year={2005},
  publisher={INFORMS}
}

@article{wiesemann2013robust,
  title={Robust Markov decision processes},
  author={Wiesemann, Wolfram and Kuhn, Daniel and Rustem, Ber{\c{c}}},
  journal={Mathematics of Operations Research},
  volume={38},
  number={1},
  pages={153--183},
  year={2013},
  publisher={INFORMS}
}

@article{mannor2012lightning,
  title={Lightning does not strike twice: Robust MDPs with coupled uncertainty},
  author={Mannor, Shie and Mebel, Ofir and Xu, Huan},
  journal={arXiv preprint arXiv:1206.4643},
  year={2012}
}

@inproceedings{derman2023robustness,
  title={Robustness and regularization in reinforcement learning},
  author={Derman, Esther and Men, Yevgeniy and Geist, Matthieu and Mannor, Shie},
  booktitle={NeurIPS 2023 Workshop on Generalization in Planning},
  year={2023}
}

@article{eysenbach2021maximum,
  title={Maximum entropy RL (provably) solves some robust RL problems},
  author={Eysenbach, Benjamin and Levine, Sergey},
  journal={arXiv preprint arXiv:2103.06257},
  year={2021}
}

@inproceedings{ng1999policy,
  title={Policy invariance under reward transformations: Theory and application to reward shaping},
  author={Ng, Andrew Y and Harada, Daishi and Russell, Stuart},
  booktitle={Icml},
  volume={99},
  pages={278--287},
  year={1999},
  organization={Citeseer}
}

@misc{
anonymous2025robust,
title={Robust Optimization for Mitigating Reward Hacking with Correlated Proxies},
author={Liu, Zixuan and Sun, Xiaolin and Zheng, Zizhan},
year={2025},
url={https://openreview.net/forum?id=0wW6Ml0qku}
}

@article{ashlag2025state,
  title={State entropy regularization for robust reinforcement learning},
  author={Ashlag, Yonatan and Koren, Uri and Mutti, Mirco and Derman, Esther and Bacon, Pierre-Luc and Mannor, Shie},
  journal={arXiv preprint arXiv:2506.07085},
  year={2025}
}

@book{hiriart2004fundamentals,
  title={Fundamentals of convex analysis},
  author={Hiriart-Urruty, Jean-Baptiste and Lemar{\'e}chal, Claude},
  year={2004},
  publisher={Springer Science \& Business Media}
}

@inproceedings{mensch2018differentiable,
  title={Differentiable dynamic programming for structured prediction and attention},
  author={Mensch, Arthur and Blondel, Mathieu},
  booktitle={International Conference on Machine Learning},
  pages={3462--3471},
  year={2018},
  organization={PMLR}
}

@article{bagnell2001solving,
  title={Solving uncertain Markov decision processes},
  author={Bagnell, J Andrew and Ng, Andrew Y and Schneider, Jeff G},
  year={2001},
  publisher={Carnegie Mellon University}
}

@inproceedings{grand2021scalable,
  title={Scalable first-order methods for robust mdps},
  author={Grand-Cl{\'e}ment, Julien and Kroer, Christian},
  booktitle={Proceedings of the AAAI Conference on Artificial Intelligence},
  volume={35},
  number={13},
  pages={12086--12094},
  year={2021}
}

@inproceedings{husain2021regularized,
  title={Regularized policies are reward robust},
  author={Husain, Hisham and Ciosek, Kamil and Tomioka, Ryota},
  booktitle={International Conference on Artificial Intelligence and Statistics},
  pages={64--72},
  year={2021},
  organization={PMLR}
}

@article{bubeck2015convex,
  title={Convex optimization: Algorithms and complexity},
  author={Bubeck, S{\'e}bastien and others},
  journal={Foundations and Trends{\textregistered} in Machine Learning},
  volume={8},
  number={3-4},
  pages={231--357},
  year={2015},
  publisher={Now Publishers, Inc.}
}

@article{zhan2023policy,
  title={Policy mirror descent for regularized reinforcement learning: A generalized framework with linear convergence},
  author={Zhan, Wenhao and Cen, Shicong and Huang, Baihe and Chen, Yuxin and Lee, Jason D and Chi, Yuejie},
  journal={SIAM Journal on Optimization},
  volume={33},
  number={2},
  pages={1061--1091},
  year={2023},
  publisher={SIAM}
}

@article{bossens2025mirror,
  title={Mirror Descent Policy Optimisation for Robust Constrained Markov Decision Processes},
  author={Bossens, David M and Nitanda, Atsushi},
  journal={arXiv preprint arXiv:2506.23165},
  year={2025}
}

@article{tomar2020mirror,
  title={Mirror descent policy optimization},
  author={Tomar, Manan and Shani, Lior and Efroni, Yonathan and Ghavamzadeh, Mohammad},
  journal={arXiv preprint arXiv:2005.09814},
  year={2020}
}

@Inbook{Borkar2008,
author="Borkar, Vivek S.",
title="Multiple Timescales",
bookTitle="Stochastic Approximation: A Dynamical Systems Viewpoint",
year="2008",
publisher="Hindustan Book Agency",
address="Gurgaon",
pages="64--77",
isbn="978-93-86279-38-5",
doi="10.1007/978-93-86279-38-5_6",
url="https://doi.org/10.1007/978-93-86279-38-5_6"
}

@article{doi:10.1137/S0363012997331639,
author = {Borkar, V. S. and Meyn, S. P.},
title = {The O.D.E. Method for Convergence of Stochastic Approximation and Reinforcement Learning},
journal = {SIAM Journal on Control and Optimization},
volume = {38},
number = {2},
pages = {447-469},
year = {2000},
doi = {10.1137/S0363012997331639},
URL = {https://doi.org/10.1137/S0363012997331639},
eprint = {https://doi.org/10.1137/S0363012997331639},
}

@incollection{ROBBINS1971233,
title = {A CONVERGENCE THEOREM FOR NON NEGATIVE ALMOST SUPERMARTINGALES AND SOME APPLICATIONS**Research supported by NIH Grant 5-R01-GM-16895-03 and ONR Grant N00014-67-A-0108-0018.},
editor = {Jagdish S. Rustagi},
booktitle = {Optimizing Methods in Statistics},
publisher = {Academic Press},
pages = {233-257},
year = {1971},
isbn = {978-0-12-604550-5},
doi = {https://doi.org/10.1016/B978-0-12-604550-5.50015-8},
url = {https://www.sciencedirect.com/science/article/pii/B9780126045505500158},
author = {H. Robbins and D. Siegmund},
}

@inproceedings{todorov2012mujoco,
  title={Mujoco: A physics engine for model-based control},
  author={Todorov, Emanuel and Erez, Tom and Tassa, Yuval},
  booktitle={2012 IEEE/RSJ international conference on intelligent robots and systems},
  pages={5026--5033},
  year={2012},
  organization={IEEE}
}

@article{brockman2016openai,
  title={Openai gym},
  author={Brockman, Greg and Cheung, Vicki and Pettersson, Ludwig and Schneider, Jonas and Schulman, John and Tang, Jie and Zaremba, Wojciech},
  journal={arXiv preprint arXiv:1606.01540},
  year={2016}
}

@article{paszke2019pytorch,
  title={Pytorch: An imperative style, high-performance deep learning library},
  author={Paszke, A},
  journal={arXiv preprint arXiv:1912.01703},
  year={2019}
}

@article{stable-baselines3,
  author  = {Antonin Raffin and Ashley Hill and Adam Gleave and Anssi Kanervisto and Maximilian Ernestus and Noah Dormann},
  title   = {Stable-Baselines3: Reliable Reinforcement Learning Implementations},
  journal = {Journal of Machine Learning Research},
  year    = {2021},
  volume  = {22},
  number  = {268},
  pages   = {1-8},
  url     = {http://jmlr.org/papers/v22/20-1364.html}
}

@article{liu2024enhancing,
  title={Enhancing llm safety via constrained direct preference optimization},
  author={Liu, Zixuan and Sun, Xiaolin and Zheng, Zizhan},
  journal={arXiv preprint arXiv:2403.02475},
  year={2024}
}

@article{kiran2021deep,
  title={Deep reinforcement learning for autonomous driving: A survey},
  author={Kiran, B Ravi and Sobh, Ibrahim and Talpaert, Victor and Mannion, Patrick and Al Sallab, Ahmad A and Yogamani, Senthil and P{\'e}rez, Patrick},
  journal={IEEE transactions on intelligent transportation systems},
  volume={23},
  number={6},
  pages={4909--4926},
  year={2021},
  publisher={IEEE}
}

@article{christiano2017deep,
  title={Deep reinforcement learning from human preferences},
  author={Christiano, Paul F and Leike, Jan and Brown, Tom and Martic, Miljan and Legg, Shane and Amodei, Dario},
  journal={Advances in neural information processing systems},
  volume={30},
  year={2017}
}

@article{kober2013reinforcement,
  title={Reinforcement learning in robotics: A survey},
  author={Kober, Jens and Bagnell, J Andrew and Peters, Jan},
  journal={The International Journal of Robotics Research},
  volume={32},
  number={11},
  pages={1238--1274},
  year={2013},
  publisher={SAGE Publications Sage UK: London, England}
}

@article{morimoto2005robust,
  title={Robust reinforcement learning},
  author={Morimoto, Jun and Doya, Kenji},
  journal={Neural computation},
  volume={17},
  number={2},
  pages={335--359},
  year={2005},
  publisher={MIT Press}
}

@article{lim2013reinforcement,
  title={Reinforcement learning in robust markov decision processes},
  author={Lim, Shiau Hong and Xu, Huan and Mannor, Shie},
  journal={Advances in Neural Information Processing Systems},
  volume={26},
  year={2013}
}

@inproceedings{pinto2017robust,
  title={Robust adversarial reinforcement learning},
  author={Pinto, Lerrel and Davidson, James and Sukthankar, Rahul and Gupta, Abhinav},
  booktitle={International conference on machine learning},
  pages={2817--2826},
  year={2017},
  organization={PMLR}
}

@article{muller2025improving,
  title={Improving the Effectiveness of Potential-Based Reward Shaping in Reinforcement Learning},
  author={M{\"u}ller, Henrik and Kudenko, Daniel},
  journal={arXiv preprint arXiv:2502.01307},
  year={2025}
}

@inproceedings{gadot2024solving,
  title={Solving non-rectangular reward-robust MDPs via frequency regularization},
  author={Gadot, Uri and Derman, Esther and Kumar, Navdeep and Elfatihi, Maxence Mohamed and Levy, Kfir and Mannor, Shie},
  booktitle={Proceedings of the AAAI Conference on Artificial Intelligence},
  volume={38},
  number={19},
  pages={21090--21098},
  year={2024}
}

@inproceedings{ahmed2019understanding,
  title={Understanding the impact of entropy on policy optimization},
  author={Ahmed, Zafarali and Le Roux, Nicolas and Norouzi, Mohammad and Schuurmans, Dale},
  booktitle={International conference on machine learning},
  pages={151--160},
  year={2019},
  organization={PMLR}
}

@article{neu2017unified,
  title={A unified view of entropy-regularized markov decision processes},
  author={Neu, Gergely and Jonsson, Anders and G{\'o}mez, Vicen{\c{c}}},
  journal={arXiv preprint arXiv:1705.07798},
  year={2017}
}

@inproceedings{haarnoja2018soft,
  title={Soft actor-critic: Off-policy maximum entropy deep reinforcement learning with a stochastic actor},
  author={Haarnoja, Tuomas and Zhou, Aurick and Abbeel, Pieter and Levine, Sergey},
  booktitle={International conference on machine learning},
  pages={1861--1870},
  year={2018},
  organization={Pmlr}
}

@article{islam2019marginalized,
  title={Marginalized state distribution entropy regularization in policy optimization},
  author={Islam, Riashat and Ahmed, Zafarali and Precup, Doina},
  journal={arXiv preprint arXiv:1912.05128},
  year={2019}
}

@inproceedings{seo2021state,
  title={State entropy maximization with random encoders for efficient exploration},
  author={Seo, Younggyo and Chen, Lili and Shin, Jinwoo and Lee, Honglak and Abbeel, Pieter and Lee, Kimin},
  booktitle={International conference on machine learning},
  pages={9443--9454},
  year={2021},
  organization={PMLR}
}

@article{yuan2022renyi,
  title={R{\'e}nyi state entropy maximization for exploration acceleration in reinforcement learning},
  author={Yuan, Mingqi and Pun, Man-On and Wang, Dong},
  journal={IEEE transactions on artificial intelligence},
  volume={4},
  number={5},
  pages={1154--1164},
  year={2022},
  publisher={IEEE}
}

@article{bolland2024off,
  title={Off-policy maximum entropy rl with future state and action visitation measures},
  author={Bolland, Adrien and Lambrechts, Gaspard and Ernst, Damien},
  journal={arXiv preprint arXiv:2412.06655},
  year={2024}
}

@article{kim2023accelerating,
  title={Accelerating reinforcement learning with value-conditional state entropy exploration},
  author={Kim, Dongyoung and Shin, Jinwoo and Abbeel, Pieter and Seo, Younggyo},
  journal={Advances in Neural Information Processing Systems},
  volume={36},
  pages={31811--31830},
  year={2023}
}

@inproceedings{choromanski2020provably,
  title={Provably robust blackbox optimization for reinforcement learning},
  author={Choromanski, Krzysztof and Pacchiano, Aldo and Parker-Holder, Jack and Tang, Yunhao and Jain, Deepali and Yang, Yuxiang and Iscen, Atil and Hsu, Jasmine and Sindhwani, Vikas},
  booktitle={Conference on robot learning},
  pages={683--696},
  year={2020},
  organization={PMLR}
}

@article{huttenrauch2024robust,
  title={Robust black-box optimization for stochastic search and episodic reinforcement learning},
  author={H{\"u}ttenrauch, Maximilian and Neumann, Gerhard},
  journal={Journal of Machine Learning Research},
  volume={25},
  number={153},
  pages={1--44},
  year={2024}
}

@inproceedings{kirschner2020distributionally,
  title={Distributionally robust Bayesian optimization},
  author={Kirschner, Johannes and Bogunovic, Ilija and Jegelka, Stefanie and Krause, Andreas},
  booktitle={International Conference on Artificial Intelligence and Statistics},
  pages={2174--2184},
  year={2020},
  organization={PMLR}
}

@inproceedings{inatsu2022bayesian,
  title={Bayesian optimization for distributionally robust chance-constrained problem},
  author={Inatsu, Yu and Takeno, Shion and Karasuyama, Masayuki and Takeuchi, Ichiro},
  booktitle={International Conference on Machine Learning},
  pages={9602--9621},
  year={2022},
  organization={PMLR}
}

@article{inatsu2025bayesian,
  title={Bayesian Optimization of Robustness Measures Using Randomized GP-UCB-based Algorithms under Input Uncertainty},
  author={Inatsu, Yu},
  journal={arXiv preprint arXiv:2504.03172},
  year={2025}
}

@book{bertsekas2009convex,
  title={Convex optimization theory},
  author={Bertsekas, Dimitri},
  volume={1},
  year={2009},
  publisher={Athena Scientific}
}

@article{schulman2017proximal,
  title={Proximal policy optimization algorithms},
  author={Schulman, John and Wolski, Filip and Dhariwal, Prafulla and Radford, Alec and Klimov, Oleg},
  journal={arXiv preprint arXiv:1707.06347},
  year={2017}
}

@book{borkar2008stochastic,
  title={Stochastic approximation: a dynamical systems viewpoint},
  author={Borkar, Vivek S and Borkar, Vivek S},
  volume={100},
  year={2008},
  publisher={Springer}
}

@article{chevalier2018minimalistic,
  title={Minimalistic gridworld environment for openai gym},
  author={Chevalier-Boisvert, Maxime and Willems, Lucas and Pal, Suman},
  year={2018}
}

@article{yan2024reward,
  title={Reward-robust rlhf in llms},
  author={Yan, Yuzi and Lou, Xingzhou and Li, Jialian and Zhang, Yiping and Xie, Jian and Yu, Chao and Wang, Yu and Yan, Dong and Shen, Yuan},
  journal={arXiv preprint arXiv:2409.15360},
  year={2024}
}

@inproceedings{wang2020reinforcement,
  title={Reinforcement learning with perturbed rewards},
  author={Wang, Jingkang and Liu, Yang and Li, Bo},
  booktitle={Proceedings of the AAAI conference on artificial intelligence},
  volume={34},
  number={04},
  pages={6202--6209},
  year={2020}
}

@article{he2023robotic,
  title={Robotic control in adversarial and sparse reward environments: A robust goal-conditioned reinforcement learning approach},
  author={He, Xiangkun and Lv, Chen},
  journal={IEEE Transactions on Artificial Intelligence},
  volume={5},
  number={1},
  pages={244--253},
  year={2023},
  publisher={IEEE}
}

@article{mannor2016robust,
  title={Robust MDPs with k-rectangular uncertainty},
  author={Mannor, Shie and Mebel, Ofir and Xu, Huan},
  journal={Mathematics of Operations Research},
  volume={41},
  number={4},
  pages={1484--1509},
  year={2016},
  publisher={INFORMS}
}

@article{goyal2023robust,
  title={Robust markov decision processes: Beyond rectangularity},
  author={Goyal, Vineet and Grand-Clement, Julien},
  journal={Mathematics of Operations Research},
  volume={48},
  number={1},
  pages={203--226},
  year={2023},
  publisher={INFORMS}
}

@article{behzadian2021fast,
  title={Fast Algorithms for L-constrained S-rectangular Robust MDPs},
  author={Behzadian, Bahram and Petrik, Marek and Ho, Chin Pang},
  journal={Advances in Neural Information Processing Systems},
  volume={34},
  pages={25982--25992},
  year={2021}
}

@article{ho2021partial,
  title={Partial policy iteration for l1-robust markov decision processes},
  author={Ho, Chin Pang and Petrik, Marek and Wiesemann, Wolfram},
  journal={Journal of Machine Learning Research},
  volume={22},
  number={275},
  pages={1--46},
  year={2021}
}

@inproceedings{tessler2019action,
  title={Action robust reinforcement learning and applications in continuous control},
  author={Tessler, Chen and Efroni, Yonathan and Mannor, Shie},
  booktitle={International Conference on Machine Learning},
  pages={6215--6224},
  year={2019},
  organization={PMLR}
}

@inproceedings{haarnoja2017reinforcement,
  title={Reinforcement learning with deep energy-based policies},
  author={Haarnoja, Tuomas and Tang, Haoran and Abbeel, Pieter and Levine, Sergey},
  booktitle={International conference on machine learning},
  pages={1352--1361},
  year={2017},
  organization={PMLR}
}

@article{lee2018sparse,
  title={Sparse markov decision processes with causal sparse tsallis entropy regularization for reinforcement learning},
  author={Lee, Kyungjae and Choi, Sungjoon and Oh, Songhwai},
  journal={IEEE Robotics and Automation Letters},
  volume={3},
  number={3},
  pages={1466--1473},
  year={2018},
  publisher={IEEE}
}

@inproceedings{schulman2015trust,
  title={Trust region policy optimization},
  author={Schulman, John and Levine, Sergey and Abbeel, Pieter and Jordan, Michael and Moritz, Philipp},
  booktitle={International conference on machine learning},
  pages={1889--1897},
  year={2015},
  organization={PMLR}
}

@article{xu2009robustness,
  title={Robustness and regularization of support vector machines.},
  author={Xu, Huan and Caramanis, Constantine and Mannor, Shie},
  journal={Journal of machine learning research},
  volume={10},
  number={7},
  year={2009}
}

@article{hoffman2019robust,
  title={Robust learning with jacobian regularization},
  author={Hoffman, Judy and Roberts, Daniel A and Yaida, Sho},
  journal={arXiv preprint arXiv:1908.02729},
  year={2019}
}

@article{duchi2021statistics,
  title={Statistics of robust optimization: A generalized empirical likelihood approach},
  author={Duchi, John C and Glynn, Peter W and Namkoong, Hongseok},
  journal={Mathematics of Operations Research},
  volume={46},
  number={3},
  pages={946--969},
  year={2021},
  publisher={INFORMS}
}

@article{shafieezadeh2015distributionally,
  title={Distributionally robust logistic regression},
  author={Shafieezadeh Abadeh, Soroosh and Mohajerin Esfahani, Peyman M and Kuhn, Daniel},
  journal={Advances in neural information processing systems},
  volume={28},
  year={2015}
}

@article{wiewiora2003potential,
  title={Potential-based shaping and Q-value initialization are equivalent},
  author={Wiewiora, Eric},
  journal={Journal of Artificial Intelligence Research},
  volume={19},
  pages={205--208},
  year={2003}
}

@inproceedings{asmuth2008potential,
  title={Potential-based Shaping in Model-based Reinforcement Learning.},
  author={Asmuth, John and Littman, Michael L and Zinkov, Robert},
  booktitle={AAAI},
  pages={604--609},
  year={2008}
}

@article{ma2024highly,
  title={Highly efficient self-adaptive reward shaping for reinforcement learning},
  author={Ma, Haozhe and Luo, Zhengding and Vo, Thanh Vinh and Sima, Kuankuan and Leong, Tze-Yun},
  journal={arXiv preprint arXiv:2408.03029},
  year={2024}
}

@article{berner2019dota,
  title={Dota 2 with large scale deep reinforcement learning},
  author={Berner, Christopher and Brockman, Greg and Chan, Brooke and Cheung, Vicki and Debiak, Przemyslaw and Dennison, Christy and Farhi, David and Fischer, Quirin and Hashme, Shariq and Hesse, Chris and others},
  journal={arXiv preprint arXiv:1912.06680},
  year={2019}
}

@article{ye2020towards,
  title={Towards playing full moba games with deep reinforcement learning},
  author={Ye, Deheng and Chen, Guibin and Zhang, Wen and Chen, Sheng and Yuan, Bo and Liu, Bo and Chen, Jia and Liu, Zhao and Qiu, Fuhao and Yu, Hongsheng and others},
  journal={Advances in Neural Information Processing Systems},
  volume={33},
  pages={621--632},
  year={2020}
}

@inproceedings{yu2020meta,
  title={Meta-world: A benchmark and evaluation for multi-task and meta reinforcement learning},
  author={Yu, Tianhe and Quillen, Deirdre and He, Zhanpeng and Julian, Ryan and Hausman, Karol and Finn, Chelsea and Levine, Sergey},
  booktitle={Conference on robot learning},
  pages={1094--1100},
  year={2020},
  organization={PMLR}
}

@article{mahankali2024random,
  title={Random latent exploration for deep reinforcement learning},
  author={Mahankali, Srinath and Hong, Zhang-Wei and Sekhari, Ayush and Rakhlin, Alexander and Agrawal, Pulkit},
  journal={arXiv preprint arXiv:2407.13755},
  year={2024}
}

@article{liu2025skill,
  title={Skill expansion and composition in parameter space},
  author={Liu, Tenglong and Li, Jianxiong and Zheng, Yinan and Niu, Haoyi and Lan, Yixing and Xu, Xin and Zhan, Xianyuan},
  journal={arXiv preprint arXiv:2502.05932},
  year={2025}
}

@article{devidze2022exploration,
  title={Exploration-guided reward shaping for reinforcement learning under sparse rewards},
  author={Devidze, Rati and Kamalaruban, Parameswaran and Singla, Adish},
  journal={Advances in Neural Information Processing Systems},
  volume={35},
  pages={5829--5842},
  year={2022}
}

@article{badia2020never,
  title={Never give up: Learning directed exploration strategies},
  author={Badia, Adri{\`a} Puigdom{\`e}nech and Sprechmann, Pablo and Vitvitskyi, Alex and Guo, Daniel and Piot, Bilal and Kapturowski, Steven and Tieleman, Olivier and Arjovsky, Mart{\'\i}n and Pritzel, Alexander and Bolt, Andew and others},
  journal={arXiv preprint arXiv:2002.06038},
  year={2020}
}

@inproceedings{lobel2023flipping,
  title={Flipping coins to estimate pseudocounts for exploration in reinforcement learning},
  author={Lobel, Sam and Bagaria, Akhil and Konidaris, George},
  booktitle={International Conference on Machine Learning},
  pages={22594--22613},
  year={2023},
  organization={PMLR}
}

@inproceedings{machado2020count,
  title={Count-based exploration with the successor representation},
  author={Machado, Marlos C and Bellemare, Marc G and Bowling, Michael},
  booktitle={Proceedings of the AAAI Conference on Artificial Intelligence},
  volume={34},
  number={04},
  pages={5125--5133},
  year={2020}
}

@article{choshen2018dora,
  title={Dora the explorer: Directed outreaching reinforcement action-selection},
  author={Choshen, Leshem and Fox, Lior and Loewenstein, Yonatan},
  journal={arXiv preprint arXiv:1804.04012},
  year={2018}
}

@article{tang2017exploration,
  title={\# exploration: A study of count-based exploration for deep reinforcement learning},
  author={Tang, Haoran and Houthooft, Rein and Foote, Davis and Stooke, Adam and Xi Chen, OpenAI and Duan, Yan and Schulman, John and DeTurck, Filip and Abbeel, Pieter},
  journal={Advances in neural information processing systems},
  volume={30},
  year={2017}
}

@article{bellemare2016unifying,
  title={Unifying count-based exploration and intrinsic motivation},
  author={Bellemare, Marc and Srinivasan, Sriram and Ostrovski, Georg and Schaul, Tom and Saxton, David and Munos, Remi},
  journal={Advances in neural information processing systems},
  volume={29},
  year={2016}
}

@article{cheng2021heuristic,
  title={Heuristic-guided reinforcement learning},
  author={Cheng, Ching-An and Kolobov, Andrey and Swaminathan, Adith},
  journal={Advances in Neural Information Processing Systems},
  volume={34},
  pages={13550--13563},
  year={2021}
}

@article{gupta2022unpacking,
  title={Unpacking reward shaping: Understanding the benefits of reward engineering on sample complexity},
  author={Gupta, Abhishek and Pacchiano, Aldo and Zhai, Yuexiang and Kakade, Sham and Levine, Sergey},
  journal={Advances in Neural Information Processing Systems},
  volume={35},
  pages={15281--15295},
  year={2022}
}

@inproceedings{jeon2020reward,
  title={Reward-rational (implicit) choice: A unifying formalism for reward learning},
  author={Jeon, Hong Jun and Milli, Smitha and Dragan, Anca},
  booktitle={Advances in Neural Information Processing Systems},
  year={2020}
}

@inproceedings{filippi2010optimism,
  title={Optimism in reinforcement learning and Kullback-Leibler divergence},
  author={Filippi, Sarah and Capp{\'e}, Olivier and Garivier, Aur{\'e}lien},
  booktitle={2010 48th Annual Allerton Conference on Communication, Control, and Computing (Allerton)},
  pages={115--122},
  year={2010},
  organization={IEEE}
}

@article{lidayan2024bamdp,
  title={Bamdp shaping: a unified framework for intrinsic motivation and reward shaping},
  author={Lidayan, Aly and Dennis, Michael and Russell, Stuart},
  journal={arXiv preprint arXiv:2409.05358},
  year={2024}
}

@article{gao2024hierarchical,
  title={Hierarchical reinforcement learning from demonstration via reachability-based reward shaping},
  author={Gao, Xiaozhu and Liu, Jinhui and Wan, Bo and An, Lingling},
  journal={Neural Processing Letters},
  volume={56},
  number={3},
  pages={184},
  year={2024},
  publisher={Springer}
}

@article{lu2019predictive,
  title={Predictive coding for boosting deep reinforcement learning with sparse rewards},
  author={Lu, Xingyu and Tiomkin, Stas and Abbeel, Pieter},
  journal={arXiv preprint arXiv:1912.13414},
  year={2019}
}

@article{xie2023text2reward,
  title={Text2reward: Reward shaping with language models for reinforcement learning},
  author={Xie, Tianbao and Zhao, Siheng and Wu, Chen Henry and Liu, Yitao and Luo, Qian and Zhong, Victor and Yang, Yanchao and Yu, Tao},
  journal={arXiv preprint arXiv:2309.11489},
  year={2023}
}

@article{zhang2024orso,
  title={ORSO: Accelerating Reward Design via Online Reward Selection and Policy Optimization},
  author={Zhang, Chen Bo Calvin and Hong, Zhang-Wei and Pacchiano, Aldo and Agrawal, Pulkit},
  journal={arXiv preprint arXiv:2410.13837},
  year={2024}
}

@inproceedings{chen2024improving,
  title={Improving Discriminative Capability of Reward Models in RLHF Using Contrastive Learning},
  author={Chen, Lu and Zheng, Rui and Wang, Binghai and Jin, Senjie and Huang, Caishuang and Ye, Junjie and Zhang, Zhihao and Zhou, Yuhao and Xi, Zhiheng and Gui, Tao and others},
  booktitle={Proceedings of the 2024 Conference on Empirical Methods in Natural Language Processing},
  pages={15270--15283},
  year={2024}
}

@article{gupta2023behavior,
  title={Behavior alignment via reward function optimization},
  author={Gupta, Dhawal and Chandak, Yash and Jordan, Scott and Thomas, Philip S and C da Silva, Bruno},
  journal={Advances in Neural Information Processing Systems},
  volume={36},
  pages={52759--52791},
  year={2023}
}

@article{enders2024risk,
  title={Risk-sensitive soft actor-critic for robust deep reinforcement learning under distribution shifts},
  author={Enders, Tobias and Harrison, James and Schiffer, Maximilian},
  journal={arXiv preprint arXiv:2402.09992},
  year={2024}
}

@article{kumar2023policy,
  title={Policy gradient for rectangular robust markov decision processes},
  author={Kumar, Navdeep and Derman, Esther and Geist, Matthieu and Levy, Kfir Y and Mannor, Shie},
  journal={Advances in Neural Information Processing Systems},
  volume={36},
  pages={59477--59501},
  year={2023}
}

@article{kumar2025dual,
  title={Dual Formulation for Non-Rectangular Lp Robust Markov Decision Processes},
  author={Kumar, Navdeep and Gupta, Adarsh and Elfatihi, Maxence Mohamed and Ramponi, Giorgia and Levy, Kfir Yehuda and Mannor, Shie},
  journal={arXiv preprint arXiv:2502.09432},
  year={2025}
}

@article{bellemare2013arcade,
  title={The arcade learning environment: An evaluation platform for general agents},
  author={Bellemare, Marc G and Naddaf, Yavar and Veness, Joel and Bowling, Michael},
  journal={Journal of artificial intelligence research},
  volume={47},
  pages={253--279},
  year={2013}
}

@article{rajeswaran2017learning,
  title={Learning complex dexterous manipulation with deep reinforcement learning and demonstrations},
  author={Rajeswaran, Aravind and Kumar, Vikash and Gupta, Abhishek and Vezzani, Giulia and Schulman, John and Todorov, Emanuel and Levine, Sergey},
  journal={arXiv preprint arXiv:1709.10087},
  year={2017}
}

@article{hoeffding1963probability,
  title={Probability inequalities for sums of bounded random variables},
  author={Hoeffding, Wassily},
  journal={Journal of the American statistical association},
  volume={58},
  number={301},
  pages={13--30},
  year={1963},
  publisher={Taylor \& Francis}
}

@book{vershynin2018high,
  title={High-dimensional probability: An introduction with applications in data science},
  author={Vershynin, Roman},
  volume={47},
  year={2018},
  publisher={Cambridge university press}
}

@article{zhang2018three,
  title={Three mechanisms of weight decay regularization},
  author={Zhang, Guodong and Wang, Chaoqi and Xu, Bowen and Grosse, Roger},
  journal={arXiv preprint arXiv:1810.12281},
  year={2018}
}

@article{jiang2022holderian,
  title={H{\"o}lderian error bounds and kurdyka-{\l}ojasiewicz inequality for the trust region subproblem},
  author={Jiang, Rujun and Li, Xudong},
  journal={Mathematics of Operations Research},
  volume={47},
  number={4},
  pages={3025--3050},
  year={2022},
  publisher={INFORMS}
}

@inproceedings{liao2024error,
  title={Error bounds, PL condition, and quadratic growth for weakly convex functions, and linear convergences of proximal point methods},
  author={Liao, Feng-Yi and Ding, Lijun and Zheng, Yang},
  booktitle={6th Annual Learning for Dynamics \& Control Conference},
  pages={993--1005},
  year={2024},
  organization={PMLR}
}

@inproceedings{skalse2023invariance,
  title={Invariance in policy optimisation and partial identifiability in reward learning},
  author={Skalse, Joar Max Viktor and Farrugia-Roberts, Matthew and Russell, Stuart and Abate, Alessandro and Gleave, Adam},
  booktitle={International Conference on Machine Learning},
  pages={32033--32058},
  year={2023},
  organization={PMLR}
}

@book{puterman2014markov,
  title={Markov decision processes: discrete stochastic dynamic programming},
  author={Puterman, Martin L},
  year={2014},
  publisher={John Wiley \& Sons}
}

@article{kearns2002near,
  title={Near-optimal reinforcement learning in polynomial time},
  author={Kearns, Michael and Singh, Satinder},
  journal={Machine learning},
  volume={49},
  number={2},
  pages={209--232},
  year={2002},
  publisher={Springer}
}

@article{macua2021fully,
  title={Fully distributed actor-critic architecture for multitask deep reinforcement learning},
  author={Macua, Sergio Valcarcel and Davies, Ian and Tukiainen, Aleksi and De Cote, Enrique Munoz},
  journal={The Knowledge Engineering Review},
  volume={36},
  pages={e6},
  year={2021},
  publisher={Cambridge University Press}
}

@article{liu2026robust,
  title={Robust general utility for reinforcement learning},
  author={Liu, Zixuan and Wu, Fangzheng and Summa, Brian and Zheng, Zizhan},
  journal={arXiv preprint arXiv:2608.03562},
  year={2026}
}

@inproceedings{liu2026risk,
  title={Risk-Conditioned Fine-Tuning of Large Language Models},
  author={Liu, Zixuan and Fangzheng, WU and Summa, Brian and Zheng, Zizhan},
  booktitle={Proceedings of the 2026 Conference on Empirical Methods in Natural Language Processing},
  pages={22036--22076},
  year={2026}
}

@inproceedings{liu2026ua,
  title={UA-TTRL: Uncertainty-Aware Test-Time Reinforcement Learning},
  author={Liu, Zixuan and Khajavi, Siavash and Jiang, Guangkai and Liu, Xinru},
  booktitle={ICASSP 2026-2026 IEEE International Conference on Acoustics, Speech and Signal Processing (ICASSP)},
  pages={17147--17151},
  year={2026},
  organization={IEEE}
}

@inproceedings{liu2026targeting,
  title={Targeting misalignment: A conflict-aware framework for reward-model-based llm alignment},
  author={Liu, Zixuan and Khajavi, Siavash H and Jiang, Guangkai and Liu, Xinru},
  booktitle={Proceedings of the AAAI Conference on Artificial Intelligence},
  volume={40},
  number={44},
  pages={37692--37700},
  year={2026}
}

}


\appendix
\newpage

\section*{Acknowledgments}
This work was supported in part by NSF grant CNS-2146548, a grant from the Louisiana Board of Regents, a gift from Coefficient Giving, and funding from Tulane University Jurist Center for
Artificial Intelligence. We thank the anonymous reviewers for their insightful and constructive feedback.

\section{Limitations}
\label{appendix:limitations}
Despite the effectiveness of our framework, several limitations remain. First, our current convergence guarantee relies on a strong convexity assumption for the projection objective, which does not directly apply to the neural-network setting and remains a limitation of the current work. Theoretically, this assumption is used to ensure a well-behaved projection and support convergence to the desired optimum over the projection parameters $\varphi$. In practice, to improve the conditioning of the projection objective and facilitate optimization, we add a ridge term $\frac{\lambda}{2}|\varphi|^2$ to the projection loss, as described in Appendix~\ref{appendix:implementation-mujuco}. Although the projection is still learned only approximately in this deep setting, this regularization helps stabilize the projection-learning process in practice~\cite{zhang2018three} and leads to promising empirical performance in our experiments. We also emphasize that, even without strong convexity, one can still analyze convergence under weaker conditions such as Hölderian error bounds~\cite{jiang2022holderian} or the Polyak-Łojasiewicz inequality~\cite{liao2024error}, which are widely used in the nonconvex optimization literature and are more realistic for neural networks. However, pursuing such an extension is beyond the scope of the current paper. We believe it is common in RL research to present theory under stronger, more tractable assumptions than those satisfied by the neural-network implementation used in experiments. In particular, much of RL theory is still built around tabular, linear, or otherwise structured function classes, while practical success often relies on deep neural networks, for which comparable end-to-end guarantees are much harder to obtain. For example,~\cite{macua2021fully} derives a two time-scale convergence result under Assumptions 6-7, which do not hold for the deep neural network used in their experiments, but are introduced to illustrate the behavior of the proposed algorithm (page 12). Overall, we view our current theory as a simplified analysis that highlights the main idea, and the lack of strong convexity for neural networks remains a limitation.

Second, our theory builds on PBRS equivalence relations, and thus, the benefit of the method may be limited when reward misspecification is not PBRS-type redundant. In fact, many widely used uncertainty sets in reward-robust RL literature, including $\ell_p$-norm, entropy-based, and KL-based uncertainty sets, do contain PBRS-type redundancy as shown in the paper. We also acknowledge that there may be settings where the redundancy is not of PBRS type. For example,~\cite{skalse2023invariance} shows that reward learning and policy optimization can exhibit additional forms of reward non-identifiability beyond PBRS: rewards that differ on unreachable transitions, differ only by next-state redistribution, differ by positive linear scaling, or belong to broader optimality-preserving classes may induce the same downstream object or policy behavior. Extending our framework to handle such forms of redundancy, or more general behavior-discriminative uncertainty sets beyond PBRS, is not covered in the current paper and is an important direction for future work.

\section{Related Work}
\label{appendix:related-work}
\subsection{Reward-Robust RL and Regularization RL}

Reward-robust RL studies how to learn policies that remain effective when the reward signal is perturbed in an adversarial manner~\cite{mannor2007bias}, and has been explored in a broad range of settings~\cite{morimoto2005robust,wang2020reinforcement,he2023robotic,yan2024reward,liu2026robust}. A common view is to think of the problem as a dynamic zero-sum game: the agent selects a policy to maximize return, while an adversary selects a reward within a prescribed uncertainty set to minimize it~\cite{iyengar2005robust,wiesemann2013robust}. This leads to a max-min optimization problem that is often expensive to solve, creating practical bottlenecks for large-scale applications~\cite{tessler2019action}. To improve tractability, recent work has pursued several directions. Some approaches exploit structured uncertainty sets that admit more efficient optimization~\cite{mannor2016robust,mannor2012lightning,goyal2023robust}. In contrast, a larger body of work adopts rectangular uncertainty across states or state-action pairs to derive scalable solvers~\cite{behzadian2021fast,ho2021partial,bagnell2001solving,grand2021scalable}. In parallel, the strong empirical performance of regularized policy optimization has motivated many methods that use explicit regularizers to encourage desirable behavior, such as better exploration~\cite{haarnoja2017reinforcement,lee2018sparse,ahmed2019understanding,neu2017unified}, improved training stability~\cite{schulman2015trust,haarnoja2018soft}, and improved sample efficiency~\cite{seo2021state}.

More recently, there is growing recognition that certain robust RL objectives are closely connected to regularized RL formulations. Such robustness-regularization duality typically relies on the Legendre-Fenchel transform, which has been well established in statistical learning~\cite{xu2009robustness,hoffman2019robust,duchi2021statistics,shafieezadeh2015distributionally}. In the RL setting,~\cite{derman2023robustness} formally proves that policy regularization methods solve a particular instance of robust MDPs with uncertain rewards. A representative example is~\cite{eysenbach2021maximum}, which shows that maximum policy entropy regularized RL maximizes a lower bound on a reward-robust RL objective, and thus can be used to learn policies that are robust to some disturbances in the reward function. Subsequent work has expanded this perspective beyond policy regularization to state- and state-action-based regularizer, which induce robustness against adversaries that depend on global patterns rather than only local information~\cite{islam2019marginalized,seo2021state,yuan2022renyi,kim2023accelerating,bolland2024off}. In particular,~\cite{husain2021regularized} studies a broad family of regularized RL objectives and derives the corresponding adversarial-reward dual problems, and~\cite{ashlag2025state} analyzes the special case of state-entropy regularization, demonstrating improved robustness to structured, spatially correlated perturbations.

In this work, we build on the same Legendre-Fenchel duality and show that the regularization-robustness correspondence can be extended to our shaping-aware reward-robust objective. This equivalence enables us to avoid solving the inner minimization on the projected uncertainty set directly and instead optimize an associated regularized objective, substantially reducing the computational overhead of our approach.

\subsection{Potential-Based Reward Shaping}
Potential-based reward shaping (PBRS) was introduced by~\cite{ng1999policy} as a principled way to modify rewards without changing the optimal policy. The key idea is to add a shaping term defined by the difference of a potential function evaluated at two states, which can be canceled out in the Bellman equation, thus preserving the optimal policy~\cite{wiewiora2003potential,asmuth2008potential}. PBRS is generally used to incorporate heuristics or task-agnostic metrics that encourage broader state exploration~\cite{ma2024highly}. Accordingly, much of the literature focuses on how to construct effective potential functions. In some large-scale practical RL tasks, the potential functions are heavily handcrafted based on prior knowledge~\cite{berner2019dota,ye2020towards,yu2020meta}. Besides reward
engineering, a distinct line of work applies exploration bonuses to reward novel or infrequently visited states~\cite{mahankali2024random,liu2025skill,devidze2022exploration,badia2020never}. For example, count-based reward shaping tracks visitation counts and assigns higher rewards to less frequently visited states~\cite {lobel2023flipping,machado2020count,choshen2018dora}. However, in continuous spaces, exact state counting is typically infeasible, and prior work mainly relies on approximations. For example,~\cite{tang2017exploration} uses hashing to discretize the state space and enable count-style bonuses, while~\cite{bellemare2016unifying} proposes pseudo-counts derived from recording probabilities. Beyond count-based reward shaping, several works incorporate structured domain knowledge into shaping to guide exploration more directly, for example, by using heuristic shaping to reduce the effective horizon~\cite{cheng2021heuristic} or by exploiting shaping to narrow the portion of the state space that must be searched~\cite{gupta2022unpacking}. Recent approaches also consider alternative feedback signals, such as success rates derived from historical experience as the shaped rewards, to balance exploration and exploitation in a more adaptive way~\cite{ma2024highly}.

Beyond exploration, PBRS has also been used as a general mechanism for integrating task structure and prior knowledge into RL objectives. For instance,~\cite{gao2024hierarchical} proposes a reachability-based shaping signal for hierarchical RL to alleviate the negative interference of suboptimal demonstrations, where shaped rewards encode progress toward subgoals, enabling efficient skill learning. In addition, reward shaping has been adopted in deep RL practice as a way to mitigate sparse reward problems.~\cite{lu2019predictive} introduces a predictive coding-based shaping objective that understands the structure and dynamics of the environment and emphasize on features most useful for learning, improving learning stability under sparse rewards. More recently, reward shaping has been revisited through the lens of foundation models. Text2Reward~\cite{xie2023text2reward} uses language models to generate reward signals from natural-language task descriptions, reflecting an emerging direction where shaping is produced from semantic feedback rather than hand-designed potentials. 

Our paper is also related to a broader line of work on reward design
and reward optimization as a method of shaping agent behavior. For instance, behavior alignment via reward optimization directly treats the reward
as a design variable to induce desired behaviors~\cite{gupta2023behavior},
while online reward selection methods intertwine reward selection with
policy optimization to improve sample efficiency in reward design loops~\cite{zhang2024orso}.
In RLHF, recent work studies how to improve the discriminative power of learned
reward models (e.g., via contrastive objectives), which are conceptually adjacent to our emphasis on behavior-discriminative reward representations~\cite{chen2024improving}.

In contrast to prior work, which primarily designs shaped rewards to improve exploration or induce desired behaviors, we use reward shaping as a tool for robustness modeling. To the best of our knowledge, we are the first to bring PBRS into reward-robust RL to construct uncertainty sets over behaviorally meaningful variations rather than variations in raw reward values. Leveraging PBRS’s policy invariance property, we identify and remove shaping-induced redundancies in standard uncertainty sets, yielding a reduced set in which each remaining reward corresponds to a behaviorally distinct objective.

\section{Proofs}

\subsection{Proof of Lemma~\ref{lemma:shaping}}
\label{appendix:lemma-pbrs}
\begin{proof}
    Fix any policy \(\pi\) and initial state \(s_0=s\). For all \(s \in \mathcal{S}\), define $v^{\pi}_{r}(s)
= \mathbb{E}\!\left[\sum_{t=0}^{\infty} \gamma^{t} r(s_t,a_t,s_{t+1})
  \,\Big|\, s_0 = s,\pi,P \right]$ as the value function at state \(s\).
Consider the discounted return under reward \(r'\):
\begin{align*}
v^{\pi}_{r'}(s)
&= \mathbb{E}\!\left[\sum_{t=0}^{\infty}\gamma^t\bigl(r(s_t,a_t,s_{t+1})+(\gamma\Phi(s_{t+1})-\Phi(s_t))\bigr)\,\Big|\,s_0=s,\pi,P\right] \\
&= v^{\pi}_{r}(s)
\;+\;\,\mathbb{E}\!\left[\sum_{t=0}^{\infty}\gamma^{t+1}\Phi(s_{t+1 })-\sum_{t=0}^{\infty}\gamma^t\Phi(s_t)\,\Big|\,s_0=s,\pi,P\right].
\end{align*}
The first sum can be re-indexed as
\(\sum_{t=0}^{\infty}\gamma^{t+1}\Phi(s_{t+1})=\sum_{t=1}^{\infty}\gamma^t\Phi(s_t)\).
Therefore the two series telescope:
\[
\sum_{t=1}^{\infty}\gamma^t\Phi(s_t)-\sum_{t=0}^{\infty}\gamma^t\Phi(s_t)
= -\Phi(s_0).
\]
Hence
\[
v^{\pi}_{r'}(s)=v^{\pi}_{r}(s)-\Phi(s_0),
\]
and consequently
\[
J(\pi,r')
=\mathbb{E}_{s_0\sim\mu_0}\!\bigl[v^{\pi}_{r}(s_0)\bigr]
=J(\pi,r)-\mathbb{E}_{s_0\sim\mu_0}[\Phi(s_0)].
\]
The shift \(-\mathbb{E}_{s_0\sim\mu_0}[\Phi(s_0)]\) does not depend on \(\pi\). So within each equivalence class, all rewards differ by a
policy-independent constant shift. This completes the proof.
\end{proof}

\subsection{Proof of Proposition~\ref{prop:affine_redundancy}}
\label{appendix:propo-redudant}
\begin{proof}
Fix \((\mu_0,\pi,P)\) and view rewards as vectors in the finite-dimensional space
\(\mathbb{R}^{\mathcal{X'}}\). Since \(\mathcal{U}\) has nonempty interior around \(\tilde r\), for any chosen norm \(\|\cdot\|\) on \(\mathbb{R}^{\mathcal{X'}}\), there exists
\(\rho>0\) such that the open ball
\[
B(\tilde r,\rho)\;:=\;\{u\in \mathbb{R}^{\mathcal{X'}}:\|u-\tilde r\|<\rho\}
\subseteq \mathcal{U}.
\]

Next, choose any potential function \(\Phi\in\mathbb{R}^{\mathcal{S}}\) that is not constant
(e.g., pick two states \(s_1\neq s_2\) and set \(\Phi(s_1)=1,\Phi(s_2)=0\), and \(\Phi=0\) elsewhere).
Define the corresponding shaping term \(F\in \mathbb{R}^{\mathcal{X'}}\) by
\(F(s,a,s')=\gamma\Phi(s')-\Phi(s)\).
Because \(\Phi\) is not constant, \(F\neq 0\), hence \(\|F\|>0\).
Let \(\delta := \rho/\|F\|\). Then for any \(\alpha\in\mathbb{R}\) with \(|\alpha|\le \delta\),
\[
\|\tilde r+\alpha F-\tilde r\| = \|\alpha F\| \le |\alpha|\,\|F\| \le \delta \|F\|=\rho,
\]
which implies \(\tilde r+\alpha F \in B(\tilde r,\rho)\subseteq \mathcal{U}\).

It remains to show PBRS equivalence. Fix any policy \(\pi\) and initial state \(s_0=s\). For all \(s \in \mathcal{S}\), define $v^{\pi}_{r}(s)
= \mathbb{E}\!\left[\sum_{t=0}^{\infty} \gamma^{t} r(s_t,a_t,s_{t+1})
  \,\Big|\, s_0 = s,\pi,P \right]$ as the value function at state \(s\).
Consider the discounted return under reward \(\tilde r+\alpha F\):
\begin{align*}
v^{\pi}_{\tilde r+\alpha F}(s)
&= \mathbb{E}\!\left[\sum_{t=0}^{\infty}\gamma^t\bigl(\tilde r(s_t,a_t,s_{t+1})+\alpha(\gamma\Phi(s_{t+1})-\Phi(s_t))\bigr)\,\Big|\,s_0=s,\pi,P\right] \\
&= v^{\pi}_{\tilde r}(s)
\;+\;\alpha\,\mathbb{E}\!\left[\sum_{t=0}^{\infty}\gamma^{t+1}\Phi(s_{t+1 })-\sum_{t=0}^{\infty}\gamma^t\Phi(s_t)\,\Big|\,s_0=s,\pi,P\right].
\end{align*}
The first sum can be re-indexed as
\(\sum_{t=0}^{\infty}\gamma^{t+1}\Phi(s_{t+1})=\sum_{t=1}^{\infty}\gamma^t\Phi(s_t)\).
Therefore the two series telescope:
\[
\sum_{t=1}^{\infty}\gamma^t\Phi(s_t)-\sum_{t=0}^{\infty}\gamma^t\Phi(s_t)
= -\Phi(s_0).
\]
Hence
\[
v^{\pi}_{\tilde r+\alpha F}(s)=v^{\pi}_{\tilde r}(s)-\alpha\,\Phi(s_0),
\]
and consequently
\[
J(\pi,\tilde r+\alpha F)
=\mathbb{E}_{s_0\sim\mu_0}\!\bigl[v^{\pi}_{\tilde r+\alpha F}(s_0)\bigr]
=J(\pi,\tilde r)-\alpha\,\mathbb{E}_{s_0\sim\mu_0}[\Phi(s_0)].
\]
The shift \(-\alpha\,\mathbb{E}_{s_0\sim\mu_0}[\Phi(s_0)]\) does not depend on \(\pi\), so for any
two policies \(\pi_1,\pi_2\),
\[
J(\pi_1,\tilde r+\alpha F)-J(\pi_2,\tilde r+\alpha F)
=J(\pi_1,\tilde r)-J(\pi_2,\tilde r).
\]
Thus all rewards in \(\{\tilde r+\alpha F:|\alpha|\le\delta\}\) induce identical policy rankings,
i.e., they are PBRS-equivalent.
This completes the proof.
\end{proof}

\subsection{Proof of Theorem~\ref{theorem:core-eq}}
\label{appendix:theorem-core-eq}
\begin{proof}
Take any \(\tilde r \in \mathcal{R}\). By construction of the projection \(\text{Proj}\), we have \(
\text{Proj}(\tilde r) = \tilde r + F_\Phi\) for some shaping term \(F_\Phi \in \mathcal{S}_{\text{shape}}\). By Lemma~\ref{lemma:shaping}, we have
\begin{equation}
\label{eq:proj-equal}
J(\pi, \text{Proj}(\tilde r))
= J(\pi, \tilde r + F_\Phi)
= J(\pi, \tilde r) 
\qquad \forall\, \pi.
\end{equation}
So for each reward, its value is exactly the same as its projected version.

Let
\[
m := \min_{\tilde r\in\mathcal{U}} J(\pi,\tilde r),
\qquad
m_{\text{core}} := \min_{\tilde r_{\text{core}}\in\mathcal{U}_{\text{core}}}
J(\pi,\tilde r_{\text{core}}).
\]

Take any \(\tilde r \in \mathcal{U}\). From Equation~\ref{eq:proj-equal}, we have
\[
J(\pi,\tilde r) = J(\pi,\text{Proj}(\tilde r)) \ge m_{\text{core}},
\]
because \(\text{Proj}(\tilde r) \in \mathcal{U}_{\text{core}}\) by definition and
\(m_{\text{core}}\) is the minimum over that set.

Since this holds for every \(\tilde r \in \mathcal{U}\), we can take the minimum over
\(\tilde r\in\mathcal{U}\) on the left:
\begin{equation}
\label{eq:m_large_mcore}
m = \min_{\tilde r\in\mathcal{U}} J(\pi,\tilde r) \ge m_{\text{core}}.
\end{equation}
Now take any \(\tilde r_{\text{core}} \in \mathcal{U}_{\text{core}}\).
By definition of \(\mathcal{U}_{\text{core}}\), there exists some
\(\tilde r \in \mathcal{U}\) such that
\[
\tilde r_{\text{core}} = \text{Proj}(\tilde r).
\]

Using Lemma~\ref{lemma:shaping} again,
\[
J(\pi, \tilde r_{\text{core}}) = J(\pi, \tilde r).
\]

By definition of \(m\),
\[
m := \min_{\tilde r\in\mathcal{U}} J(\pi,\tilde r)
\;\le\; J(\pi,\tilde r) = J(\pi,\tilde r_{\text{core}}).
\]

Since this holds for every \(\tilde r_{\text{core}} \in \mathcal{U}_{\text{core}}\),
we can take the minimum over \(\tilde r_{\text{core}} \in \mathcal{U}_{\text{core}} \) on the right-hand side:
\begin{equation}
\label{eq:m_less_mcore}
m \;\le\; \min_{\tilde r_{\text{core}}\in\mathcal{U}_{\text{core}}} J(\pi,\tilde r_{\text{core}})
= m_{\text{core}}.
\end{equation}

From Equation~\ref{eq:m_large_mcore} and Equation~\ref{eq:m_less_mcore}, we have
\[
m \ge m_{\text{core}}
\quad\text{and}\quad
m \le m_{\text{core}}
\;\;\Rightarrow\;\;
m = m_{\text{core}}.
\]

Thus, for any policy \(\pi\),
\[
\min_{\tilde r\in\mathcal{U}} J(\pi,\tilde r)
=
\min_{\tilde r_{\text{core}}\in\mathcal{U}_{\text{core}}} J(\pi,\tilde r_{\text{core}}).
\]

Now take the max over policies:
\[
\max_{\pi} \min_{\tilde r\in\mathcal{U}} J(\pi,\tilde r)
=
\max_{\pi} \min_{\tilde r_{\text{core}}\in\mathcal{U}_{\text{core}}} J(\pi,\tilde r_{\text{core}}).
\]

Because the inner quantities are identical for each \(\pi\), the set of maximizers
\(\arg\max_{\pi}\) is also the same for the original and projected uncertainty
sets. This shows that the robust optimal value is unchanged when we pass
from \(\mathcal{U}\) to \(\mathcal{U}_{\text{core}}\).

\end{proof}

\subsection{Dimension Drop for Shaping-Aware Projected Uncertainty Set}
\label{appendix:dimension-drop}
In this section, we quantify how projection reduces the adversary’s effective search space. Since \(\mathcal{U}_{\text{core}}\) collapses all PBRS-equivalent rewards to a single representative, it removes entire directions in reward space that provably do not affect policy ranking. Under a linear parameterization, this reduction can be characterized explicitly as a decrease in affine dimension:
\begin{proposition}
\label{prop:dimension-shrink}
Assume a linear parameterization of the reward function: \(
r_\theta(s,a,s') = \phi(s,a,s')^\top \theta,\theta \in \Theta,
\) where \(\Theta \subseteq \mathbb{R}^d\) is a convex ambiguity set in parameter
space corresponding to $\mathcal{U}$ (i.e, $\mathcal{U}=\{r_\theta:\theta \in \Theta\}$), and let \(\mathcal{S}_{\text{shape}} \subseteq \mathbb{R}^d\) be the shaping subspace. Let $\operatorname{aff}(\Theta)$ denote the affine hull of $\Theta$.
Then the projected set can be carried out over the
lower-dimensional set
\(\Theta_{\mathrm{core}} =\{\text{Proj}(\theta):\theta\in \Theta\}\),
whose affine dimension is at most \(
\dim(\Theta_{\mathrm{core}})
\;\le\;
\dim(\Theta) - \dim\bigl(\mathcal{S}_{\text{shape}} \cap \operatorname{aff}(\Theta)\bigr).
\) In particular, whenever \(\mathcal{S}_{\text{shape}}\) is nontrivial and intersects the span of
\(\Theta\), the projected search space strictly shrinks.
\end{proposition}

\begin{proof}
Let \(A := \operatorname{aff}(\Theta)\) be the affine hull of \(\Theta\), and let
\(
k := \dim(A) = \dim(\Theta),
\) the affine dimension of \(\Theta\). Pick any fixed \(\theta_0 \in \Theta\), and define the associated linear
subspace
\[
V := A - \theta_0
   = \{ \theta - \theta_0 : \theta \in A \}
   \subseteq \mathbb{R}^d.
\]
Then \(V\) is a \(k\)-dimensional linear subspace, i.e. \(
\dim(V) = k.
\) Similarly, define the intersection of the shaping subspace with this affine
subspace: \(
S' := \mathcal{S}_{\text{shape}} \cap A.
\)
Then the corresponding linear subspace inside \(V\) is
\[
S := S' - \theta_0
   = \mathcal{S}_{\text{shape}} \cap V.
\]
where \(\dim(S) = \dim(\mathcal{S}_{\text{shape}} \cap A).\) Define a shifted version of \(\mathrm{Proj}\) on \(V\):
\[
T : V \to \mathbb{R}^d, \qquad
T(v) := \mathrm{Proj}(\theta_0 + v) - \mathrm{Proj}(\theta_0).
\]
Intuitively, \(T\) describes how \(\mathrm{Proj}\) moves points relative to the
base point \(\theta_0\). Observe that, for any \(v \in V\) and any \(u \in S\) (so \(u \in \mathcal{S}_{\text{shape}} \cap V\)), we have \(\theta_0 + v + u \in A\). This is because $u\in V$ by definition of $u$ and $V$ is a linear subspace, so $v+u\in V$. Thus, we have $\theta_0+(v+u)\in \theta_0+V=A$ by definition of $V$. Next, by shaping invariance, we have \(
\mathrm{Proj}(\theta_0 + v + u) = \mathrm{Proj}(\theta_0 + v).\) Therefore,
\[
T(v+u)
= \mathrm{Proj}(\theta_0 + v + u) - \mathrm{Proj}(\theta_0)
= \mathrm{Proj}(\theta_0 + v) - \mathrm{Proj}(\theta_0)
= T(v).
\]

So \(T\) is constant along directions in \(S\).
That means points that differ by something in $S$ are indistinguishable to \(T\).
In other words, \(T\) factors through the quotient space \(V/S\). Let $C: V\to V/S, C(v):=[v]$ be the canonical quotient map sending each $v\in V$ to its equivalence class $[v]$. Because \(T(v+u) = T(v)\) for all \(u \in S\), if two vectors are
equivalent (\(v' \sim v\)), then they get mapped to the same output: \(
v' = v + u,\; u \in S
      \;\;\Longrightarrow\;\;
T(v') = T(v).\) So \(T\) only depends on the equivalence class \([v]\), not the specific representative. We define:
\[
\tilde T : V/S \to \mathbb{R}^d, \qquad
\tilde T([v]) := T(v).
\]
And we can write:
\[
T = \tilde T \circ C,
\]
because for any \(v \in V\),
\[
(\tilde T \circ C)(v) = \tilde T([v]) = T(v).
\]
By construction, \(\tilde T\) is linear, and its image is the linear span of
\(T(V)\):
\[
\operatorname{Im}(\tilde T) = \operatorname{span}(T(V)).
\]

The dimension of the quotient space is
\[
\dim(V/S) = \dim(V) - \dim(S) = k - \dim(\mathcal{S}_{\text{shape}} \cap A).
\]

Since \(\tilde T\) is a linear map from \(V/S\) to \(\mathbb{R}^d\), its rank is
at most the dimension of its domain:
\[
\operatorname{rank}(\tilde T)
\;\le\;
\dim(V/S)
= k - \dim(\mathcal{S}_{\text{shape}} \cap A).
\]

And notice that \(\operatorname{rank}(\tilde T) = \dim(\operatorname{Im}(\tilde T))
= \dim(\operatorname{span}(T(V)))\).
By definition, \(
\Theta_{\text{core}} = \{\mathrm{Proj}(\theta) : \theta \in \Theta\}.
\) Any \(\theta \in \Theta\) can be written as \(\theta = \theta_0 + v\) with
\(v \in V\). Then
\[
\mathrm{Proj}(\theta)
= \mathrm{Proj}(\theta_0 + v)
= \mathrm{Proj}(\theta_0) + T(v).
\]

Therefore
\[
\Theta_{\text{core}} \subseteq \mathrm{Proj}(\theta_0) + T(V),
\]
and consequently,
\[
\operatorname{aff}(\Theta_{\text{core}})
\subseteq
\mathrm{Proj}(\theta_0) + \operatorname{span}(T(V)).
\]

Thus the affine dimension of \(\Theta_{\text{core}}\) satisfies
\[
\dim(\Theta_{\text{core}})
= \dim\bigl(\operatorname{aff}(\Theta_{\text{core}})\bigr)
\le
\dim\bigl(\operatorname{span}(T(V))\bigr)
= \operatorname{rank}(\tilde T)
\le
k - \dim(\mathcal{S}_{\text{shape}} \cap A).
\]

Recall \(k = \dim(A) = \dim(\Theta)\), so we obtain: 
\[
\dim(\Theta_{\mathrm{core}})
\;\le\;
\dim(\Theta) - \dim\bigl(\mathcal{S}_{\text{shape}} \cap \operatorname{aff}(\Theta)\bigr).
\]
which completes the proof. 
If \(\mathcal{S}_{\text{shape}} \cap \operatorname{aff}(\Theta)\) is nontrivial
(i.e., has positive dimension), then
\[
\dim\bigl(\mathcal{S}_{\text{shape}} \cap \operatorname{aff}(\Theta)\bigr) > 0,
\]
so
\[
\dim(\Theta_{\text{core}})
\;\le\;
\dim(\Theta)
-
\underbrace{\dim\bigl(\mathcal{S}_{\text{shape}} \cap \operatorname{aff}(\Theta)\bigr)}_{>0}
\;<\;
\dim(\Theta),
\]
i.e., the affine dimension of the projected ambiguity set is
strictly smaller than that of the original ambiguity set.
\end{proof}

In general, replacing \(\mathcal{U}\) with \(\mathcal{U}_{\text{core}}\) reduces the adversary's search space, and the extent of this reduction scales with the number of independent shaping directions present in \(\mathcal{U}\). 
The toy example in Figure~\ref{fig:toy-mdp-dim} is a concrete illustration: here \(\dim(\mathcal{U})=2\) and the shaping subspace \(\mathcal{S}_{\text{shape}}=\mathrm{span}\{(1,1)\}\) has dimension \(1\), so \(\mathcal{U}_{\text{core}}\) is one-dimensional. At the other extreme, if \(\mathcal{U}\) consists only of shaping rewards, then \(\mathcal{U}_{\text{core}}\) collapses to a single representative and removes all dimensions. We further show that for any discounted multi-armed bandit, the projection removes exactly one degree of freedom. 

\begin{proposition}
Consider a discounted multi-armed bandit with a single state \(s\) and action set \(\mathcal{A}\).
Let \(\mathcal{R}=\mathbb{R}^{\mathcal{A}}\) denote the reward space, and let \(\mathcal{S}_{\mathrm{shape}}\) be the shaping subspace. Then the quotient space
\(\mathcal{R}_{\mathrm{core}}=\mathcal{R}/\mathcal{S}_{\mathrm{shape}}\) has dimension \(|\mathcal{A}|-1\).
Equivalently, projecting to canonical representatives removes exactly one behaviorally redundant direction.
\end{proposition}
\begin{proof}
PBRS shaping is
\[
F(s,a,s') \;=\; \gamma \Phi(s') - \Phi(s).
\]
For multi-armed bandit MPD, there is only one state, so $s' = s$, hence the PBRS shaping term
\[
F(s,a,s) \;=\; \gamma \Phi(s) - \Phi(s)
           \;=\; (\gamma - 1)\Phi(s)
           \;=:\; F
\]
is a constant independent of the action. Thus, shaping can only add the same constant $F$ to every reward. Therefore, two reward vectors $r, r' \in \mathbb{R}^{|\mathcal{A}|}$ are PBRS-equivalent iff
\[
        r' \;=\; r + F \cdot \mathbf{1},
\]
i.e., they differ by the all-ones direction. Then the shaping
subspace is
\[
    \mathcal{S}_{\text{shape}}
    \;=\;
    \{ F\cdot  \mathbf{1} : F \in \mathbb{R} \}
    \;=\;
    \operatorname{span}\{\mathbf{1}\},
\]
which has dimension $1$. Therefore, the quotient space has dimension
\[
    \dim\bigl(\mathcal{R} / \mathcal{S}_{\text{shape}}\bigr)
    \;=\;
    |\mathcal{A}| - 1.
\]   
This completes the proof.
\end{proof}

\subsection{Additional Properties of Shaping-aware Projected Uncertainty Set}
\label{appendix:property-proj}

In this section, we propose additional properties of the projected uncertainty set.

\begin{proposition}
    Let \(\mathcal{U} \subseteq \mathcal{R}\) be a compact uncertainty set and  \( \mathcal{U}_{\text{core}} := \mathrm{Proj}(\mathcal{U})=\{\mathrm{Proj}(r): r\in \mathcal{U}\}\) is the corresponding projected uncertainty set. If  \(\mathrm{Proj}:\mathcal{R}\to\mathcal{R}\) is continuous, then \(\mathcal{U}_{\text{core}}\) is compact.
\end{proposition}
\begin{proof}
Since \(\mathrm{Proj}\) is continuous, the continuous image of a compact set is compact. Hence \(\mathcal{U}_{\text{core}}\) is compact.
\end{proof}

\begin{proposition}
Let \(\mathcal{U} \subseteq \mathcal{R}\) be a convex uncertainty set and  \( \mathcal{U}_{\text{core}} := \mathrm{Proj}(\mathcal{U})=\{\mathrm{Proj}(r): r\in \mathcal{U}\}\) is the corresponding projected uncertainty set. If  \(\mathrm{Proj}:\mathcal{R}\to\mathcal{R}\) is affine, i.e., for all \(r_1,r_2\in\mathcal{R}\) and \(\lambda\in[0,1]\),
\[
\mathrm{Proj}((1-\lambda)r_1+\lambda r_2)
=(1-\lambda)\mathrm{Proj}(r_1)+\lambda\,\mathrm{Proj}(r_2),
\] then \(\mathcal{U}_{\text{core}}\) is convex.
\end{proposition}
\begin{proof}
Take any \(u'_1,u'_2\in \mathcal{U}_{\text{core}}\). Then there exist \(u_1,u_2\in \mathcal{U}\) such that \(u'_i=\mathrm{Proj}(u_i)\) for \(i\in\{1,2\}\).
For any \(\lambda\in[0,1]\), since \(\mathcal{U}\) is convex, we have \((1-\lambda)u_1+\lambda u_2\in \mathcal{U}\). If \(\mathrm{Proj}\) is affine, then we have
\[
(1-\lambda)u'_1+\lambda u'_2
=(1-\lambda)\mathrm{Proj}(u_1)+\lambda\mathrm{Proj}(u_2)
=\mathrm{Proj}((1-\lambda)u_1+\lambda u_2)\in \mathcal{U}_{\text{core}}.
\]
Thus \(\mathcal{U}_{\text{core}}\) is convex.
\end{proof}

\subsection{Proof of Theorem~\ref{theorem:core-regular-eq-robust}}
\label{appendix:core-regular-eq-robust}
To prove Theorem~\ref{theorem:core-regular-eq-robust}, we first recall a standard connection between regularization and robustness with penalty. The statement below is adapted from Theorem 1 of~\cite{husain2021regularized} to match our notation. 

\begin{theorem}
\label{theorem:regular-eq-penalty}
Let \(\Omega\) be proper, closed, and strongly convex,
and let \(\Omega^*\) denote its convex conjugate. Then, for any policy \(\pi\),
\[
J(\pi,r) - \Omega(\pi)
=
\min_{\tilde r \in \mathbb{R}^{\mathcal{X'}}}
\Bigl\{
  J(\pi,\tilde r) + \Omega^*(r-\tilde r)
\Bigr\}.
\]
\end{theorem}
\begin{proof}{(Sketch)} The key part of the proof is to rewrite \(\Omega\) in terms of the convex conjugate of \(-\Omega\), which is well-defined since we assume \(\Omega\) to be strongly convex. The detailed proof can be found in~\cite{husain2021regularized}.
\end{proof}

We now turn to budgeted uncertainty sets. For \(\epsilon\ge 0\) and a regularization weight \(\lambda>0\), define
\[
\mathcal{U}
:=\Bigl\{\tilde r \in \mathbb{R}^{\mathcal{X'}}:\;
\Omega^*\!\left(\frac{r-\tilde r}{\lambda}\right)\le \epsilon
\Bigr\}.
\]

\begin{theorem}
\label{theorem:budget-duality}
Fix any policy \(\pi\), \(\epsilon \ge 0\) and \(\lambda > 0\), then
the following lower bound holds:
\[
\min_{\tilde r\in\mathcal U}J(\pi,\tilde r)
\ge
J(\pi,r)-\lambda\Omega(\pi)-\lambda\epsilon.
\] 
\end{theorem}

\begin{proof}
Consider the inner problem
\[
\min_{\tilde r\in\mathcal U}J(\pi,\tilde r).
\]
For every feasible $\tilde r\in\mathcal U$, we have
\[
\Omega^*\left(\frac{r-\tilde r}{\lambda}\right)\le \epsilon.
\]
Therefore,
\[
\lambda\Omega^*\left(\frac{r-\tilde r}{\lambda}\right)-\lambda\epsilon
\le 0.
\]
Hence, for every $\tilde r\in\mathcal U$,
\[
J(\pi,\tilde r)
\ge
J(\pi,\tilde r)
+
\lambda\Omega^*\left(\frac{r-\tilde r}{\lambda}\right)
-
\lambda\epsilon.
\]
Taking the minimum over $\tilde r\in\mathcal U$ gives
\[
\min_{\tilde r\in\mathcal U}J(\pi,\tilde r)
\ge
\min_{\tilde r\in\mathcal U}
\left\{
J(\pi,\tilde r)
+
\lambda\Omega^*\left(\frac{r-\tilde r}{\lambda}\right)
-
\lambda\epsilon
\right\}.
\]
Since minimizing over the feasible set is no smaller than minimizing over the
entire reward space, we further have
\[
\min_{\tilde r\in\mathcal U}
\left\{
J(\pi,\tilde r)
+
\lambda\Omega^*\left(\frac{r-\tilde r}{\lambda}\right)
-
\lambda\epsilon
\right\}
\ge
\min_{\tilde r\in\mathbb R^{\mathcal X'}}
\left\{
J(\pi,\tilde r)
+
\lambda\Omega^*\left(\frac{r-\tilde r}{\lambda}\right)
-
\lambda\epsilon
\right\}.
\]
By the scaling property of convex conjugates~\cite{hiriart2004fundamentals,mensch2018differentiable,bertsekas2009convex}, for any $\lambda>0$, the convex
conjugate of the scaled regularizer $\lambda\Omega$ is
\[
(\lambda\Omega)^*(y)
=
\lambda\Omega^*\left(\frac{y}{\lambda}\right).
\]
Applying Theorem~\ref{theorem:regular-eq-penalty} to the scaled regularizer
$\lambda\Omega$ gives
\[
J(\pi,r)-\lambda\Omega(\pi)
=
\min_{\tilde r\in\mathbb R^{\mathcal X'}}
\left\{
J(\pi,\tilde r)
+
(\lambda\Omega)^*(r-\tilde r)
\right\}.
\]
Using the scaling property, this becomes
\[
J(\pi,r)-\lambda\Omega(\pi)
=
\min_{\tilde r\in\mathbb R^{\mathcal X'}}
\left\{
J(\pi,\tilde r)
+
\lambda\Omega^*\left(\frac{r-\tilde r}{\lambda}\right)
\right\}.
\]
Therefore,
\[
\begin{aligned}
&\min_{\tilde r\in\mathbb R^{\mathcal X'}}
\left\{
J(\pi,\tilde r)
+
\lambda\Omega^*\left(\frac{r-\tilde r}{\lambda}\right)
-
\lambda\epsilon
\right\} =
J(\pi,r)-\lambda\Omega(\pi)-\lambda\epsilon.
\end{aligned}
\]
Combining this with the previous inequalities yields
\[
\min_{\tilde r\in\mathcal U}J(\pi,\tilde r)
\ge
J(\pi,r)-\lambda\Omega(\pi)-\lambda\epsilon.
\]
This completes the proof.

\end{proof}

Theorem~\ref {theorem:core-regular-eq-robust} holds because the inequality above is agnostic to the particular reward domain. It only requires that the inner problem optimizes over an uncertainty set defined through \(\Omega^*\) and that Slater’s condition holds. In our setting, the projection \(\mathrm{Proj}\) maps each reward to a PBRS-equivalent canonical representative and satisfies \(\mathrm{Proj}(r+F_\Phi)=\mathrm{Proj}(r)\). Therefore, replacing the original uncertainty set by its projected counterpart \(\mathcal{U}_{\mathrm{core}}=\mathrm{Proj}(\mathcal{U})\) simply restricts the adversary to one representative per PBRS class without altering policy rankings. The same duality applies verbatim on \(\mathcal{U}_{\mathrm{core}}\). Concretely, fix any policy $\pi$, Theorem~\ref{theorem:core-eq} gives
\[
\min_{\tilde r_{\mathrm{core}}\in\mathcal U_{\mathrm{core}}}
J(\pi,\tilde r_{\mathrm{core}})
=
\min_{\tilde r\in\mathcal U}
J(\pi,\tilde r).
\]
Moreover, Lemma~\ref{lemma:shaping} shows
\[
J(\pi,r_{\mathrm{core}})=J(\pi,r).
\]
Applying Theorem~\ref{theorem:budget-duality} to the right-hand side yields
\[
\min_{\tilde r_{\mathrm{core}}\in\mathcal U_{\mathrm{core}}}
J(\pi,\tilde r_{\mathrm{core}})=\min_{\tilde r\in\mathcal U}
J(\pi,\tilde r)
\ge
J(\pi,r)-\lambda\Omega(\pi)-\lambda\epsilon.
\]
Using $J(\pi,r)=J(\pi,r_{\mathrm{core}})$, we obtain
\[
\min_{\tilde r_{\mathrm{core}}\in\mathcal U_{\mathrm{core}}}
J(\pi,\tilde r_{\mathrm{core}})
\ge
J(\pi,r_{\mathrm{core}})-\lambda\Omega(\pi)-\lambda\epsilon.
\]
Finally, taking the maximum over $\pi$ gives
\[
\max_{\pi}
\min_{\tilde r_{\mathrm{core}}\in\mathcal U_{\mathrm{core}}}
J(\pi,\tilde r_{\mathrm{core}})
\ge
\max_{\pi}
\left[
J(\pi,r_{\mathrm{core}})-\lambda\Omega(\pi)
\right]
-\lambda\epsilon.
\]
This completes the proof.

\subsection{Proof of Theorem~\ref{theorem:entropy-kl-exact-duality}}
\label{appendix:entropy-kl-duality-proof}

\begin{proof}
Following a similar idea used in the proof of
Theorem~\ref{theorem:core-regular-eq-robust}, it suffices to prove the result for the unprojected uncertainty set and then replace $\mathcal U$ by $\mathcal{U_{\mathrm{core}}}$ and $r$ by
$r_{\mathrm{core}}$.

Fix a policy $\pi$. Let
\[
u=\frac{r-\tilde r}{\lambda}.
\]
Then $\tilde r=r-\lambda u$, and the constraint
$\Omega^*((r-\tilde r)/\lambda)\le \epsilon/\lambda$ becomes
\[
\Omega^*(u)\le \frac{\epsilon}{\lambda}.
\]
Therefore,
\[
\begin{aligned}
\min_{\tilde r\in\mathcal U}J(\pi,\tilde r)
&=
\min_{u:\Omega^*(u)\le \epsilon/\lambda}
J(\pi,r-\lambda u)\\
&=
J(\pi,r)
-
\lambda
\sup_{u:\Omega^*(u)\le \epsilon/\lambda}
\mathbb E\!\left[
\sum_t\gamma^t u(s_t,a_t)
\Bigm|\mu_0,\pi,P
\right].
\end{aligned}
\]
Define
\[
d_\pi(s,a)
:=
\mathbb E\!\left[
\sum_{t=0}^{\infty}\gamma^t
\mathbf 1\{s_t=s,a_t=a\}
\Bigm|\mu_0,\pi,P
\right],
\qquad
d_\pi(s):=\sum_a d_\pi(s,a).
\]
Then
\[
d_\pi(s,a)=d_\pi(s)\pi(a|s),
\]
and the objective in the inner maximization can be written as
\[
\mathbb E\!\left[
\sum_t\gamma^t u(s_t,a_t)
\Bigm|\mu_0,\pi,P
\right]
=
\sum_s\sum_a d_\pi(s,a)u(s,a).
\]
For the entropy-based uncertainty set, 
\[
\Omega^*(u)= \mathbb{E}\left[
            \sum_{t} \gamma^{t}
            \log \sum_{a }
\exp(u(s_t,a_t))\Big| \mu_0, \pi, P \right],\]
Equivalently, 
\[
\Omega^*(u)
=
\sum_s d_\pi(s)\log\sum_a \exp(u(s,a)).
\]
Therefore, the constrained maximization is
\[
\sup_{u:\Omega^*(u)\le \epsilon/\lambda}
\sum_s\sum_a d_\pi(s,a)u(s,a).
\]
Equivalently, we minimize the negative objective:
\[
\min_u
-\sum_s\sum_a d_\pi(s,a)u(s,a)
\quad
\text{s.t.}
\quad
\Omega^*(u)\le \frac{\epsilon}{\lambda}.
\]
The Lagrangian is
\[
\mathcal L(u,\eta)
=
-\sum_s\sum_a d_\pi(s,a)u(s,a)
+
\eta\left(
\sum_s d_\pi(s)\log\sum_a \exp(u(s,a))
-
\frac{\epsilon}{\lambda}
\right).
\]
The KKT conditions are
\[
\eta\ge 0,
\]
\[
\sum_s d_\pi(s)\log\sum_a \exp(u(s,a))
\le \frac{\epsilon}{\lambda},
\]
\[
\eta\left(
\sum_s d_\pi(s)\log\sum_a \exp(u(s,a))
-
\frac{\epsilon}{\lambda}
\right)=0,
\]
and, for every \((s,a)\),
\[
\frac{\partial \mathcal L}{\partial u(s,a)}
=
-d_\pi(s,a)
+
\eta d_\pi(s)
\frac{\exp(u(s,a))}{\sum_{a'}\exp(u(s,a'))}
=0.
\]
Using \(d_\pi(s,a)=d_\pi(s)\pi(a|s)\), this becomes
\[
-d_\pi(s)\pi(a|s)
+
\eta d_\pi(s)
\frac{\exp(u(s,a))}{\sum_{a'}\exp(u(s,a'))}
=0.
\]
For every visited state with \(d_\pi(s)>0\), dividing by \(d_\pi(s)\) gives
\[
\pi(a|s)
=
\eta
\frac{\exp(u(s,a))}{\sum_{a'}\exp(u(s,a'))}.
\]
Summing both sides over \(a\) yields \(\eta=1\). Hence
\[
\pi(a|s)
=
\frac{\exp(u(s,a))}{\sum_{a'}\exp(u(s,a'))}.
\]
Equivalently,
\[
u(s,a)=\log\pi(a|s)+c(s),
\]
where \(c(s)\) is independent of \(a\). We next show that the constraint must be active at the optimum. Suppose, for
contradiction, that an optimal solution \(u\) satisfies
\[
\Omega^*(u)<\frac{\epsilon}{\lambda}.
\]
For any scalar \(c>0\), define
\[
u_c(s,a)=u(s,a)+c .
\]
Then
\[
\log\sum_a \exp(u_c(s,a))
=
\log\sum_a \exp(u(s,a)+c)
=
c+\log\sum_a \exp(u(s,a)).
\]
Therefore,
\[
\Omega^*(u_c)
=
\Omega^*(u)
+
c\sum_s d_\pi(s).
\]
For sufficiently small \(c>0\), we still have
\[
\Omega^*(u_c)\le \frac{\epsilon}{\lambda}.
\]
However, the objective strictly increases:
\[
\sum_s\sum_a d_\pi(s,a)u_c(s,a)
=
\sum_s\sum_a d_\pi(s,a)u(s,a)
+
c\sum_s\sum_a d_\pi(s,a)
>
\sum_s\sum_a d_\pi(s,a)u(s,a).
\]
This contradicts the optimality of \(u\). Hence the constraint must be active:
\[
\Omega^*(u)=\frac{\epsilon}{\lambda}.
\]
Substituting \(u(s,a)=\log\pi(a|s)+c(s)\) into \(\Omega^*(u)\), we obtain
\[
\sum_s d_\pi(s)\log\sum_a \exp(u(s,a))
=
\sum_s d_\pi(s)\log\sum_a \exp(\log\pi(a|s)+c(s))
=
\sum_s d_\pi(s)c(s),
\]
because \(\sum_a\pi(a|s)=1\). Hence
\[
\sum_s d_\pi(s)c(s)
=
\frac{\epsilon}{\lambda}.
\]
Therefore,
\[
\begin{aligned}
&\sup_{u:\Omega^*(u)\le \epsilon/\lambda}
\mathbb E\!\left[
\sum_t\gamma^t u(s_t,a_t)
\Bigm|\mu_0,\pi,P
\right] \\
&=
\sum_s\sum_a d_\pi(s,a)
\bigl(\log\pi(a|s)+c(s)\bigr)\\
&=
\sum_s d_\pi(s)\sum_a\pi(a|s)\log\pi(a|s)
+
\sum_s d_\pi(s)c(s)\\
&=
-\Omega_{\mathrm{Ent}}(\pi)+\frac{\epsilon}{\lambda}.
\end{aligned}
\]
For the KL-based uncertainty set, write
\[
\Omega^*(u)
=
\sum_s d_\pi(s)
\log\sum_a \pi_{\mathrm{ref}}(a|s)\exp(u(s,a)).
\]
The Lagrangian is
\[
\mathcal L(u,\eta)
=
-\sum_s\sum_a d_\pi(s,a)u(s,a)
+
\eta\left(
\sum_s d_\pi(s)
\log\sum_a \pi_{\mathrm{ref}}(a|s)\exp(u(s,a))
-
\frac{\epsilon}{\lambda}
\right).
\]
The stationarity condition is
\[
-d_\pi(s,a)
+
\eta d_\pi(s)
\frac{\pi_{\mathrm{ref}}(a|s)\exp(u(s,a))}
{\sum_{a'}\pi_{\mathrm{ref}}(a'|s)\exp(u(s,a'))}
=0.
\]
Using \(d_\pi(s,a)=d_\pi(s)\pi(a|s)\) and summing over \(a\) again gives
\(\eta=1\), so
\[
\pi(a|s)
=
\frac{\pi_{\mathrm{ref}}(a|s)\exp(u(s,a))}
{\sum_{a'}\pi_{\mathrm{ref}}(a'|s)\exp(u(s,a'))}.
\]
Equivalently,
\[
u(s,a)
=
\log\frac{\pi(a|s)}{\pi_{\mathrm{ref}}(a|s)}
+
c(s).
\]
Since the constraint is active,
\[
\sum_s d_\pi(s)c(s)=\frac{\epsilon}{\lambda}.
\]
Thus
\[
\begin{aligned}
&\sup_{u:\Omega^*(u)\le \epsilon/\lambda}
\mathbb E\!\left[
\sum_t\gamma^t u(s_t,a_t)
\Bigm|\mu_0,\pi,P
\right]\\
&=
\sum_s d_\pi(s)
\sum_a\pi(a|s)
\log\frac{\pi(a|s)}{\pi_{\mathrm{ref}}(a|s)}
+
\frac{\epsilon}{\lambda}\\
&=
-\Omega_{\mathrm{KL}}(\pi)+\frac{\epsilon}{\lambda}.
\end{aligned}
\]

Substituting either identity into
\[
\min_{\tilde r\in\mathcal U}J(\pi,\tilde r)
=
J(\pi,r)
-
\lambda
\sup_{u:\Omega^*(u)\le \epsilon/\lambda}
\mathbb E\!\left[
\sum_t\gamma^t u(s_t,a_t)
\Bigm|\mu_0,\pi,P
\right]
\]
gives
\[
\min_{\tilde r\in\mathcal U}J(\pi,\tilde r)
=
J(\pi,r)+\lambda\Omega(\pi)-\epsilon,
\]
where \(\Omega=\Omega_{\mathrm{Ent}}\) in the entropy case and
\(\Omega=\Omega_{\mathrm{KL}}\) in the KL case. This completes the proof.
\end{proof}

\subsection{Properties of \(\Omega^*\)-induced Projection}
\label{appendix:property-projection-omega}
Recall that we estimate the projection using
\[
\operatorname{Proj}(r)=\arg\min_{\Phi}
\Omega^*\Bigl(
  r(s, a, s') - \bigl(\gamma \Phi(s') - \Phi(s)\bigr)
\Bigr).
\]
Next, we establish conditions under which the projection is well-defined, i.e., the minimizer exists and is unique. We begin with a technical lemma that will be used in the proof.
\begin{lemma}
\label{lemma:sshape-linear} 
    $\mathcal{S}_{\text{shape}}$ is a linear subspce.
\end{lemma}

\begin{proof}
Recall that PBRS shaping terms are of the form
\[
F_{\Phi}(s,a,s') \;=\; \gamma \Phi(s') - \Phi(s).
\]
Then we define the shaping set as
\[
S_{\text{shape}} := \{ F_{\Phi} : \Phi \in \mathbb{R}^{\mathcal{S}}\}.
\]

Now take any \(\Phi_1, \Phi_2\) and scalars \(a,b\). Because the map
\(\Phi \mapsto F_{\Phi}\) is linear,
\[
F_{a\Phi_1 + b\Phi_2}(s,a,s')
= \gamma (a\Phi_1 + b\Phi_2)(s') - (a\Phi_1 + b\Phi_2)(s)
= a F_{\Phi_1}(s,a,s') + b F_{\Phi_2}(s,a,s').
\]

So:
\begin{itemize}
    \item \(F_{\Phi_1} + F_{\Phi_2} = F_{\Phi_1 + \Phi_2} \in S_{\text{shape}}\),
    \item \(a F_{\Phi_1} = F_{a\Phi_1} \in S_{\text{shape}}\).
\end{itemize}

This completes the proof.

\end{proof}

\begin{theorem}
\label{thm:proj-existence-uniqueness}
Assume $\mathcal S$ and $\mathcal X'$ are finite and $\Omega^*$
is proper, lower semicontinuous, and convex. Let $[r]:=\{r+F_{\Phi}:\Phi\in\mathbb R^{\mathcal S}\}$ denote the PBRS equivalence class of $r$.
Assume that $\Omega^*$ is level-bounded on $[r]$, i.e., for every $c\in\mathbb R$ the set
\[
\{\,r'\in [r]: \Omega^*(r')\le c\,\}
\quad \text{is bounded.}
\]
Then the minimization $\min_{r'\in[r]} \Omega^*(r')$ exists, equivalently,
$\operatorname{Proj}(r)$ is nonempty.

If additionally $\Omega^*$ is strictly convex, then the projected reward $\operatorname{Proj}(r)$ is unique.
\end{theorem}

\begin{proof}
Since $\Omega^*$ is proper, lower semicontinuous, and convex, it is a closed convex function on the
finite-dimensional space $\mathbb R^{\mathcal X'}$. The set $[r]=r+\mathcal S_{\mathrm{shape}}$ is a nonempty affine set (it contains $r$) and is closed and convex
because $\mathcal S_{\mathrm{shape}}$ is a linear subspace as proved in Lemma~\ref{lemma:sshape-linear}.
By the assumed level-boundedness on $[r]$, each sublevel set
\[
L_c := \{\,r'\in[r]: \Omega^*(r')\le c\,\}
\]
is bounded. Since $\Omega^*$ is lower semicontinuous and $[r]$ is closed, $L_c$ is also closed, hence compact
in finite dimensions. Therefore, the infimum of $\Omega^*$ over $[r]$ is attained on $L_c$ for any $c$ above
the infimum, and $\arg\min_{r'\in[r]}\Omega^*(r')$ is nonempty.

If $\Omega^*$ is strictly convex, then its restriction to the convex set $[r]$ is strictly convex. Hence,
there can be at most one minimizer on $[r]$. Therefore $\operatorname{Proj}(r)$ is unique.
\end{proof}

Beyond the conceptual motivation in the main paper, namely that the \(\Omega^*\)-aligned projection yields a consistent and conservative canonicalization, the projection can also improve inner optimization of the reward-robust RL problem from a max-min perspective. A growing line of work solves the inner minimization using dimension-free convex optimization methods such as mirror descent~\cite{zhan2023policy,bossens2025mirror,tomar2020mirror}. For these methods, iteration complexity typically depends on the size of the feasible set. In our setting, the projection selects, within each PBRS equivalence class, the representative that is smallest under the
$\Omega^*$-geometry. As a consequence, projection cannot increase the $\Omega^*$-radius of the uncertainty set, which directly tightens the corresponding mirror-descent convergence bound. We first prove that the projection cannot increase the $\Omega^*$-radius of the uncertainty set.

\begin{proposition}
\label{prop:diameter-shrink}
Assume $\Omega^*:\mathcal R\to\mathbb R$ is convex and bounded below.
Define $\mathrm{Rad}_{\Omega^*}(\mathcal U):=\sup_{r\in\mathcal U}\ \Omega^*(r)$.
Then for any uncertainty set $\mathcal U\subseteq\mathcal R$,
\[
\mathrm{Rad}_{\Omega^*}(\mathrm{Proj}(\mathcal U))
\;\le\;
\mathrm{Rad}_{\Omega^*}(\mathcal U).
\]
\end{proposition}

\begin{proof}
Fix any $r\in\mathcal R$. By definition,
\[
\mathrm{Proj}(r)\in\arg\min_{\Phi}\ \Omega^*\!\bigl(r-(\gamma\Phi(s')-\Phi(s))\bigr).
\]
Since $\Phi\equiv 0$ is feasible, we have
$\Omega^*(\mathrm{Proj}(r))\le \Omega^*(r)$.
Taking the supremum over $r\in\mathcal U$ yields
\[
\sup_{\tilde r\in \mathrm{Proj}(\mathcal U)}\Omega^*(\tilde r)
=
\sup_{r\in\mathcal U}\Omega^*(\mathrm{Proj}(r))
\le
\sup_{r\in\mathcal U}\Omega^*(r).
\]
This completes the proof.
\end{proof}

Since standard mirror-descent bounds scale with a geometry-dependent size term of the feasible set
(e.g., a Bregman radius under the mirror map), reducing $\mathrm{Rad}_{\Omega^*}$
tightens the worst-case iteration complexity constants for the inner minimization. To make this dependence explicit, consider minimizing a convex function $f$ over a convex set
$\mathcal K$ using mirror descent with mirror map $\Psi$ that is $\rho$-strongly convex with
respect to a norm $\|\cdot\|$. Assume the subgradients are bounded as
$\|g_t\|_* \le L$ for all iterations. A standard guarantee (e.g., Theorem~4.3 of~\cite{bubeck2015convex})
states that with a suitable step size,
\[
f(\bar x_t)-f(x^\star)\ \le\ L\sqrt{\frac{2D}{\rho\,t}},
\qquad
D:=\sup_{x\in\mathcal K} D_{\Psi}(x,x_1),
\]
where $D_{\Psi}$ is the Bregman radius induced by $\Psi$ and $x_1$ is the initialization.
Equivalently, to reach accuracy $f(\bar x_t)-f(x^\star)\le \delta$, it suffices that
\[
t\ \ge\ \frac{2L^2}{\rho\,\delta^2}\,D .
\]

In our setting, we take $\Psi=\Omega^*$ (we additionally assume \(\Omega^*\) is $\rho$-strongly convex) and initialize at $x_1=0$.\footnote{This holds after an affine normalization: replace $\Omega^*(x)$ by
$\widetilde{\Omega}^*(x) := \Omega^*(x) - \Omega^*(0) - \langle \nabla
\Omega^*(0), x \rangle$, which does not change the induced Bregman
divergence up to this choice of origin.}
Then the domain-size term becomes
\[
D \;=\; \sup_{x\in\mathcal K} D_{\Omega^*}(x,0)
\;=\; \sup_{x\in\mathcal K} \Omega^*(x)
\;=:\; \mathrm{Rad}_{\Omega^*}(\mathcal K).
\]
Therefore, the mirror-descent iteration complexity scales linearly with
$\mathrm{Rad}_{\Omega^*}(\mathcal K)$:
\[
t(\delta)\ \ge\ \frac{2L^2}{\rho\,\delta^2}\,\mathrm{Rad}_{\Omega^*}(\mathcal K).
\]
As proved in Proposition~\ref{prop:diameter-shrink}, for any feasible set $\mathcal K \subseteq \mathcal R$ ,
projection cannot increase the $\Omega^*$-radius:
\[
\mathrm{Rad}_{\Omega^*}(\operatorname{Proj}(\mathcal K))
\;\le\;
\mathrm{Rad}_{\Omega^*}(\mathcal K).
\]
Combining with the bound above, optimizing the inner problem over the projected set tightens the
worst-case mirror-descent complexity guarantee, implying that fewer inner-loop iterations suffice to reach a target accuracy.

Since the \(\Omega^*\)-induced geometry can be abstract in full generality, we now illustrate the projection in a simple and concrete setting. Specifically, we consider the Hilbert space case and \(\Omega^*\) is the squared \(L^2\) norm. In this case, learning the projection reduces to fitting the shaping network (i.e., the potential \(\Phi_\varphi\)) by solving:
\[
\varphi^{*}
= \arg\min_{\varphi}
\Bigl(
  r(s, a, s') - \bigl(\gamma \Phi_{\varphi}(s') - \Phi_{\varphi}(s)\bigr)
\Bigr)^{2}.
\]
To further understand this, we will continue to use the simple MDP defined in Figure~\ref{fig:toy-mdp-dim} as an example. Here, every transition from $s_1$ to $s_2$ is deterministic, so the shaping term is the same constant $F$ for both actions. Then the projection becomes to fit a single constant $F$ to best match the two rewards \(r_1 = r(s_1,a_1,s_2)\) and \(r_2 = r(s_1,a_2,s_2)\):
\[
\min_{F} \ (r_1 - F)^2 + (r_2 - F)^2 .
\]
This has a unique solution
\[
F^{\ast} = \frac{r_1 + r_2}{2}.
\]
And then the projected rewards:
\[
r_1' = r_1 - F^{\ast} = \frac{r_1 - r_2}{2},
\qquad
r_2' = r_2 - F^{\ast} = \frac{r_2 - r_1}{2},
\]
which is exactly the reward example in Figure~\ref{fig:dimension-shrink} (right).

\subsection{Analysis of Algorithm~\ref{alg:projected-ppo}}
\label{appendix:algorithm}

In this section, we justify our key design choices and provide an overall analysis of Algorithm~\ref{alg:projected-ppo}.

\paragraph{Pre-computing vs. on-policy learning of the projection.}
Theoretically, Theorem~\ref{theorem:core-regular-eq-robust} is stated for an ideal canonicalized base reward $r_{\mathrm{core}}=\mathrm{Proj}(r)$.
In principle, one may approximate this projection offline prior to policy optimization. For example, one can collect a large dataset, solve the projection problem to obtain $r_{\mathrm{core}}$, and then run a standard RL optimizer using this fixed projected reward. This approach is viable when the discrepancy
$\Omega^*$ is fixed (e.g., $\ell_p$ constrained uncertainty) and one is willing to pay the upfront data collection cost.

However, offline pre-computation can be undesirable for two reasons. First, it can be data-inefficient,
since obtaining an accurate projection may require a large coverage dataset that is collected solely for
canonicalization. Second, our framework allows general choices of $\Omega^*$. In some instances $\Omega^*$
is fixed, while in others it is defined through a trajectory-weighted objective (e.g., entropy/KL-induced forms)
that depends on the current policy and therefore changes as learning proceeds. In this latter case, a single
pre-computed projection may become stale as the policy evolves.

To obtain an algorithm that is both data-efficient and applicable to both policy-independent and
policy-dependent $\Omega^*$, we adopt the on-policy strategy in Algorithm~\ref{alg:projected-ppo}. The same rollouts are used to update the projection estimate and perform the policy update, so no
additional data collection is required. Importantly, if $\Omega^*$ depends on the policy, then the target
projection itself changes over training. Estimating the projection from rollouts under the current policy
allows the learned canonicalization to adapt online as $\pi$ evolves.

\paragraph{Convergence Analysis.}
Algorithm~1 can be shown to converge asymptotically using standard two-timescale stochastic approximation arguments~\cite{borkar2008stochastic,gadot2024solving}. We place the projection update on the fast timescale and the policy update on the slow timescale. Intuitively, the projection parameters track their near-optimal values for the current policy, so the policy updates evolve as if they were computed using an almost-converged projection. Under standard conditions, this separation of timescales yields convergence to a stationary point of the target objective. Next, we outlined the detailed analysis. We begin by stating the assumptions used throughout this section.

\begin{assumption}
\label{ass:compact-convex}
The iterates $\{\theta_i\}$ generated by the algorithm remain bounded almost surely:
\[
    \sup_i \|\theta_i\| < \infty \qquad \text{a.s.}
\]
\end{assumption}
This is a standard boundedness assumption used to ensure that the parameter iterates remain in a compact region, thereby enabling the convergence analysis.

\begin{assumption}
\label{ass:stationary}
For each fixed $\theta \in \Theta$, the Markov chain induced by $(\mu_0, \pi_\theta, P)$ on states is ergodic with unique stationary distribution, and the trajectory used in Algorithm~\ref{alg:projected-ppo} are sampled from the stationary regime (or from sufficiently long rollouts).
\end{assumption}
This assumption typically holds when the environment is a continuous task where every state remains reachable, and the policy continues to explore. Sampling from the stationary regime ensures unbiased stochastic gradients.

\begin{assumption}
\label{ass:strong-convexity}
    Let \(l(\varphi,\theta) := \Omega^*_{\pi_\theta}\!\bigl(r - F_{\Phi_\varphi}\bigr)\) denote the projection loss in Algorithm~\ref{alg:projected-ppo}, and let \(\ell(\varphi,\theta;\mathcal{D}_{\pi_\theta})\) denote its empirical estimate computed from a rollout dataset \(\mathcal{D}_{\pi_\theta}\). The population objective $L(\varphi,\theta):=\mathbb{E}[l(\varphi,\theta)|\theta]$ (the expectation of \(l\) over the data-generating distribution induced by the policy \(\pi_\theta\)) is $\mu$-strongly convex with $L_\varphi$-Lipschitz gradient in $\varphi$, uniformly over $\theta \in \Theta$, and $L_\theta$-Lipschitz gradient in $\theta$, uniformly over $\varphi \in \Psi$, hence admits a unique minimizer:
    \[
    \varphi^*(\theta):=\arg\min_{\varphi \in \Psi} L(\varphi, \theta).
    \]
    Moreover, the stochastic gradient satisfies:
    \[
\mathbb{E}\!\left[ \left\| \nabla_{\varphi} \ell(\varphi,\theta; \mathcal D_{\pi_\theta})
      - \nabla_{\varphi} L(\varphi,\theta) \right\|^{2}
      \,\big|\, \varphi,\theta \right] \le \sigma^{2}.
\]

\end{assumption}
This assumption holds when $\Phi_\varphi$ is tabular or linear in parameters (i.e., $\Phi_\varphi=\phi(s)^\top\varphi$ for some features $\phi(s)$) since in this case, minimizing $L(\varphi,\theta)$ actually solves a strongly convex problem, and the minimizer $\varphi^*$ is unique. For other function classes, for example, neural networks, this assumption does not necessarily hold, and in practice, we run SGD and treat the resulting $\Phi$ as an approximate projection and empirically observe stable behavior. The stochastic gradient assumption ensures the SGD update is stable.

\begin{assumption}
\label{ass:lipschitz}
The policy class $\pi_\theta$ is differentiable in $\theta$, and $\nabla_\theta \bar J(\theta, \varphi):=J(\pi_{\theta},r_{\varphi})-\Omega(\pi_{\theta})$ is Lipschitz in $(\theta,\varphi)$.
\end{assumption}

This usually holds for smooth policy-gradient surrogates, and PPO is one practical instantiation.

\begin{assumption}
\label{ass:two-timescale}
Let $t$ index every update, including inner $\varphi$ update and outer $\theta$ update. Stepsizes \(\{\alpha_t\}\) (for \(\phi\)) and
\(\{\beta_t\}\) (for \(\theta\)) satisfy
\[
\sum_t \alpha_t = \infty, \qquad \sum_t \alpha_t^2 < \infty, \qquad
\sum_t \beta_t = \infty, \qquad \sum_t \beta_t^2 < \infty, \qquad
\frac{\beta_t}{\alpha_t} \to 0.
\]
\end{assumption}
This assumption make sure that Algorithm~\ref{alg:projected-ppo} satisfies a two timescale argument where we update $\varphi$ (by SGD on $l(\varphi, \theta)$) fast while update $\theta$ (by policy gradient for $\bar J(\theta,\varphi)$) slow so that $\varphi$ stays near $\varphi^*(\pi_\theta)$ and if $\pi_\theta \rightarrow \pi_{\theta^*}$, then $\varphi \rightarrow \varphi^*(\pi_{\theta^*})$.

Next, before presenting the convergence proof, we first establish several supporting lemmas.

\begin{lemma}
\label{lemma:young}
    \[
\|a - b\|^{2} \;\le\; (1 + \eta)\,\|a\|^{2}
    + \left(1 + \frac{1}{\eta}\right)\|b\|^{2},
    \qquad \forall\, \eta > 0.
\]
\end{lemma}
\begin{proof}
Notice that
\[
\|a-b\|^{2}
= \|a\|^{2} + \|b\|^{2} - 2\langle a,b\rangle
\le \|a\|^{2} + \|b\|^{2} + 2|\langle a,b\rangle|
\le \|a\|^{2} + \|b\|^{2} + 2\|a\|\,\|b\|.
\]

Consider the nonnegative square
\[
    \Bigl(\sqrt{\eta}\,u - \frac{1}{\sqrt{\eta}}\,v\Bigr)^{2} \;\ge\; 0.
\]
Expanding gives
\[
    \eta u^{2} + \frac{1}{\eta} v^{2} - 2uv \;\ge\; 0
    \quad\Longrightarrow\quad
    2uv \;\le\; \eta u^{2} + \frac{1}{\eta} v^{2}.
\]
Let \(u=\|a\|, v=\|b\|\):
\[
2\|a\|\,\|b\| \le \eta \|a\|^{2} + \frac{1}{\eta}\|b\|^{2}.
\]

Hence
\[
\|a-b\|^{2}
\le (1+\eta)\|a\|^{2}
  + \left(1+\frac{1}{\eta}\right)\|b\|^{2}.
\]
This completes the proof.
\end{proof}

\begin{lemma}
    \label{lemma:lipschitz-varphi-theta}
Under Assumption~\ref{ass:strong-convexity}, there exists some constant $K$, such that
\[
  \bigl\|\varphi^{\star}(\theta_1)-\varphi^{\star}(\theta_2)\bigr\|
  \;\le\; K\,\bigl\|\theta_1-\theta_2\bigr\|,
  \qquad \forall\,\theta_1,\theta_2 .
\]
\end{lemma}
\begin{proof}
 Under Assumption~\ref{ass:strong-convexity},there exists $\mu > 0$ such that for all $\theta$ and all
$\varphi_1,\varphi_2$,
\[
    \big\langle
        \nabla_\varphi L(\varphi_1,\theta) - \nabla_\varphi L(\varphi_2,\theta),
        \,\varphi_1 - \varphi_2
    \big\rangle
    \;\ge\;
    \mu \,\|\varphi_1 - \varphi_2\|^2 .
\]
And there exists $L_{\theta} > 0$ such that for all $\varphi$ and all
$\theta_1,\theta_2$,
\[
    \big\|
        \nabla_\varphi L(\varphi,\theta_1)
        - \nabla_\varphi L(\varphi,\theta_2)
    \big\|
    \;\le\;
    L_{\theta}\,\|\theta_1 - \theta_2\| .
\]
Fix any $\theta_1,\theta_2$ and set
\[
    \varphi_1 := \varphi^{\star}(\theta_1),
    \qquad
    \varphi_2 := \varphi^{\star}(\theta_2).
\]
Because $\varphi_1$ and $\varphi_2$ minimize $L(\cdot,\theta_1)$ and
$L(\cdot,\theta_2)$, they satisfy the first-order optimality conditions:
\[
  \nabla_{\varphi} L(\varphi_1,\theta_1) = 0,
  \qquad
  \nabla_{\varphi} L(\varphi_2,\theta_2) = 0.
\]

Consider
\begin{align*}
  0
  &= \nabla_{\varphi} L(\varphi_1,\theta_1) - \nabla_{\varphi} L(\varphi_2,\theta_2) \\
  &= \underbrace{\bigl(\nabla_{\varphi} L(\varphi_1,\theta_1)
         - \nabla_{\varphi} L(\varphi_2,\theta_1)\bigr)}_{(1)}
   \;+\;
     \underbrace{\bigl(\nabla_{\varphi} L(\varphi_2,\theta_1)
         - \nabla_{\varphi} L(\varphi_2,\theta_2)\bigr)}_{(2)} .
\end{align*}

Take the inner product of both sides with $\varphi_1-\varphi_2$:
\[
  \bigl\langle(1),\varphi_1-\varphi_2\bigr\rangle
  \;=\;
  -\bigl\langle(2),\varphi_1-\varphi_2\bigr\rangle .
\]

Lower bound the left-hand side:
\[
  \bigl\langle(1),\varphi_1-\varphi_2\bigr\rangle
  = \bigl\langle
      \nabla_{\varphi} L(\varphi_1,\theta_1)
      - \nabla_{\varphi} L(\varphi_2,\theta_1),
      \,\varphi_1-\varphi_2
    \bigr\rangle
  \;\ge\; \mu \|\varphi_1-\varphi_2\|^2 .
\]

Upper bound the right-hand side:
\begin{align*}
  \bigl|\langle(2),\varphi_1-\varphi_2\rangle\bigr|
  &\le \|(2)\|\,\|\varphi_1-\varphi_2\|  \\
  &= \bigl\|
      \nabla_{\varphi} L(\varphi_2,\theta_1)
      - \nabla_{\varphi} L(\varphi_2,\theta_2)
     \bigr\|\,\|\varphi_1-\varphi_2\|              \\
  &\le L_{\theta}\,\|\theta_1-\theta_2\|\,\|\varphi_1-\varphi_2\| .
\end{align*}

Combining above gives
\[
  \mu \|\varphi_1-\varphi_2\|^2
  \;\le\;
  L_{\theta}\,\|\theta_1-\theta_2\|\,\|\varphi_1-\varphi_2\| .
\]
If $\varphi_1=\varphi_2$, the claim is trivial.  Otherwise, divide both sides by
$\|\varphi_1-\varphi_2\|$ to obtain
\[
  \|\varphi_1-\varphi_2\|
  \;\le\;
  \frac{L_{\theta}}{\mu}\,\|\theta_1-\theta_2\| =K\|\theta_1-\theta_2\|.
\]
This completes the proof.
\end{proof}

\begin{theorem}
    Consider Algorithm~\ref{alg:projected-ppo} and suppose Assumptions~\ref{ass:compact-convex},~\ref{ass:stationary},~\ref{ass:strong-convexity},~\ref{ass:lipschitz} and~\ref{ass:two-timescale} hold. Then, the iterates $(\theta_i, \varphi_i)$ generated by Algorithm~\ref{alg:projected-ppo} satisfy, almost surely:
    \begin{enumerate}
  \item The iterate for $\varphi$ tracks the instantaneous minimizer:
  \[
\|\varphi_i-\varphi^*(\theta_i)\|\to 0,
\qquad
\varphi^*(\theta):=\arg\min_\varphi L(\varphi,\theta).
\]

  \item Define the mean drift as $h(\theta,\varphi):=\mathbb{E}[\nabla_\theta \bar J\big(\theta,\varphi\big)|\theta,\varphi]$ The iterate $\{\theta_i\}$ is an asymptotic pseudo-trajectory of the ODE
    \[
    \dot\theta =h\big(\theta,\varphi^*(\theta)\big),
    \]
    hence every limit point of $\{\theta_i\}$ lies in an internally chain transitive invariant set of this ODE.

  \item  If the mean $\theta$-update is an exact gradient field
(i.e., the stochastic gradient estimator is unbiased so that the expected update $h(\theta,\varphi)$ equals $\nabla_\theta \bar J(\theta,\varphi)$),
then every limit point $\theta_\infty$ satisfies
\[
\nabla_\theta \bar J\big(\theta_\infty,\varphi^*(\theta_\infty)\big)=0.
\]
\end{enumerate}
\end{theorem}

\begin{proof}
1. For fixed outer index $i$, define the gradient noise at inner step $k$:
\[
M_{i,k+1}
    := \nabla_\varphi \ell(\varphi_{i,k},\theta_i; D_{\pi_{\theta_i}})
       - \nabla_\varphi L(\varphi_{i,k},\theta_i).
\]
By Assumption~\ref{ass:strong-convexity}, we have \(
\mathbb{E}\bigl[\|M_{i,k+1}\|^2 \mid \varphi_{i,k},\theta_i\bigr] \le \sigma^2.
\) Then the inner update can be written as
\[
\varphi_{i,k+1}
    = \varphi_{i,k}
      - \alpha_{i,k}\bigl(\nabla_\varphi L(\varphi_{i,k},\theta_i) + M_{i,k+1}\bigr),
\]
which is standard SGD on $L(\cdot,\theta_i)$. Let \(
e_{i,k} := \varphi_{i,k} - \varphi^*(\theta_i).
\)
Since $\varphi^*(\theta_i)$ minimizes $L(\cdot,\theta_i)$, we have \(
\nabla_\varphi L\bigl(\varphi^*(\theta_i),\theta_i\bigr) = 0.
\) Then we have,
\begin{align*}
\|e_{i,k+1}\|^2
 &=   \bigl\|  \varphi_{i,k+1} - \varphi^*(\theta_i)  \bigr\|^2   \\
 &=   \bigl\|  \varphi_{i,k}- \varphi^*(\theta_i)
      - \alpha_{i,k}\bigl(\nabla_\varphi L(\varphi_{i,k},\theta_i) + M_{i,k+1}\bigr)      \bigr\|^2     \\
 &= \bigl\|e_{i,k} - \alpha_{i,k}\bigl(\nabla_\varphi L(\varphi_{i,k},\theta_i)
     + M_{i,k+1}\bigr)\bigr\|^2 
\end{align*}
Take the conditional expectation given \(
\mathcal{F}_{i,k}
 := \sigma\bigl(\theta_i,\varphi_{i,0},\ldots,\varphi_{i,k},D_{\pi_{\theta_i}}\bigr).
\) We obtain:
\begin{align*}
\mathbb{E}\bigl[\|e_{i,k+1}\|^2 \mid \mathcal{F}_{i,k}\bigr]
&= \|e_{i,k}\|^2
   - 2\alpha_{i,k} \bigl\langle e_{i,k},
        \nabla_\varphi L(\varphi_{i,k},\theta_i) \bigr\rangle - 2\alpha_{i,k} \bigl\langle e_{i,k},
        \mathbb{E}\bigl[M_{i,k+1} \mid \mathcal{F}_{i,k}\bigr] \bigr\rangle\\
&\quad   + \alpha_{i,k}^2
     \bigl\|\nabla_\varphi L(\varphi_{i,k},\theta_i)\bigr\|^2 
   + \alpha_{i,k}^2 \,
     \mathbb{E}\bigl[\|M_{i,k+1}\|^2 \mid \mathcal{F}_{i,k}\bigr]. \\
     &\overset{(i)}{=} \|e_{i,k}\|^2
   - 2\alpha_{i,k} \bigl\langle e_{i,k},\nabla_\varphi L(\varphi_{i,k},\theta_i) \bigr\rangle  \\
   &\quad + \alpha_{i,k}^2
     \bigl\|\nabla_\varphi L(\varphi_{i,k},\theta_i)\bigr\|^2 
   + \alpha_{i,k}^2 \,
     \mathbb{E}\bigl[\|M_{i,k+1}\|^2 \mid \mathcal{F}_{i,k}\bigr]. \\
   &\overset{(ii)}{\le} \bigl(1 - 2\mu\alpha_{i,k} + L_\varphi^2 \alpha_{i,k}^2\bigr)\|e_{i,k}\|^2
       + \sigma^2 \alpha_{i,k}^2
\end{align*}
(i) uses $\mathbb{E}[M_{i,k+1} \mid \mathcal{F}_{i,k}] = 0$ (unbiased stochastic gradient under stationary sampling from Assumption~\ref{ass:stationary}); (ii) use the $\mu$-strong convexity, $L_\varphi$-Lipschitz and bounded second moment of stochastic gradient from Assumption~\ref{ass:strong-convexity}
\[
\bigl\langle e_{i,k}, \nabla_\varphi L(\varphi_{i,k},\theta_i) \bigr\rangle
    = \bigl\langle e_{i,k}, \nabla_\varphi L(\varphi_{i,k},\theta_i) -\nabla_\varphi L(\varphi^*(\theta_i),\theta_i)\bigr\rangle \ge \mu \|e_{i,k}\|^2,
\]
\[
\bigl\|\nabla_\varphi L(\varphi_{i,k},\theta_i)\bigr\|
    =  \bigl\|\nabla_\varphi L(\varphi_{i,k},\theta_i)-\nabla_\varphi L(\varphi^*(\theta_i),\theta_i)\bigr\| \le L_\varphi \|e_{i,k}\|.
\] 

Because $\sum_k \alpha_{i,k}^2 < \infty$ (Assumption~\ref{ass:two-timescale}) and
$\sum_k \alpha_{i,k} = \infty$, the Robbins-Siegmund supermartingale theorem~\cite{ROBBINS1971233} implies
\[
\|e_{i,k}\|^2 \longrightarrow 0
    \qquad \text{a.s.\ as } k \to \infty
    \quad (\text{for fixed } \theta_i).
\]
Next, according to Assumption~\ref{ass:two-timescale}, when we use $t$ index every update, then $\frac{\beta_t}{\alpha_t}\rightarrow 0$ means that $\theta$ moves so slowly that $\varphi$ has effectively infinite time to equilibrate between meaningful changes in $\theta$. Define \(e_t := \varphi_t - \varphi^*(\theta_t)\), where now $\theta$ changes over time, so the optimum is moving. Now when we update $\varphi$ at time $t$, the next error is \(
e_{t+1} = \varphi_{t+1} - \varphi^*(\theta_{t+1}) =(\varphi_{t+1}-\varphi^*(\theta_{t}))+(\varphi^*(\theta_{t})-\varphi^*(\theta_{t+1})).
\) Now use the inequality from Lemma~\ref{lemma:young}:
\[
\|a - b\|^{2} \;\le\; (1 + \eta)\,\|a\|^{2}
    + \left(1 + \frac{1}{\eta}\right)\|b\|^{2},
    \qquad \forall\, \eta > 0.
\]
Apply this with $a = \varphi_{t+1}-\varphi^*(\theta_{t})$, $b = \varphi^*(\theta_{t})-\varphi^*(\theta_{t+1}))$:
\[
\|e_{t+1}\|^{2}
\;\le\; (1 + \eta)\,\|\varphi_{t+1}-\varphi^*(\theta_{t})\|^{2}
    + \left(1 + \frac{1}{\eta}\right)\|\varphi^*(\theta_{t})-\varphi^*(\theta_{t+1}))\|^{2}.
\]
Take conditional expectation given $\mathcal{F}_t$:
\[
\mathbb{E}\bigl[\|e_{t+1}\|^{2} \mid \mathcal{F}_t\bigr]
\;\le\; (1 + \eta)\,\mathbb{E}\bigl[\|\varphi_{t+1}-\varphi^*(\theta_{t})\|^{2} \mid \mathcal{F}_t\bigr]
    + \left(1 + \frac{1}{\eta}\right)\|\varphi^*(\theta_{t})-\varphi^*(\theta_{t+1}))\|^{2}.
\]
As for bounding
$\mathbb{E}\bigl[\|\varphi_{t+1}-\varphi^*(\theta_{t})\|^{2}\mid \mathcal{F}_t\bigr]$, i.e., the error to the old minimizer $\varphi^*(\theta_t)$, based on the previous results regarding $e_{i,k+1}$, we have that there exist constant $c_1,c_2,c_3>0$, such that
\[
\mathbb{E}\bigl[\|\varphi_{t+1}-\varphi^*(\theta_{t})\|^{2}\mid \mathcal{F}_t\bigr] \le \bigl(1 - c_1\alpha_{t} + c_2 \alpha_{t}^2\bigr)\|e_{t}\|^2
       + c_3 \alpha_{t}^2
\]
Next, we bound \(\|\varphi^*(\theta_{t})-\varphi^*(\theta_{t+1}))\|^{2}\). Under Lemma~\ref{lemma:lipschitz-varphi-theta}, $\varphi^*(\theta)$ is Lipschitz:
\[
\|\varphi^*(\theta') - \varphi^*(\theta)\|
    \;\le\; K \,\|\theta' - \theta\| \quad \text{for some } K>0.
\]
So
\[
\|\varphi^*(\theta_{t})-\varphi^*(\theta_{t+1}))\| \;\le\; K \,\|\theta_{t+1} - \theta_t\|.
\]
But $\theta$ changes with step size $\beta_t$, i.e.,
\[
\|\theta_{t+1} - \theta_t\| = O(\beta_t).
\]
Hence we have:
\[
\|\varphi^*(\theta_{t})-\varphi^*(\theta_{t+1}))\|^{2}\le c_4 \beta_t^2, \qquad \text{for some } c_4>0.
\]
Now apply Robbins-Siegmund again. Since $\sum_t \alpha_t^2 < \infty$ and
$\sum_t \beta_t^2 < \infty$, the error terms are summable, and because
$\sum_t \alpha_t = \infty$, we conclude that:
\[
  \|e_{t+1}\| \longrightarrow 0 \quad \text{a.s.}
\]
Once we have $\|e_{t+1}\| \to 0$ for the global time $t$, it implies that at the
subsequence of times corresponding to the end of each outer iteration
(i.e., when we set $\varphi_{i+1} = \varphi_{i,K_i}$), we also have
\[
  \|\varphi_i - \varphi^*(\theta_i)\| \longrightarrow 0 \quad \text{a.s.}
\]

2. The $\theta$-update can be written as
\[
  \theta_{i+1}
  = \theta_i + \beta_i \bigl( h(\theta_i, \varphi_{i+1}) + \xi_{i+1} \bigr),
\]
where $\{\xi_{i+1}\}$ is a noise term:
\[
  \xi_{i+1}
  := \nabla_\theta \bar J(\theta_i, \varphi_{i+1}; D'_{\pi_{\theta_i}})
     - h(\theta_i, \varphi_{i+1}) .
\]
The Lipschitz assumption on $\nabla_\theta \bar J(\theta, \varphi)$ (Assumption~\ref{ass:lipschitz})
implies that $h$ is Lipschitz in $(\theta,\varphi)$ on the bounded region visited
by the iterates since expectation preserves Lipschitzness. Boundedness of $\{\theta_i\}$ in Assumption~\ref{ass:compact-convex} and also $\sum_i \beta_i^2 < \infty$ (Assumption~\ref{ass:two-timescale}) ensures that the accumulated noise is controlled in the ODE method. Now use $\|\varphi_{i+1} - \varphi^*(\theta_i)\| \to 0$ a.s.
By Lipschitzness of $h$,
\[
  \bigl\| h(\theta_i, \varphi_{i+1}) - h(\theta_i, \varphi^*(\theta_i)) \bigr\|
  \;\le\; L_h \bigl\|\varphi_{i+1} - \varphi^*(\theta_i)\bigr\|
  \;\longrightarrow\; 0
  \quad \text{a.s.}
\]
Hence the $\theta$-recursion is an SA scheme with vanishing bias:
\[
  \theta_{i+1}
  = \theta_i
    + \beta_i \Bigl( h(\theta_i, \varphi^*(\theta_i))
      + o(1)
      + \xi_{i+1} \Bigr).
\]
Standard ODE-method conclusions (Kushner-Clark two-timescale SA~\cite{Borkar2008,doi:10.1137/S0363012997331639}) then imply that the continuous-time interpolation of $\{\theta_i\}$ is an
asymptotic pseudo-trajectory of the limiting ODE
\[
  \dot{\theta} = h\bigl(\theta, \varphi^*(\theta)\bigr).
\]
3. If we assume \(h\bigl(\theta, \varphi^*(\theta)\bigr)=\nabla_\theta \bar J(\theta,\varphi)\), then the limiting ODE is
\[
  \dot{\theta} = \nabla_\theta \bar J\bigl(\theta, \varphi^*(\theta)\bigr).
\]
Use \(\bar J(\theta)\) short for \(J\bigl(\theta, \varphi^*(\theta)\bigr)\), consider any solution trajectory $\theta(t)$ of this ODE. By the chain rule,
\[
  \frac{d}{dt}\,\bar J(\theta(t))
  \;=\;
  \big\langle \nabla \bar J(\theta(t)),\, \dot\theta(t) \big\rangle.
\]
Substituting $\dot\theta(t) = \nabla \bar J(\theta(t))$ yields
\[
  \frac{d}{dt}\,\bar J(\theta(t))
  \;=\;
  \big\langle \nabla \bar J(\theta(t)),\, \nabla \bar J(\theta(t)) \big\rangle
  \;=\;
  \big\|\nabla \bar J(\theta(t))\big\|^2
  \;\ge\; 0.
\]
Therefore $\bar J(\theta(t))$ is non-decreasing along trajectories, and the
only points where $\bar J$ stops increasing are those for which
$\nabla \bar J(\theta) = 0$. Hence every limit point of the ODE must
lie in the set of stationary points
\[
  \{\theta : \nabla \bar J(\theta) = 0\}.
\]
Since the stochastic iterates $\{\theta_i\}$ form an asymptotic
pseudo-trajectory of the ODE, their limit points are
contained in the same stationary set, i.e.,
\[
  \nabla \bar J(\theta_\infty) = 0.
\]
Equivalently,
\[
  \nabla_\theta \bar J\bigl(\theta_\infty, \varphi^*(\theta_\infty)\bigr) = 0,
\]
This completes the proof.
\end{proof}

\paragraph{Projection Error Analysis.} 
In Algorithm~\ref{alg:projected-ppo}, the projection objective is estimated from a finite batch of rollout samples, and the projector is learned approximately using a neural network. As a result, the learned projection may deviate from the exact population projector. We next quantify this approximation error. Fix the policy parameter \(\theta\), let \(l(\varphi,\theta) := \Omega^*_{\pi_\theta}\!\bigl(r - F_{\Phi_\varphi}\bigr)\) denote the projection loss in Algorithm~\ref{alg:projected-ppo}, and Let \(l(\varphi,\theta;\tau)\) denote the
projection loss evaluated on a rollout sample \(\tau\), and define the population
projection objective
\[
L(\varphi,\theta)
:=
\mathbb E\!\left[
l(\varphi,\theta;\tau)\mid \theta
\right],
\qquad
\tau\sim\mathcal D_{\pi_\theta}.
\]
Given \(N\) i.i.d. rollouts \(\tau_1,\ldots,\tau_N\sim\mathcal D_{\pi_\theta}\), define
the empirical objective
\[
\hat L(\varphi,\theta)
:=
\frac{1}{N}\sum_{n=1}^{N}l(\varphi,\theta;\tau_n).
\]
We make the following standard assumptions:
\begin{assumption}
\label{ass:bounded-projection-class}
The projection parameters $\varphi$ are restricted to a bounded set \(\Psi\subset\mathbb R^p\), where \(p\) is the dimension of the projection parameter. In particular, there exists \(R>0\) such that
\[
    \|\varphi\|_2\le R,\qquad \forall \varphi\in\Psi,
\]
\end{assumption}
This is similar to Assumption~\ref{ass:compact-convex}. Here, we assume that the projection parameter \(\varphi\) also remains bounded.

\begin{assumption}
\label{ass:bounded-projection-loss}
For fixed \(\theta\), the sample projection loss is uniformly bounded. That is,
there exists \(B_\ell>0\) such that, for all \(\varphi\in\Psi\) and rollout samples
\(\tau\),
\[
    |l(\varphi,\theta;\tau)|\le B_\ell .
\]
\end{assumption}
This is a standard bounded-loss assumption, which allows us to control the finite-sample estimation error of the projection objective.

\begin{assumption}
\label{ass:lipschitz-projection-objectives}
For fixed \(\theta\), the empirical projection objective \(\hat L(\cdot,\theta)\) is Lipschitz on \(\Psi\). That is, there exists \(G_\varphi>0\) such that, for all \(\varphi,\varphi'\in\Psi\),
\[
|\hat L(\varphi,\theta)-\hat L(\varphi',\theta)|
\le
G_\varphi\|\varphi-\varphi'\|_2 .
\]
\end{assumption}
This is the empirical-objective counterpart of Assumption~\ref{ass:strong-convexity}, which imposes regularity on the population projection objective. Here, we similarly assume that the empirical projection objective is Lipschitz.

\begin{theorem}
\label{theorem:projection-error}
Under Assumption~\ref{ass:bounded-projection-class},~\ref{ass:bounded-projection-loss},~\ref{ass:lipschitz-projection-objectives} and~\ref{ass:strong-convexity}, let
\[
\varphi^\star\in\arg\min_{\varphi\in\Psi}L(\varphi,\theta),
\qquad
\hat\varphi\in\arg\min_{\varphi\in\Psi}\hat L(\varphi,\theta).
\]
For fixed $\theta$, with probability at least \(1-\delta\), we have
\[
L(\hat\varphi,\theta)-L(\varphi^\star,\theta)
=
\mathcal O\left(
\sqrt{
\frac{
p\log N+\log(1/\delta)
}{N}
}
\right).
\]
\end{theorem}

\begin{proof}
Fix \(\theta\). For any fixed \(\varphi\in\Psi\), the random variables
\[
l(\varphi,\theta;\tau_1),\ldots,l(\varphi,\theta;\tau_N)
\]
are i.i.d. under the rollout distribution \(\mathcal D_{\pi_\theta}\), and by
Assumption~\ref{ass:bounded-projection-loss}, each of them is bounded in
\([-B_l,B_l]\). Hoeffding's inequality~\cite{hoeffding1963probability} says that if $l(\varphi,\theta;\tau_n) \in [a_n, b_n]$, then:
\[
\Pr\!\left(\left|\sum_{n=1}^{N} l(\varphi,\theta;\tau_n)-\mathbb{E}\sum_{n=1}^{N}l(\varphi,\theta;\tau_n)\right|\ge t\right)
\le
2\exp\!\left(
-\frac{2t^2}{\sum_{n=1}^{N}(b_n-a_n)^2}
\right).
\]

Here $a_n=-B_\ell$, $b_n=B_\ell$, so $b_n-a_n=2B_\ell$, and therefore
\[
\sum_{n=1}^{N}(b_n-a_n)^2 = 4N B_\ell^2.
\]
Thus,
\[
\Pr\!\left(\left|\sum_{n=1}^{N} l(\varphi,\theta;\tau_n)-\mathbb{E}\sum_{n=1}^{N}l(\varphi,\theta;\tau_n)\right|\ge t\right)
\le
2\exp\!\left(-\frac{t^2}{2N B_\ell^2}\right).
\]

Now divide by $N$. Since
\[
\hat L(\varphi,\theta)=\frac{1}{N}\sum_{n=1}^{N}l(\varphi,\theta;\tau_n),
\qquad
L(\varphi,\theta)=\mathbb{E}[l(\varphi,\theta;\tau_n)],
\]
we have
\[
\bigl|\hat L(\varphi,\theta)-L(\varphi,\theta)\bigr|
=
\frac{1}{N}\left|\sum_{n=1}^{N}l(\varphi,\theta;\tau_n)-\mathbb{E}\sum_{n=1}^{N}l(\varphi,\theta;\tau_n)\right|.
\]
Setting $t=N\varepsilon$, we get
\[
\Pr\!\left(\bigl|\hat L(\varphi,\theta)-L(\varphi,\theta)\bigr|\ge \varepsilon\right)
\le
2\exp\!\left(-\frac{N\varepsilon^2}{2B_\ell^2}\right).
\]

Now set the right-hand side equal to $\delta$:
\[
2\exp\!\left(-\frac{N\varepsilon^2}{2B_\ell^2}\right)=\delta.
\]
Solving for $\varepsilon$,
\[
\varepsilon
=
B_\ell\sqrt{\frac{2\log(2/\delta)}{N}}.
\]

Therefore, for any fixed $\varphi$, with probability at least $1-\delta$,
\[
\bigl|\hat L(\varphi,\theta)-L(\varphi,\theta)\bigr|
\le
B_\ell\sqrt{\frac{2\log(2/\delta)}{N}}.
\]
Notice that this is a pointwise concentration bound, as it holds for one fixed value of
\(\varphi\). To obtain a uniform bound over all \(\varphi\in\Psi\), we use a
standard metric-covering argument~\cite{vershynin2018high}. Let \(\mathcal N_\rho\subset\Psi\) be a finite
\(\rho\)-cover of \(\Psi\) under the Euclidean norm, meaning that for every
\(\varphi\in\Psi\), there exists \(\varphi_\rho\in\mathcal N_\rho\) such that
\[
    \|\varphi-\varphi_\rho\|_2\le \rho .
\]
Since \(\Psi\subset\mathbb R^p\) is contained in a radius-\(R\) ball by
Assumption~\ref{ass:bounded-projection-class}, every $\varphi\in\Psi$
lies inside the Euclidean ball
\[
B_2(R)=\{\varphi\in\mathbb{R}^p:\ \|\varphi\|_2\le R\}.
\]
So it is enough to cover the larger set $B_2(R)$. If we can cover $B_2(R)$, then we automatically
cover $\Psi$. By the standard covering-number bound for Euclidean balls (Corollary~4.2.11 in \cite{vershynin2018high}), the radius-\(R\) Euclidean ball in
\(\mathbb R^p\) admits a \(\rho\)-cover with cardinality at most
\[
    \left(1+\frac{2R}{\rho}\right)^p .
\]
For \(\rho\le R\), this is bounded by
\[
    \left(\frac{3R}{\rho}\right)^p .
\]
Thus we may choose \(\mathcal N_\rho\) such that
\[
    |\mathcal N_\rho|
    \le
    \left(\frac{3R}{\rho}\right)^p .
\]
Intuitively, the factor \(R/\rho\) counts the number of grid points needed along
one coordinate direction, and the exponent \(p\) comes from covering a
\(p\)-dimensional parameter space.

Now apply the pointwise concentration bound to each element of
\(\mathcal N_\rho\). For any fixed \(\varphi\in\mathcal N_\rho\), the previous
argument gives
\[
\Pr\!\left(
\bigl|\hat L(\varphi,\theta)-L(\varphi,\theta)\bigr|
\ge \varepsilon
\right)
\le
2\exp\!\left(-\frac{N\varepsilon^2}{2B_\ell^2}\right).
\]
Therefore, 
\[
\begin{aligned}
&\Pr\!\left(
\sup_{\varphi\in\mathcal N_\rho}
\bigl|\hat L(\varphi,\theta)-L(\varphi,\theta)\bigr|
\ge \varepsilon
\right) \\
&\le
\sum_{\varphi\in\mathcal N_\rho}
\Pr\!\left(
\bigl|\hat L(\varphi,\theta)-L(\varphi,\theta)\bigr|
\ge \varepsilon
\right) \\
&\le
2|\mathcal N_\rho|
\exp\!\left(-\frac{N\varepsilon^2}{2B_\ell^2}\right).
\end{aligned}
\]
Using
\[
|\mathcal N_\rho|
\le
\left(\frac{3R}{\rho}\right)^p,
\]
we obtain
\[
\Pr\!\left(
\sup_{\varphi\in\mathcal N_\rho}
\bigl|\hat L(\varphi,\theta)-L(\varphi,\theta)\bigr|
\ge \varepsilon
\right)
\le
2
\exp\!\left(
p\log(3R/\rho)
-
\frac{N\varepsilon^2}{2B_\ell^2}
\right).
\]
Set the right-hand side equal to \(\delta\):
\[
2
\exp\!\left(
p\log(3R/\rho)
-
\frac{N\varepsilon^2}{2B_\ell^2}
\right)
=
\delta.
\]
Solving for \(\varepsilon\), we get
\[
\varepsilon
=
B_\ell
\sqrt{
\frac{
2\left(p\log(3R/\rho)+\log(2/\delta)\right)
}{N}
}.
\]
Thus, with probability at least \(1-\delta\),
\[
\sup_{\varphi\in\mathcal N_\rho}
\bigl|
\hat L(\varphi,\theta)-L(\varphi,\theta)
\bigr|
\le
B_\ell
\sqrt{
\frac{
2\left(p\log(3R/\rho)+\log(2/\delta)\right)
}{N}
}.
\]
It remains to extend the bound from the finite cover \(\mathcal N_\rho\) to the
whole parameter set \(\Psi\). 
By Assumption~\ref{ass:strong-convexity}, \(L(\cdot,\theta)\) is
\(L_\varphi\)-Lipschitz in \(\varphi\) on \(\Psi\). By
Assumption~\ref{ass:lipschitz-projection-objectives},
\(\hat L(\cdot,\theta)\) is \(G_\varphi\)-Lipschitz in \(\varphi\) on \(\Psi\).
Let
\[
\bar L_\varphi := \max\{L_\varphi,G_\varphi\}.
\]
Then, for any \(\varphi\in\Psi\), choose
\(\varphi_\rho\in\mathcal N_\rho\) such that
\[
\|\varphi-\varphi_\rho\|_2\le \rho .
\]
We have
\[
\left|
L(\varphi,\theta)-L(\varphi_\rho,\theta)
\right|
\le
L_\varphi\rho
\le
\bar L_\varphi\rho,
\]
and
\[
\left|
\hat L(\varphi,\theta)-\hat L(\varphi_\rho,\theta)
\right|
\le
G_\varphi\rho
\le
\bar L_\varphi\rho.
\]
Therefore,
\[
\begin{aligned}
\left|
\hat L(\varphi,\theta)-L(\varphi,\theta)
\right|
&\le
\left|
\hat L(\varphi_\rho,\theta)-L(\varphi_\rho,\theta)
\right|\\
&\quad+
\left|
\hat L(\varphi,\theta)-\hat L(\varphi_\rho,\theta)
\right|
+
\left|
L(\varphi,\theta)-L(\varphi_\rho,\theta)
\right|\\
&\le
\sup_{\varphi'\in\mathcal N_\rho}
\left|
\hat L(\varphi',\theta)-L(\varphi',\theta)
\right|
+
2\bar L_\varphi\rho .
\end{aligned}
\]
Taking the supremum over \(\varphi\in\Psi\), we obtain that, with probability at
least \(1-\delta\),
\[
\sup_{\varphi\in\Psi}
\left|
\hat L(\varphi,\theta)-L(\varphi,\theta)
\right|
\le
B_\ell
\sqrt{
\frac{
2\left(p\log(3R/\rho)+\log(2/\delta)\right)
}{N}
}
+
2\bar L_\varphi\rho .
\]
Now choose
\[
\rho=N^{-1/2}.
\]
Then
\[
p\log(3R/\rho)
=
p\log(3R\sqrt N)
=
p\log(3R)+\frac{p}{2}\log N
=
\mathcal O(p\log N).
\]
Moreover,
\[
2\bar L_\varphi\rho
=
\frac{2\bar L_\varphi}{\sqrt N}
=
\mathcal O\left(\sqrt{\frac{1}{N}}\right)
\le
\mathcal O\left(
\sqrt{
\frac{
p\log N+\log(1/\delta)
}{N}
}
\right),
\]
for \(p\ge 1\), \(N\ge 2\), and \(\delta\in(0,1)\). Therefore,
\[
\sup_{\varphi\in\Psi}
\left|
\hat L(\varphi,\theta)-L(\varphi,\theta)
\right|
=
\mathcal O\left(
\sqrt{
\frac{
p\log N+\log(1/\delta)
}{N}
}
\right).
\]
Since absolute value is symmetric, this is equivalent to
\[
\sup_{\varphi\in\Psi}
\left|
L(\varphi,\theta)-\hat L(\varphi,\theta)
\right|
=
\mathcal O\left(
\sqrt{
\frac{
p\log N+\log(1/\delta)
}{N}
}
\right).
\]
Since
\[
\varphi^\star\in\arg\min_{\varphi\in\Psi}L(\varphi,\theta),
\qquad
\hat\varphi\in\arg\min_{\varphi\in\Psi}\hat L(\varphi,\theta),
\]
we have
\[
\begin{aligned}
L(\hat\varphi,\theta)-L(\varphi^\star,\theta)
&=
\bigl[L(\hat\varphi,\theta)-\hat L(\hat\varphi,\theta)\bigr]
+
\bigl[\hat L(\hat\varphi,\theta)-\hat L(\varphi^\star,\theta)\bigr] \\
&\quad+
\bigl[\hat L(\varphi^\star,\theta)-L(\varphi^\star,\theta)\bigr].
\end{aligned}
\]
Because \(\hat\varphi\) minimizes \(\hat L(\cdot,\theta)\), we have
\[
\hat L(\hat\varphi,\theta)-\hat L(\varphi^\star,\theta)\le 0.
\]
Therefore,
\[
L(\hat\varphi,\theta)-L(\varphi^\star,\theta)
\le
2\sup_{\varphi\in\Psi}
\left|
L(\varphi,\theta)-\hat L(\varphi,\theta)
\right|.
\]
Combining this with the previous uniform convergence bound gives, with probability at least \(1-\delta\),
\[
L(\hat\varphi,\theta)-L(\varphi^\star,\theta)
=
\mathcal O\left(
\sqrt{
\frac{
p\log N+\log(1/\delta)
}{N}
}
\right).
\]
This completes the proof.
\end{proof}
Notice that the proof does not require the strong convexity part of Assumption~\ref{ass:strong-convexity}. It only uses the Lipschitz regularity of the projection objectives. Theorem~\ref{theorem:projection-error} shows that the statistical complexity of projection learning does not scale directly with the size of the state-action space, but instead with the dimension \(p\) of the projection parameterization. Thus, when the potential function is represented by a compact model, the number of rollout samples needed to learn a reliable projection can remain moderate even in large environments. The bound also clarifies the role of the minibatch size \(N\): increasing \(N\) reduces the projection error at the standard \(1/\sqrt{N}\) rate, up to a logarithmic factor. Therefore, projection error can be controlled by using larger rollout batches, while the dependence on the confidence level is only logarithmic through \(\log(1/\delta)\). 

Next, we translate the projection-objective error into a bound on the distance between the exact projected reward and its empirical approximation. We use the following regularity assumption:
\begin{assumption}
\label{ass:lipschitz-shaping-map}
There exists \(L_F>0\) such that, for all \(\varphi,\varphi'\in\Psi\),
\[
\left\|
F_{\Phi_\varphi}-F_{\Phi_{\varphi'}}
\right\|
\le
L_F\|\varphi-\varphi'\|_2 .
\]
\end{assumption}
This is a standard assumption ensuring that the potential function is Lipschitz in the projection parameter.

\begin{theorem}
\label{theorem:reward-projection-error}
Fix \(\theta\). Define the exact and empirical projected rewards as
\[
r_{\mathrm{core}}
:=
r-F_{\Phi_{\varphi^\star}},
\qquad
\hat r_{\mathrm{core}}
:=
r-F_{\Phi_{\hat\varphi}}.
\]
Then, under Assumption~\ref{ass:bounded-projection-class},~\ref{ass:bounded-projection-loss},~\ref{ass:lipschitz-projection-objectives},~\ref{ass:strong-convexity}, and~\ref{ass:lipschitz-shaping-map}, with
probability at least \(1-\delta\),
\[
\left\|
\hat r_{\mathrm{core}}-r_{\mathrm{core}}
\right\|
=
\mathcal O\!\left(
\left(
\frac{
p\log N+\log(1/\delta)
}{N}
\right)^{1/4}
\right).
\]
\end{theorem}
\begin{proof}
By Theorem~\ref{theorem:projection-error}, with probability at least
\(1-\delta\),
\[
L(\hat\varphi,\theta)-L(\varphi^\star,\theta)
=
\mathcal O\!\left(
\sqrt{
\frac{
p\log N+\log(1/\delta)
}{N}
}
\right).
\]
Since \(L(\cdot,\theta)\) is \(\mu\)-strongly convex in \(\varphi\) by
Assumption~\ref{ass:strong-convexity}, and \(\varphi^\star\) minimizes
\(L(\cdot,\theta)\), we have
\[
L(\hat\varphi,\theta)-L(\varphi^\star,\theta)
\ge
\frac{\mu}{2}
\|\hat\varphi-\varphi^\star\|_2^2 .
\]
Therefore,
\[
\|\hat\varphi-\varphi^\star\|_2
\le
\sqrt{
\frac{2}{\mu}
\left(
L(\hat\varphi,\theta)-L(\varphi^\star,\theta)
\right)
}.
\]
Combining this with Theorem~\ref{theorem:projection-error} gives
\[
\|\hat\varphi-\varphi^\star\|_2
=
\mathcal O\!\left(
\left(
\frac{
p\log N+\log(1/\delta)
}{N}
\right)^{1/4}
\right).
\]
Finally, by the Lipschitzness of the potential function in Assumption~\ref{ass:lipschitz-shaping-map},
\[
\begin{aligned}
\left\|
\hat r_{\mathrm{core}}-r_{\mathrm{core}}
\right\|
&=
\left\|
\bigl(r-F_{\Phi_{\hat\varphi}}\bigr)
-
\bigl(r-F_{\Phi_{\varphi^\star}}\bigr)
\right\| \\
&=
\left\|
F_{\Phi_{\varphi^\star}}-F_{\Phi_{\hat\varphi}}
\right\| \\
&\le
L_F\|\hat\varphi-\varphi^\star\|_2 .
\end{aligned}
\]
Thus,
\[
\left\|
\hat r_{\mathrm{core}}-r_{\mathrm{core}}
\right\|
=
\mathcal O\!\left(
\left(
\frac{
p\log N+\log(1/\delta)
}{N}
\right)^{1/4}
\right).
\]
This completes the proof.
\end{proof}

Finally, we quantify how the empirical projection error propagates to the robust objectives in Theorems~\ref{theorem:core-regular-eq-robust} and~\ref{theorem:entropy-kl-exact-duality}. We begin with a widely used lemma~\cite{puterman2014markov,kearns2002near}, which will be useful for the analysis.
\begin{lemma}
\label{lemma:return-error-approx-projection}
Let \(r_{\mathrm{core}}\) be the exact projected reward and let
\(\hat r_{\mathrm{core}}\) be its empirical approximation. Define
\[
\Delta_r
:=
\|r_{\mathrm{core}}-\hat r_{\mathrm{core}}\| .
\]
Then, for any policy \(\pi\),
\[
\left|
J(\pi,r_{\mathrm{core}})
-
J(\pi,\hat r_{\mathrm{core}})
\right|
\le
\frac{\Delta_r}{1-\gamma}.
\]
\end{lemma}

\begin{proof}
Define the discounted occupancy measure
\[
d_\pi(s,a,s')
:=
\mathbb E\left[
\sum_{t=0}^{\infty}
\gamma^t
\mathbf 1\{s_t=s,a_t=a,s_{t+1}=s'\}
\Bigm|\mu_0,\pi,P
\right].
\]
Then, for any reward \(r\),
\[
J(\pi,r)=\langle d_\pi,r\rangle.
\]
Therefore,
\[
\begin{aligned}
\left|
J(\pi,r_{\mathrm{core}})
-
J(\pi,\hat r_{\mathrm{core}})
\right|
&=
\left|
\left\langle
d_\pi,
r_{\mathrm{core}}-\hat r_{\mathrm{core}}
\right\rangle
\right|\\
&\le
\|d_\pi\|
\,
\|r_{\mathrm{core}}-\hat r_{\mathrm{core}}\| .
\end{aligned}
\]
Moreover,
\[
\|d_\pi\|
\le
\|d_\pi\|_1
=
\sum_{s,a,s'}d_\pi(s,a,s')
=
\sum_{t=0}^{\infty}\gamma^t
=
\frac{1}{1-\gamma}.
\]
Thus,
\[
\left|
J(\pi,r_{\mathrm{core}})
-
J(\pi,\hat r_{\mathrm{core}})
\right|
\le
\frac{
\|r_{\mathrm{core}}-\hat r_{\mathrm{core}}\|
}{1-\gamma}.
\]
This completes the proof.
\end{proof}
Next, we bound the return difference between the exact projected reward and its empirical approximation.
\begin{lemma}
\label{lemma:return-error-finite-sample-projection}
Under the assumptions of Theorem~\ref{theorem:reward-projection-error}, with
probability at least \(1-\delta\), for any policy \(\pi\),
\[
\left|
J(\pi,r_{\mathrm{core}})
-
J(\pi,\hat r_{\mathrm{core}})
\right|
=
\mathcal O\left(
\frac{1}{1-\gamma}
\left(
\frac{
p\log N+\log(1/\delta)
}{N}
\right)^{1/4}
\right).
\]
\end{lemma}

\begin{proof}
By Theorem~\ref{theorem:reward-projection-error}, with probability at least
\(1-\delta\),
\[
\|r_{\mathrm{core}}-\hat r_{\mathrm{core}}\|_2
=
\mathcal O\left(
\left(
\frac{
p\log N+\log(1/\delta)
}{N}
\right)^{1/4}
\right).
\]
Combining this with Lemma~\ref{lemma:return-error-approx-projection} gives
\[
\left|
J(\pi,r_{\mathrm{core}})
-
J(\pi,\hat r_{\mathrm{core}})
\right|
=
\mathcal O\left(
\frac{1}{1-\gamma}
\left(
\frac{
p\log N+\log(1/\delta)
}{N}
\right)^{1/4}
\right).
\]
\end{proof}

\begin{theorem}
\label{theorem:approx-lower-bound}
Under the assumptions of Lemma~\ref{lemma:return-error-finite-sample-projection},
with probability at least \(1-\delta\), for any policy \(\pi\),
\[
\min_{\tilde r_{\mathrm{core}}\in\mathcal U_{\mathrm{core}}}
J(\pi,\tilde r_{\mathrm{core}})
\ge
J(\pi,\hat r_{\mathrm{core}})
-
\lambda\Omega(\pi)
-
\lambda\epsilon
-
\alpha_N,
\]
where
\[
\alpha_N
:=
\mathcal O\left(
\frac{1}{1-\gamma}
\left(
\frac{
p\log N+\log(1/\delta)
}{N}
\right)^{1/4}
\right).
\]
\end{theorem}

\begin{proof}
By Theorem~\ref{theorem:core-regular-eq-robust}, for any policy \(\pi\),
\[
\min_{\tilde r_{\mathrm{core}}\in\mathcal U_{\mathrm{core}}}
J(\pi,\tilde r_{\mathrm{core}})
\ge
J(\pi,r_{\mathrm{core}})
-
\lambda\Omega(\pi)
-
\lambda\epsilon.
\]
By Lemma~\ref{lemma:return-error-finite-sample-projection},
\[
J(\pi,r_{\mathrm{core}})
\ge
J(\pi,\hat r_{\mathrm{core}})-\alpha_N.
\]
Substituting this into the previous inequality gives
\[
\min_{\tilde r_{\mathrm{core}}\in\mathcal U_{\mathrm{core}}}
J(\pi,\tilde r_{\mathrm{core}})
\ge
J(\pi,\hat r_{\mathrm{core}})
-
\lambda\Omega(\pi)
-
\lambda\epsilon
-
\alpha_N.
\]
This completes the proof.
\end{proof}
Theorem~\ref{theorem:approx-lower-bound} shows that using the empirical projected reward \(\hat r_{\mathrm{core}}\) does not change the structure of Theorem~\ref{theorem:core-regular-eq-robust}. It only introduces an additional finite-sample error term \(\alpha_N\). This error depends on the number of rollout samples used to learn the projection and vanishes as \(N\) increases. Thus, the regularized objective with \(\hat r_{\mathrm{core}}\) remains a valid conservative surrogate for the shaping-aware robust objective, up to a controllable approximation error.

\begin{theorem}
\label{theorem:approx-entropy-kl-duality}
Under the assumptions of Lemma~\ref{lemma:return-error-finite-sample-projection},
for the entropy- and KL-induced uncertainty sets in Theorem~\ref{theorem:entropy-kl-exact-duality},
with probability at least \(1-\delta\), for any policy \(\pi\),
\[
\left|
\min_{\tilde r_{\mathrm{core}}\in\mathcal U_{\mathrm{core}}}
J(\pi,\tilde r_{\mathrm{core}})
-
\left[
J(\pi,\hat r_{\mathrm{core}})
+
\lambda\Omega(\pi)
-
\epsilon
\right]
\right|
\le
\mathcal O\left(
\frac{1}{1-\gamma}
\left(
\frac{
p\log N+\log(1/\delta)
}{N}
\right)^{1/4}
\right).
\]
\end{theorem}

\begin{proof}
By Theorem~\ref{theorem:entropy-kl-exact-duality},
\[
\min_{\tilde r_{\mathrm{core}}\in\mathcal U_{\mathrm{core}}}
J(\pi,\tilde r_{\mathrm{core}})
=
J(\pi,r_{\mathrm{core}})
+
\lambda\Omega(\pi)
-
\epsilon.
\]
Therefore,
\[
\begin{aligned}
&\left|
\min_{\tilde r_{\mathrm{core}}\in\mathcal U_{\mathrm{core}}}
J(\pi,\tilde r_{\mathrm{core}})
-
\left[
J(\pi,\hat r_{\mathrm{core}})
+
\lambda\Omega(\pi)
-
\epsilon
\right]
\right|\\
&=
\left|
J(\pi,r_{\mathrm{core}})
-
J(\pi,\hat r_{\mathrm{core}})
\right|.
\end{aligned}
\]
Applying Lemma~\ref{lemma:return-error-finite-sample-projection} gives the claim.
\end{proof}

Theorem~\ref{theorem:approx-entropy-kl-duality} shows that, when the projection is learned from finite rollout data, the regularized objective with the empirical projected reward \(\hat r_{\mathrm{core}}\) approximates the shaping-aware reward-robust objective up to an additive error
\(
\mathcal O\left(
\frac{1}{1-\gamma}
\left(
\frac{p\log N+\log(1/\delta)}{N}
\right)^{1/4}
\right).
\)
This error vanishes as the number of rollout samples \(N\) increases. Therefore, the empirical projection does not change the exact duality in Theorem~\ref{theorem:entropy-kl-exact-duality} in the large-sample limit. It only introduces a finite-sample approximation error.

\section{Additional Implementation Details}
\label{appendix:implementation}

All experiments were conducted on a single NVIDIA RTX 4090 GPU (24GB memory) and a 13th Gen Intel Core i9-13900KF CPU (32 threads). We implemented all methods in Python 3.9 using PyTorch 2.6.0~\citep{paszke2019pytorch}, and SB3~\cite{stable-baselines3}.

\subsection{Implementation Details for MiniGrid}
We use a simple deterministic \(N\times N\) grid-world task as in previous work~\cite{ashlag2025state,chevalier2018minimalistic}. As shown in Figure~\ref{fig:grid-world}, the agent starts in the top-left cell and aims to reach a fixed goal in the bottom-right cell. The action space consists of four deterministic moves: up, down, left, and right. Attempting to move outside the grid leaves the agent in the same cell. The goal state is terminal and absorbing. Rewards are defined on deterministic transitions. Each non-terminal step incurs a constant step cost of $-0.01$. In addition, the agent receives a $1$ reward upon entering the goal cell. Once at the goal, all subsequent rewards are zero. We use a discount factor $\gamma=0.99$ and an initial state distribution concentrated on the start state.

For the robust training, we consider an $l_2$-bounded perturbation set over state-action rewards. Concretely, letting $r$ denote the original reward and $\delta$ as additive perturbation, we define:
\[
\mathcal{U}=\{r+\delta: \|\delta\|_2\le \varepsilon\} \subset \mathbb{R}^{\mathcal{X'}}.
\]
Then the reward-robust objective is the standard max-min problem:
\[
\max_{\pi} \min_{\tilde r \in \mathcal{U}} J(\pi, \tilde r)
\]

To optimize this objective, we use a standard game-style procedure.
At each outer iteration $i$, given the current policy $\pi_i$, we approximately
solve the inner minimization by projected gradient steps on the reward
perturbation $\delta$, descending the return $J(\pi_i, r + \delta)$ while
projecting back onto the feasible set
$\{\delta : \|\delta\|_2 \le \varepsilon\}$.
This yields an approximate worst-case reward $\tilde r_i \in \mathcal{U}$.
We then perform a policy-optimization step via solving the Bellman equations to increase $J(\pi, \tilde r_i)$, producing the
next policy $\pi_{i+1}$. Repeating these two steps implements a practical
saddle-point optimization of the robust objective.

As for implementing our shaping-aware method, recall that the shaping-aware uncertainty set is defined as the image of the original uncertainty set  under the PBRS canonicalization map:
\[
\mathcal{U}_{\text{core}}
:= \operatorname{Proj}(\mathcal{U})
= \bigl\{ \operatorname{Proj}(r + \delta) : \|\delta\|_2 \le \varepsilon \bigr\}.
\]
In our grid-world, transitions are deterministic, so the rewards can be represented as a state-action vector $r \in \mathbb{R}^{\mathcal{X}}$. We use the $l_2$ geometry in our uncertainty set, consequently, \(\operatorname{Proj}\) is an orthogonal projection that removes PBRS components.

Let \(\Phi \in \mathbb{R}^\mathcal{S}\) be a potential function and define the shaping operator \(B\in \mathbb{R}^{\mathcal{SA}\times \mathcal{S}}\) by
\[
(B\Phi)(s,a,s') \;:=\; \gamma \Phi(s') - \Phi(s),
\]
where $s'$ is the deterministic next state after taking action $a$ in state $s$.
The shaping subspace is
\(
\mathcal{S}_{\text{shape}} = \operatorname{range}(B).
\) To make the decomposition unique, we fix a gauge (e.g., $\Phi(s_{0}) = 0$ ). The projection of a reward $r$ is then
\[
\operatorname{Proj}(r) = r - B\Phi^*(r), 
\qquad
\Phi^*(r) \in \arg\min_{\Phi :\, \Phi(s_{0})=0}
\bigl\| r - B\Phi \bigr\|_2^2.
\]
Equivalently, $\operatorname{Proj}(r)$ is linear and can be written as $\operatorname{Proj}(r) = P r$, where $P\in \mathbb{R}^{\mathcal{SA}\times \mathcal{SA}}$ is the orthogonal projector onto the complement of $\operatorname{range}(B)$. We denote the nominal core reward by
\[
r_{\text{core}} := \operatorname{Proj}(r) = P r.
\]
A direct characterization of $\mathcal{U}_{\text{core}}$ is therefore:
\[
\mathcal{U}_{\text{core}}
= \bigl\{ r_{\text{core}} + P\delta : \|\delta\|_2 \le \varepsilon\bigr\},
\]

Thus, the original $\ell_2$-ball uncertainty set becomes a linear image of that
set under $P$: a convex polytope in the core reward space. Importantly, the
projection collapses all PBRS-equivalent perturbations, i.e., all perturbations
differing only by an element of $\mathrm{range}(B)$, to the same core reward.

In practice, rather than explicitly parameterizing $\delta$ and repeatedly
applying $P$, we optimize the inner problem directly in a core coordinate
system. Concretely, we precompute a matrix
$Q \in \mathbb{R}^{SA \times d}$ whose columns form an orthonormal basis for
the projected subspace (so that $P = QQ^\top$), and represent core rewards as
$r_{\text{core}} = Qz$ for $z \in \mathbb{R}^d$. The inner minimization is then
carried out over $\tilde z$ with an $\ell_2$ budget:
\[
\min_{\tilde z : \lVert \tilde z - z \rVert_2 \le \varepsilon}
    J(\pi, Q \tilde z),
\qquad
z := Q^\top r,
\]
using projected gradient descent in $\tilde z$-space. The outer maximization updates the policy
parameters using the resulting worst-case core reward $Qz$. This yields the
shaping-aware max-min solver, where every adversary update is
confined to the behaviorally meaningful directions by
construction.

For the numerical simulation, we use a softmax policy and set the uncertainty budget $\varepsilon=5.0$. Both methods are trained for 300 outer policy iterations, using a policy learning rate of 0.5. For the inner adversary problem, we run 20 update steps per outer iteration with an adversary learning rate of 0.5. To quantify how much the inner adversary optimization spends updating
PBRS-invariant directions, we measure, for each inner step, the
fraction of the update that lies in the shaping subspace
$\mathrm{range}(B)$. Define the
complementary projector onto shaping directions by
\[
    P_{\text{shape}} := I - P.
\]

At outer iteration $i$, let $r^{(i,j)}$ be the reward after the $j$-th inner
adversary step ($j = 0,\ldots,J$), and define the per-step update
\[
    \Delta r^{(i,j)} := r^{(i,j+1)} - r^{(i,j)}.
\]
We then compute the PBRS-invariant update fraction for that step as
\[
    \alpha^{(i,j)}
    := \frac{\bigl\|P_{\text{shape}} \Delta r^{(i,j)}\bigr\|_2}
            {\bigl\|\Delta r^{(i,j)}\bigr\|_2 + \epsilon},
\]
where $\epsilon > 0$ is a small constant for numerical stability. Finally,
we report the average fraction over the $J$ inner updates at iteration $i$:
\[
    \mathrm{Frac}(i)
    := \frac{1}{J} \sum_{j=0}^{J-1} \alpha^{(i,j)}.
\]

By construction, values $\mathrm{Frac}(i) \approx 1$ indicate that most
adversary updates move along PBRS-equivalent directions, whereas values near $0$ indicate that the updates
primarily change the core reward.

\subsection{Implementation Details for MuJoCo, Atari, and Gymnasium-Robotics}
\label{appendix:implementation-mujuco}

For the MuJoCo, Atari, and Gymnasium-Robotics experiments, we consider two reward-robust formulations
introduced in Section~\ref{sec:preliminary}. In both cases, we report
results under the corresponding reward-robust objective
\[
    \max_{\pi} \; \min_{\tilde r \in \mathcal{U}} J(\pi, \tilde r),
\]
and implement training via the equivalent regularized RL objective. We optimize this objective using PPO~\cite{schulman2017proximal}, where the regularization term is included as an intrinsic reward. Notice that in all environments considered in our experiments, the per-step reward depends on the current state and
action. We therefore write reward as $r(s,a)$ and omit the next state argument for simplicity.

\paragraph{Entropy-based robust objective.}
We first consider the entropy-induced uncertainty set
\[
    \mathcal{U}
    =
    \left\{\tilde r :
        \mathbb{E}\!\left[
            \sum_{t=0}^{\infty} \gamma^{t}
            \log \int_{\mathcal A}
            \exp\bigl(
                r(s_t,a)
                - \tilde r(s_t,a)
            \bigr)
            \,da\Big|\, \mu_0, \pi, P
        \right]
        \le \epsilon
    \right\},
\]
which is well known to be equivalent to optimizing an
entropy-regularized RL objective:
\[
    \max_{\pi}\;
    J(\pi, r)
    \;-\;
    \mathbb{E}\!\left[
        \sum_{t=0}^{\infty} \gamma^t
        \int_{\mathcal A}
            \pi(a \mid s_t)\,\log \pi(a \mid s_t)
        \,da\Bigg|\, \mu_0, \pi, P
    \right],
\]
following prior work~\cite{haarnoja2018soft,eysenbach2021maximum}.

For our shaping-aware variant, we add a projection step that learns the
shaping potential function $\Phi_{\varphi}$ and replaces the reward with its
PBRS-canonicalized version. For the entropy case, the projection loss is
\[
\varphi^{*}\in \arg\min_{\varphi} l(\varphi)=\arg\min_{\varphi}\;
\mathbb{E}_{\mathcal D}\Biggl[\sum_{t=0}^{\infty}\gamma^t\;
\log \int_{\mathcal A}
\exp\Bigl(
r(s_t,a)
-\bigl(\gamma \Phi_{\varphi}(s_{t+1})-\Phi_{\varphi}(s_t)\bigr)
\Bigr)\,da\Biggr].
\]
We estimate the objective using a batch dataset $\mathcal D$ of $N$ on-policy trajectories.
The inner $\log \int_{\mathcal A}\exp(\cdot)\,da$ term is an action-space log-partition function,
which we approximate by Monte Carlo with a proposal density $q(\cdot\mid s)$.
In our implementation we take $q(\cdot\mid s)=\pi(\cdot\mid s)$ (the current policy) and draw
$K$ i.i.d.\ actions $a^{(1)},\ldots,a^{(K)}\sim \pi(\cdot\mid s_t)$, yielding the importance-sampled estimator
\[
\log \int_{\mathcal A} \exp(u_t(a))\,da
\;\approx\;
\log\!\left(\frac{1}{K}\sum_{k=1}^K
\frac{\exp(u_t(a^{(k)}))}{\pi(a^{(k)}\mid s_t)}\right),
\]
where $u_t(a)=r(s_t,a)-(\gamma\Phi_\varphi(s_{t+1})-\Phi_\varphi(s_t))$.
Substituting this estimator into the projection objective yields the empirical minibatch loss
\[
\widehat{l}(\varphi;\mathcal D)
:=
\frac{1}{N}\sum_{i=1}^{N}\sum_{t=0}^{T_i-1}\gamma^{t}\;
\log\!\left(\frac{1}{K}\sum_{k=1}^{K}
\frac{\exp\!\bigl(u^{(i)}_t(a^{(i,k)})\bigr)}{\pi(a^{(i,k)}\mid s^{(i)}_t)}\right),
\qquad
\varphi^{*}\in\arg\min_{\varphi}\ \widehat{l}(\varphi;\mathcal D).
\]

\paragraph{KL-based robust objective.}
Next, we consider the KL-induced uncertainty set
\[
\mathcal U
= \Biggl\{
\tilde r:\;
\mathbb{E}\!\left[
\sum_{t=0}^{\infty} \tau\gamma^t
\log \int_{\mathcal A}
\pi_{\mathrm{ref}}(a \mid s_t)
\exp\!\left(
\frac{ r(s_t,a) - \tilde r(s_t,a)}{\tau}
\right)
\;da\middle|\;
\mu_0,\pi,P
\right] \le \varepsilon
\Biggr\},
\]
where $\tau>0$ is the regularization weight and $\pi_{\mathrm{ref}}$ is a reference policy.
This robust formulation corresponds to a KL-regularized RL objective (Table~3 in~\cite{derman2023robustness}):
\[
\max_{\pi}
\; J(\pi,r)
\;-\;
\tau \,
\mathbb{E}\!\left[
\sum_{t=0}^{\infty} \gamma^t
\mathrm{KL}\bigl(\pi(\cdot \mid s_t)\,\big\|\,\pi_{\mathrm{ref}}(\cdot \mid s_t)\bigr)
\;\middle|\;
\mu_0,\pi,P
\right].
\]

For our shaping-aware variant, the projection loss becomes
\[
\varphi^*
\in \arg\min_{\varphi} l(\varphi)=
\arg \min_{\varphi}
\mathbb{E}_{\mathcal D}\Biggl[\sum_{t=0}^{\infty} \gamma^t
 \log
\int_{\mathcal A}
\pi_{\mathrm{ref}}(a \mid s_t)
\exp\!\left(
\frac{
r(s_t,a) -
\bigl(\gamma \Phi_{\varphi}(s_{t+1}) - \Phi_{\varphi}(s_t)\bigr)
}{\tau}
\right)da
\Biggr].
\]
Similarly, we estimate the expectation using a batch dataset $\mathcal D$ of $N$ on-policy trajectories.
Define
\[
u_t(a)
:= \frac{ r(s_t,a) - \bigl(\gamma \Phi_{\varphi}(s_{t+1}) - \Phi_{\varphi}(s_t)\bigr)}{\tau}.
\]
We approximate the inner log term by
Monte Carlo using a proposal density $q(\cdot\mid s)$. In our implementation, we take $q(\cdot\mid s)=\pi_{\mathrm{ref}}(\cdot \mid s)$ to be the reference policy. Drawing $K$ i.i.d.\ actions
$a^{(1)},\ldots,a^{(K)}\sim \pi_{\mathrm{ref}}(\cdot\mid s_t)$ yields the importance-sampled estimator
\[
\log
\int_{\mathcal A}
\pi_{\mathrm{ref}}(a \mid s_t)\exp\!\bigl(u_t(a)\bigr)\,da
\;\approx\;
\log\!\left(\frac{1}{K}\sum_{k=1}^{K} \exp\!\bigl(u_t(a^{(k)})\bigr)\right).
\]
Substituting this estimator yields the empirical minibatch loss
\[
\widehat{l}(\varphi;\mathcal D)
:=
\frac{1}{N}\sum_{i=1}^{N}\sum_{t=0}^{T_i-1}\gamma^{t}\;
\log\!\left(\frac{1}{K}\sum_{k=1}^{K}
\exp\!\Bigl(
u^{(i)}_t(a^{(i,k)})
\Bigr)\right),
\qquad
\varphi^{*}\in\arg\min_{\varphi}\ \widehat{l}(\varphi;\mathcal D).
\]

Finally, to satisfy the technical conditions required by our convergence analysis
(Assumption~\ref{ass:strong-convexity}), we add a small ridge term to the shaping loss,
\[
\ell_{\lambda,\beta}(\varphi)
\;:=\;
\ell(\varphi)
\;+\;
\frac{\lambda}{2}\,\|\varphi\|^2
\;+\;
\beta\,
\mathbb{E}_{s_0 \sim \mathcal D}\!\bigl[\Phi_{\varphi}(s_0)^2\bigr],
\qquad
\lambda>0,\;\beta>0,
\]
where the $\beta$-term also encourages the gauge condition corresponding to
$C(\Phi)=0$ in our PBRS normalization.
We parameterize $\Phi_{\varphi}$ by a two-layer MLP with 256 hidden units per
layer and ReLU activations, and use the same architecture across all
environments. For the reference policy $\pi_{\text{ref}}$, we use a uniform policy to ensure the baseline is not informative so that all methods have equal room for improvement. In our implementation, we fix the step size and the number of iterations for projection update. Full hyperparameters for the projection step and PPO are reported
in Table~\ref{tab:hyperparameter_ppo} and
Table~\ref{tab:hyperparameter_projection}.

\begin{table}[ht]
    \centering
    \caption{Hyperparameters used for projection across different environments.}
    \label{tab:hyperparameter_projection}
    \small
    \setlength{\tabcolsep}{6pt}
    \renewcommand{\arraystretch}{1.2}
    \resizebox{\textwidth}{!}{%
    \begin{tabular}{lcccccccc}
        \toprule
        \textbf{Hyperparameter} & \textbf{Hopper} & \textbf{Walker} & \textbf{HalfCheetah} & \textbf{Humanoid} & \textbf{Reacher} & \textbf{Seaquest} & \textbf{Pong} & \textbf{Adroit Pen} \\
        \midrule
        Iteration number & 10 & 5 & 5 & 5 & 5 & 5 & 5 & 5\\
        Batch size & 256 & 256 & 256 & 256 & 256 & 256 & 256 & 256\\
        Sampled actions $K$ & 64& 64 & 64& 64 & 64 & 64 & 64& 128\\
        Optimizer & Adam & Adam & Adam & Adam & Adam& Adam & Adam & Adam\\
        Step size & $1\times10^{-3}$ & $1\times10^{-4}$ & $3\times10^{-4}$ & $3\times10^{-4}$ & $3\times10^{-4}$ & $1\times10^{-3}$ & $1\times10^{-3}$ & $1\times10^{-3}$\\
        Ridge term weight $\lambda$ & $1\times10^{-4}$ & $1\times10^{-4}$  & $1\times10^{-4}$ & $1\times10^{-4}$   & $1\times10^{-4}$ & $1\times10^{-4}$ & $1\times10^{-4}$   & $1\times10^{-4}$   \\
        Gauge term weight $\beta$ & $1\times10^{-3} $ & $1\times10^{-3} $  & $1\times10^{-3} $ & $1\times10^{-3} $   & $1\times10^{-3} $  & $1\times10^{-3} $ & $1\times10^{-3} $   & $1\times10^{-3} $   \\
        KL coefficient $\tau$ &0.2 & 0.1 & 0.2 &  0.1& 0.1 &0.2 &0.2 &0.2\\
        \bottomrule
    \end{tabular}}
\end{table}

\begin{table}[ht]
    \centering
    \caption{Hyperparameters used for PPO training for all methods across different environments.}
    \label{tab:hyperparameter_ppo}
    \small
    \setlength{\tabcolsep}{6pt}
    \renewcommand{\arraystretch}{1.2}
    \resizebox{\textwidth}{!}{%
    \begin{tabular}{lcccccccc}
        \toprule
        \textbf{Hyperparameter} & \textbf{Hopper} & \textbf{Walker} & \textbf{HalfCheetah} & \textbf{Humanoid} & \textbf{Reacher} & \textbf{Seaquest} & \textbf{Pong} & \textbf{Adroit Pen}\\
        \midrule
        Total timesteps & 500000 & 1000000 & 1000000 & 1000000  &500000  & 10000000& 1000000 & 2000000 \\
        Batch size & 64 & 64 & 64 & 64 & 64 & 64 & 64 & 64 \\
        Number steps & 2048 & 2048 & 2048 & 2048 & 2048 & 2048 & 2048 & 2048 \\
        Number envs &1 & 1 & 1 & 1 & 1 & 1 & 1 & 1\\
        Update epochs & 10 & 10 & 10 & 10 & 10 & 10 & 10 & 10 \\
        Optimizer & Adam & Adam & Adam & Adam & Adam & Adam & Adam & Adam\\
        Learning rate & $3\times10^{-4}$ & $3\times10^{-4}$ & $3\times10^{-4}$ & $3\times10^{-4}$ & $3\times10^{-4}$ & $2.5\times10^{-4}$& $2.5\times10^{-4}$ &$3\times10^{-4}$\\
        Discount factor ($\gamma$) & 0.99 & 0.99 & 0.99 & 0.99 & 0.99 & 0.99 & 0.99 & 0.99\\
        Entropy coefficient & 0.01 & 0.01 & 0.001 & 0.001 &  0.01 &  0.01 &  0.01 & 0.001\\
        KL coefficient & 0.2 & 0.1 & 0.2 &  0.1& 0.1&0.2 &0.2 &0.2 \\
        \bottomrule
    \end{tabular}}
\end{table}

\subsection{Implementation Details for Reward Perturbation}
\label{appendix:reward-perturb}

\begin{figure}[ht]
    \centering
    \includegraphics[width=0.65\textwidth]{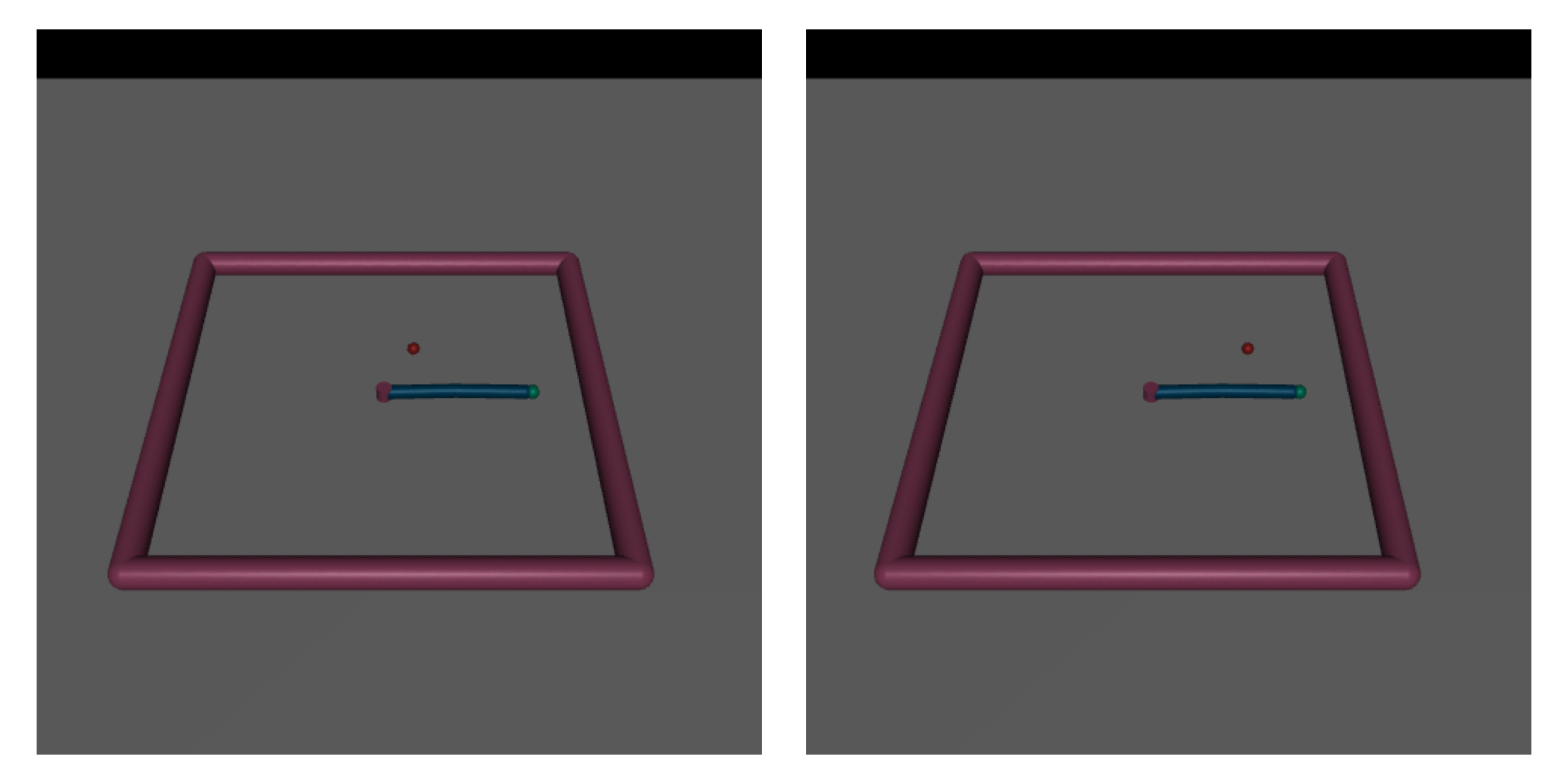}%
    \caption{Example goal shift in Reacher. Left: the original environment. Right: the perturbed environment with a shifted goal.}
    \label{fig:reacher-goal-shift}
\end{figure}

For the locomotion tasks (HalfCheetah, Hopper, Walker2d, and
Humanoid), during training, we construct the adversarial reward $r_{\text{adv}}$ using~\cite{husain2021regularized}:
\[
r_{\text{adv}} =
\begin{cases}
  r(s,a) + \delta, & \text{if } r(s,a) \le 0,\\
  r(s,a),          & \text{otherwise}.
\end{cases}
\]
where $\delta$ is drawn from a Gaussian distribution whose mean (the reward-shift radius) varies across conditions, while the variance is fixed to $0.1$. We found performance to be largely insensitive to this variance choice, so we keep it fixed and vary only the mean.

For Reacher, we relocate the goal to a random position at varying distances from its nominal location during evaluation~\cite{ashlag2025state}. Figure~\ref{fig:reacher-goal-shift} illustrates the original Reacher environment and an example perturbed setting with a shifted goal.

\subsection{Implementation Details for Ablation Study}
\label{appendix:implement-ablation}
To isolate the contribution of our $\Omega^*$-aligned canonicalization, we compare against several alternative projection choices. First, we include a random $\Phi$ variant, where the potential network $\Phi_\varphi$ is kept at its random initialization (i.e., $\varphi$ is not trained). 
Second, we include an $L^2$ projection baseline that ignores the geometry induced by the uncertainty set and instead defines $\Omega^*$ as a squared $\ell_2$ norm. 
Concretely, for both the entropy and KL settings, this corresponds to fitting $\Phi_\varphi$ by:
\[
\varphi^* \in \arg\min_{\varphi}\; l(\varphi)
\;:=\;
\mathbb{E}_{\mathcal D}\Biggl[\sum_{t=0}^{\infty}\gamma^t\;
\Bigl(
r(s_t,a_t,s_{t+1})
-\bigl(\gamma \Phi_{\varphi}(s_{t+1})-\Phi_{\varphi}(s_t)\bigr)
\Bigr)^2\Biggr].
\]
We estimate the objective using a batch dataset $\mathcal D$ of $N$ on-policy trajectories:
\[
\widehat l(\varphi;\mathcal D)
=
\frac{1}{N}\sum_{i=1}^{N}\sum_{t=0}^{T_i-1}\gamma^t
\Bigl(
r^{(i)}(s_t,a_t,s_{t+1})
-\bigl(\gamma \Phi_{\varphi}(s_{t+1}^{(i)})-\Phi_{\varphi}(s_t^{(i)})\bigr)
\Bigr)^2,
\]

\section{Additional Experiment Results}
\label{appendix:experiment-result}
\subsection{Analysis on Non-convex Uncertainty Set}
\label{appendix:non-convex}

\begin{figure}[ht]
    \centering
    \includegraphics[width=0.35\textwidth]{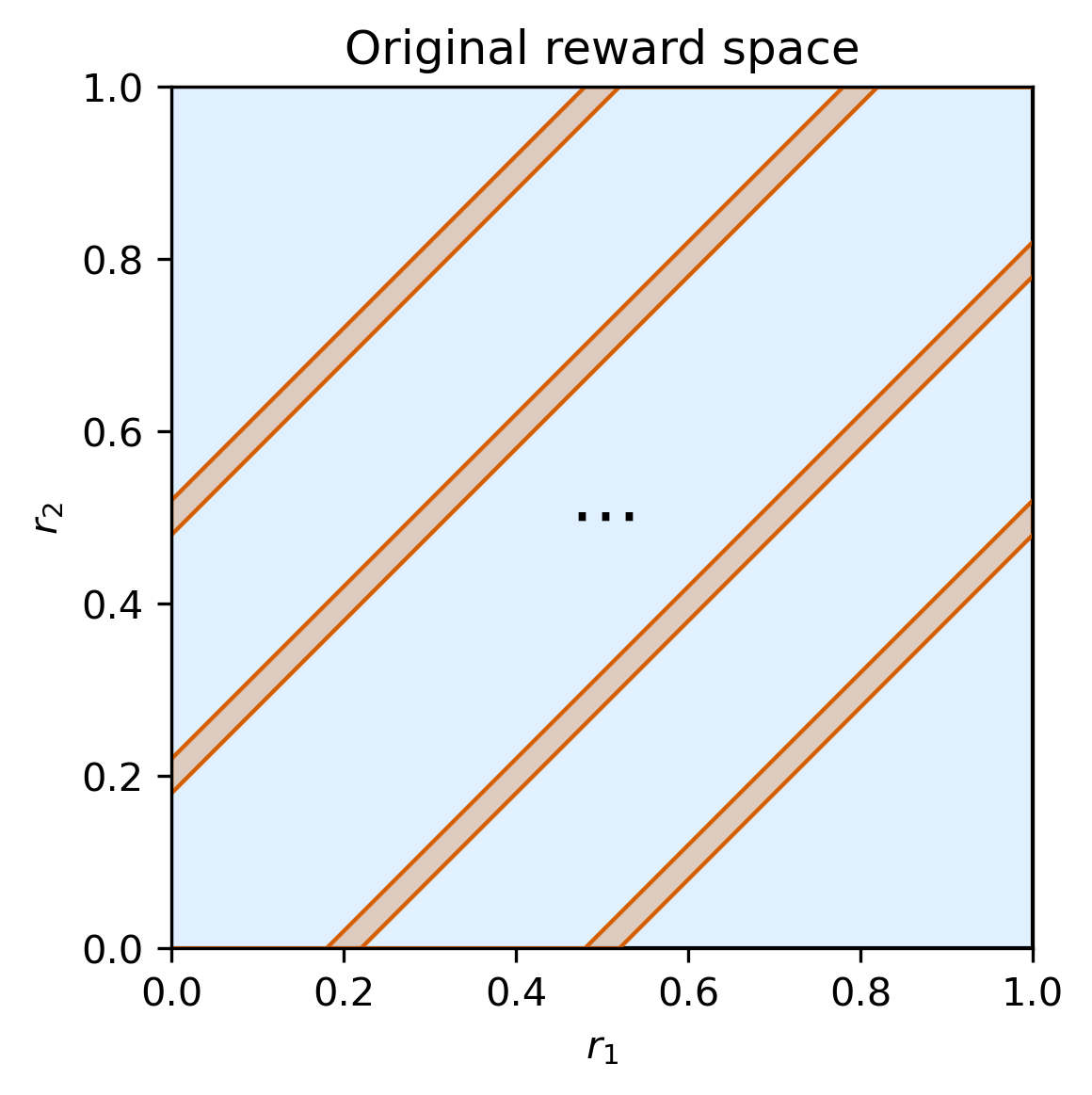}%
    \caption{Illustration of a nonconvex uncertainty set.}
    \label{fig:nonconvex-example}
\end{figure}

As discussed in the main text, when the adversary searches for a worst-case reward directly in the original reward space, it may spend many iterations moving along PBRS-invariant directions that do not change policy ranking. This issue is particularly acute for non-convex uncertainty sets (e.g., unions of scenarios) or implicitly specified feasible sets, where there may be no efficient dedicated solver for the optimization, and one must resort to black-box procedures (random search, Bayesian optimization, etc.)~\cite{choromanski2020provably,huttenrauch2024robust,kirschner2020distributionally,inatsu2022bayesian,inatsu2025bayesian}. Such methods typically suffer from sample complexity that scales badly with the dimension. Our shaping-aware projection reduces the intrinsic search dimension by removing PBRS-invariant directions, which can substantially improve the efficiency of black-box adversary search.

To illustrate, consider the same toy setting as in the main text in Figure~\ref{fig:toy-mdp-dim} and define a non-convex uncertainty set as the union of \(K\) thin diagonal bands:
\[
\mathcal{U} \;=\; \bigcup_{k=1}^K 
\left\{
  (r_1,r_2) \in [0,1]^2 \;:\;
  \bigl|(r_1 - r_2) - \delta_k\bigr| \le \eta
\right\}.
\]
When \(\eta\) is small, \(\mathcal{U}\) consists of \(K\) narrow strips aligned with the PBRS-invariant \((1,1)\) direction (Figure~\ref{fig:nonconvex-example}). The union of several scenarios makes the feasible set highly non-convex. Under the projection that fixes one coordinate (e.g., \(r_2^{\text{core}}=0\)), the set collapses to a one-dimensional union of intervals:
\[
\mathcal{U}_{\text{core}}
  \;=\; \bigcup_{k=1}^K [\delta_k - \eta,\; \delta_k + \eta]
  \;\subset\; [-1,1].
\]
Thus, projection removes the long PBRS-invariant direction, retaining only the behaviorally relevant coordinate.

Now suppose the adversary can only query feasible rewards. In the original two-dimensional set \(\mathcal{U}\), a generic black-box exploration strategy will nevertheless allocate evaluations across both directions in the ambient space. Since rewards are behaviorally identical along the invariant direction, many of these evaluations are redundant. In contrast, searching over \(\mathcal{U}_{\text{core}}\) allocates all evaluations along the identified coordinate, so each query is informative.

A simple scaling argument makes this difference explicit. Consider a single strip of width \(2\eta\). In \(\mathcal{U}_{\text{core}}\), choosing one reward that is within \(\varepsilon\) of the true worst-case reward along the interval requires on the order of
\[
N_{\mathcal{U}_{\text{core}}} \;\approx\; \frac{2\eta}{\varepsilon}
\]
feasible evaluations. In the original reward space, the corresponding strip has area approximately
\[
A \;\approx\; (2\eta)\,L,
\qquad
L \approx 1-|\delta_k|
\]
where \(L\) is the length of the diagonal segment inside \([0,1]^2\). For a two-dimensional space with \(N\) points, the typical spacing scales as \(h \approx \sqrt{A/N}\). To obtain comparable \(\varepsilon\)-level resolution, one needs
\[
N_{\mathcal{U}}
\;\approx\; \frac{A}{\varepsilon^2}
\;\approx\; \frac{2\eta L}{\varepsilon^2}.
\]
Therefore, black-box search scales as \(\mathcal{O}(1/\varepsilon)\) over \(\mathcal{U}_{\text{core}}\) but \(\mathcal{O}(1/\varepsilon^2)\) over \(\mathcal{U}\), with the extra factor arising from the algorithm implicitly spending samples to cover the PBRS-invariant direction.

For a numerical illustration, let \(\eta=0.01\), \(\delta_k=0.2\) (so \(L\approx 0.8\)), and \(\varepsilon=10^{-3}\). Then
\[
N_{\mathcal{U}_{\text{core}}} \approx \frac{0.02}{0.001}=20,
\qquad
N_{\mathcal{U}} \approx \frac{2\eta L}{\varepsilon^2}
= \frac{0.02\times 0.8}{10^{-6}}=16{,}000.
\]
In the original space, the overwhelming majority of evaluations are effectively spent spreading points along a behaviorally irrelevant direction. After projection, essentially every evaluation contributes to identifying a more adverse reward.

\subsection{Learning Curves on MuJoCo, Atari, and Gymnasium-Robotics Tasks.}
\label{appendix:learning_curve}

\begin{figure*}[ht]
    \centering
    \includegraphics[width=\linewidth]{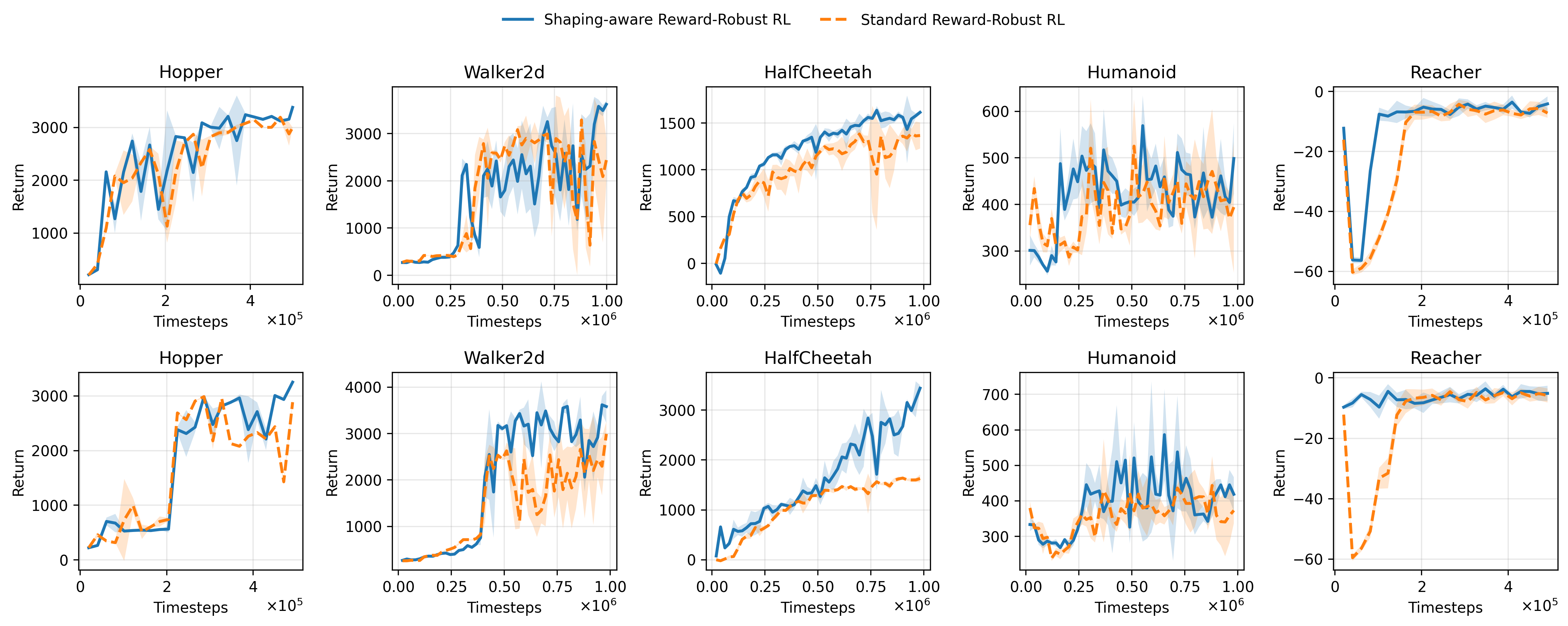}
    \caption{Mean and standard deviation return of the evaluation policy over 10 rollouts and 5 seeds, reported every 20k environment steps for MuJoCo environment. The return is in terms of the environment's original reward. Top row: Entropy-based reward-robust problem, bottom row: KL-based reward-robust problem.}
    \label{fig:learning_curve}
\end{figure*}

\begin{figure*}[ht]
    \centering
    \includegraphics[width=\linewidth]{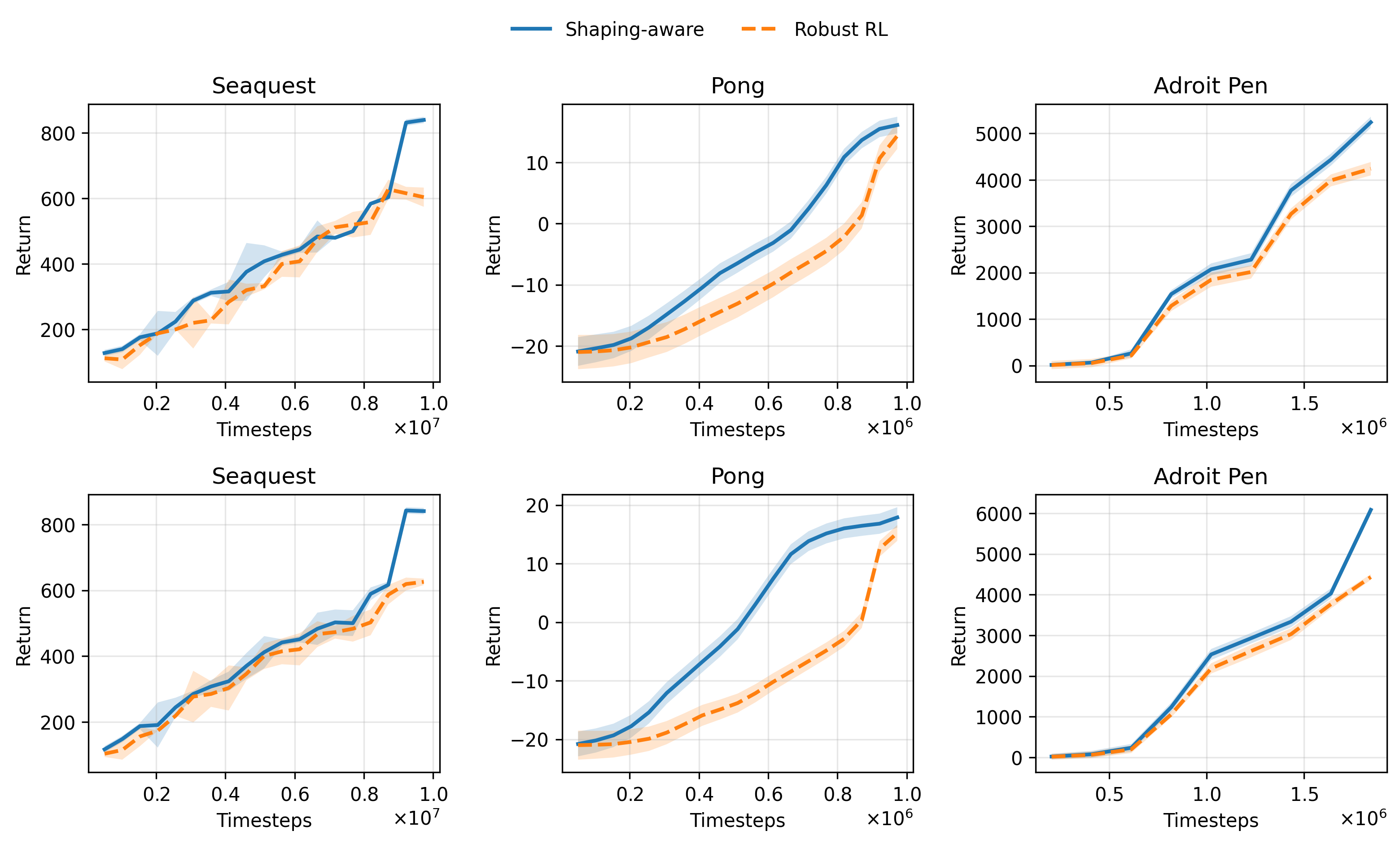}
    \caption{Mean and standard deviation return of the evaluation policy over 10 rollouts and 5 seeds, reported every 20k environment steps for Atari and Gymnasium-Robotics environment. The return is in terms of the environment's original reward. Top row: Entropy-based reward-robust problem, bottom row: KL-based reward-robust problem.}
    \label{fig:learning_curve_cont}
\end{figure*}

Figure~\ref{fig:learning_curve} and Figure\ref{fig:learning_curve_cont} compares learning curves for both methods. Overall, our method achieves higher returns throughout training and typically exhibits more stable optimization dynamics. The improvement is especially clear on HalfCheetah, Pong, and Adroit Pen, where the shaping-aware policy stays above the baseline at essentially every timestep and continues to improve late in training, while the baseline progresses more slowly and plateaus earlier. We observe a similar trend on Walker2d and Seaquest: although both methods reach comparable performance in the mid-training regime, the baseline becomes noticeably more volatile later on, whereas our method maintains a higher and more stable return trajectory. Beyond final performance, the curves also show a consistent sample-efficiency advantage. On Reacher, our method reaches near-solved performance almost immediately, while the baseline requires substantially more interaction (on the order of \(2\times 10^5\) timesteps) to catch up. On Hopper and Humanoid, the two methods share similar early-stage trends, but our method converges to a better asymptotic return and tends to reduce training instability as learning proceeds. Taken together, these results indicate that the shaping-aware projection not only improves final performance, but also enables the agent to reach a given performance level with fewer environment interactions, i.e., it converges faster than the standard reward-robust baseline.

\subsection{Reward Robustness Results}
\label{appendix:reward-robust}

\begin{figure*}[ht]
    \centering
    \includegraphics[width=\linewidth]{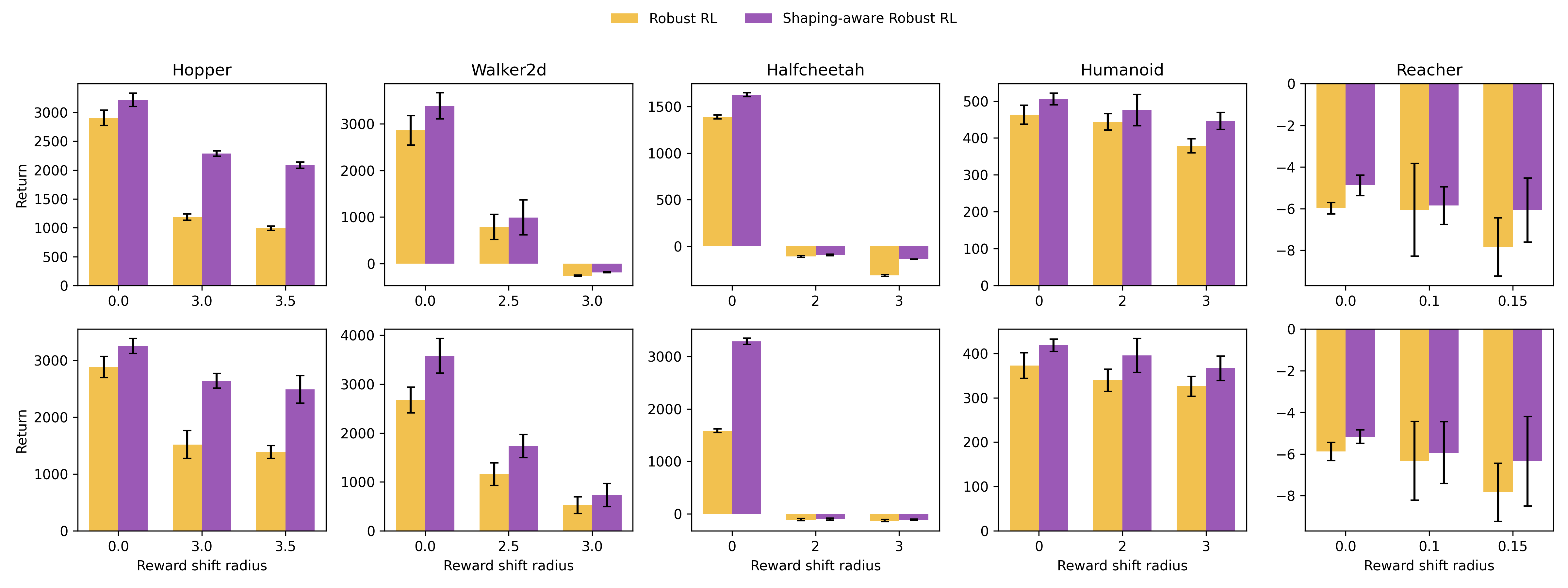}
    \caption{Mean $\pm$ 95\% confidence interval return of the evaluation policy under reward perturbation over 100 episodes and 5 seeds. Top row: Entropy-based reward-robust problem, bottom row: KL-based reward-robust problem.}
    \label{fig:reward_shift}
\end{figure*}

Figure~\ref{fig:reward_shift} shows the robustness under reward perturbations for both reward-robust problems. Overall, the shaping-aware method exhibits smaller performance degradation than the robust RL baseline as the reward shift radius increases, indicating improved stability to reward misspecification.

\subsection{Analysis on Training Time }

\begin{table}[ht]
    \centering
    \caption{Approximate training time for each algorithm across different environments.}
    \label{tab:time}
    \small
    \setlength{\tabcolsep}{8pt}
    \renewcommand{\arraystretch}{1.2}
    \begin{tabular}{lcc}
        \toprule
        \textbf{Environment} & \textbf{Standard Reward-Robust RL} & \textbf{Shaping-aware Reward-Robust RL}\\
        \midrule
        Hopper       & $\approx$50m    & $\approx$1h \\
        Walker       & $\approx$1h50m  & $\approx$2h10m \\
        HalfCheetah  & $\approx$1h30m  & $\approx$1h50m \\
        Humanoid     & $\approx$1h40m  & $\approx$2h \\
        Reacher      & $\approx$50m    & $\approx$1h \\
        Seaquest     & $\approx$5h20m  & $\approx$5h50m  \\
        Pong         & $\approx$1h            & $\approx$1h10m  \\
        Adroit Pen   & $\approx$3h50m           & $\approx$ 4h20m  \\
        \bottomrule
    \end{tabular}
\end{table}

Table~\ref{tab:time} reports the approximate training time for the standard reward-robust baseline and our shaping-aware variant. Across all tasks, the shaping-aware method incurs a small additional cost (typically around 10–30 minutes), which is attributable to the extra projection step that fits the potential function. Importantly, this overhead is modest relative to the overall training time (on the order of 
around 10–25\% depending on the environment) and remains practical in the same computational setting. Given the consistent gains in final return and faster convergence observed in Section~\ref{sec:experiment}, the added computation is a affordable trade-off for improved robustness and performance.

\subsection{Additional Ablation Study}
\label{appendix:ablation}

\begin{figure}[ht]
    \centering
    \includegraphics[width=\textwidth]{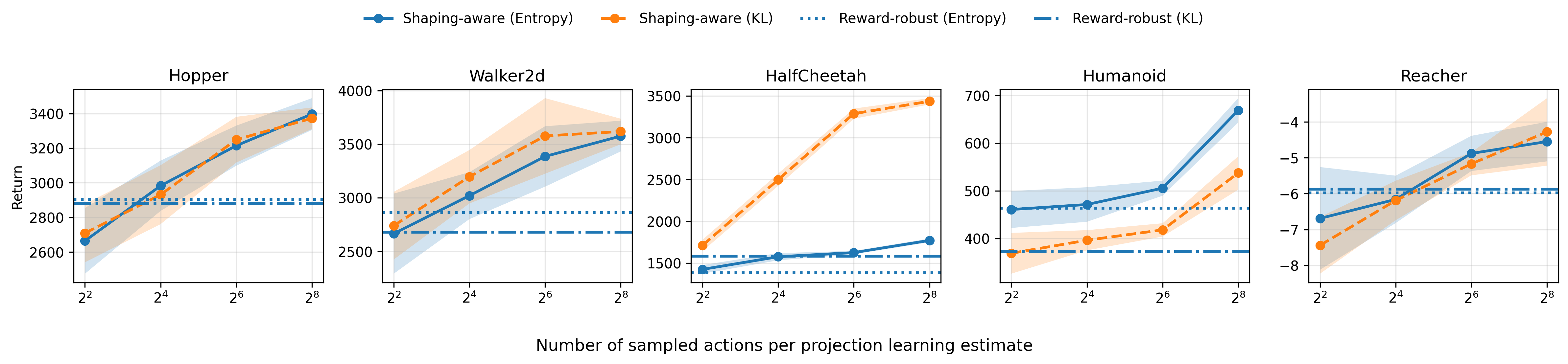}%
    \caption{Ablations on sampled actions for projection learning in MuJoCo environment. We evaluate each policy for 100 episodes and show 95\% confidence interval across 5 seeds.}
    \label{fig:action_ablation}
\end{figure}

\paragraph{Sensitivity to data coverage in the projection estimate.}
As detailed in Appendix~\ref{appendix:implementation}, the \(\Omega^*\) term used in our experiments involves a log-sum-exp over actions. Since the action space is continuous, we approximate this quantity via Monte Carlo by sampling $K$ actions (from the policy or the reference policy, depending on the objective). This approximation quality directly affects the projection step. When $K$ is small, the estimated projection loss can be noisy or biased, leading to an inaccurate canonicalization and, in turn, degraded downstream learning. To quantify this effect, we vary $K$ across all MuJoCo environments while keeping the rest of the training pipeline fixed, and evaluate the resulting performance. Figure~\ref{fig:action_ablation} shows a clear dependence on action-sample coverage. With a very small $K$, our method can underperform the standard reward-robust baselines, consistent with an unreliable estimate of the log-sum-exp term. This is particularly pronounced in Hopper, where a handful of action samples (e.g., $K=4$) provides a poor approximation. As $K$ increases, the projection estimate becomes more accurate and performance improves accordingly, eventually recovering and typically exceeding the baseline. Overall, this ablation highlights a practical tradeoff: the gains from shaping-aware projection are realized when the projection loss is estimated with sufficient action coverage, at the cost of additional sampling in the projection-learning step.

\begin{table}[t]
\centering
\caption{Ablation on canonicalization choices under reward perturbations. For Hopper, Walker2d, HalfCheetah, and Humanoid, we use a reward-shift radius of 
3. For Reacher, we use a goal-shift radius of 0.1. Results are reported as mean $\pm$
95\% confidence interval over 5 random seeds.}
\label{tab:ablation_canon}
\small
\setlength{\tabcolsep}{4.5pt}
\renewcommand{\arraystretch}{1.15}
\resizebox{\textwidth}{!}{%
\begin{tabular}{lccccc}
\toprule
\textbf{Method} & \textbf{Hopper} & \textbf{Walker2d} & \textbf{HalfCheetah} & \textbf{Humanoid} & \textbf{Reacher}\\
\midrule
\multicolumn{6}{l}{\textbf{Entropy-based robust objective}}\\
\midrule
Reward-Robust (baseline)                      & 1186.66$\pm$52.27 & -259.59$\pm$16.07 & -313.50$\pm$9.73 & 378.97$\pm$18.90 &  -6.05$\pm$2.22\\
Shaping-aware (ours, $\Omega^*$-projection)   & 2287.66$\pm$44.64 & -188.14$\pm$9.26 & -138.06$\pm$1.57 & 446.41$\pm$23.16 & -5.85$\pm$0.90 \\
Shaping-aware (random $\Phi$ canonicalization)& 1252.33$\pm$43.81 & -226.17$\pm$36.72 &  -361.70$\pm$9.97 & 338.29$\pm$22.26 & -6.35$\pm$2.14 \\
Shaping-aware ($L^2$ projection) & 1491.77$\pm$37.04 & -199.04$\pm$9.37 & -144.56$\pm$6.25 & 474.09$\pm$40.15 & -5.94$\pm$1.48 \\
\midrule
\multicolumn{6}{l}{\textbf{KL-based robust objective}}\\
\midrule
Reward-Robust (baseline)                      & 1519.33$\pm$241.86  & 525.17$\pm$171.17 & -128.73$\pm$21.13 & 325.81$\pm$22.22 & -6.33$\pm$1.89 \\
Shaping-aware (ours, $\Omega^*$-projection)   & 2638.49$\pm$129.72 & 733.78$\pm$238.30 & -110.16$\pm$10.06 & 366.31$\pm$27.34 & -5.94$\pm$1.48 \\
Shaping-aware (random $\Phi$ canonicalization)& 1230.95$\pm$124.39 & 446.50$\pm$170.87 & -125.51$\pm$21.98 & 311.47$\pm$15.03 & -6.98$\pm$1.92 \\
Shaping-aware ($L^2$ projection)& 2016.28$\pm$113.72 & 693.48$\pm$109.97 & -111.75$\pm$13.99 & 335.10$\pm$14.12  & -6.17$\pm$1.41  \\
\bottomrule
\end{tabular}
}
\end{table}

\paragraph{Ablation on canonicalization choices}
To isolate the contribution of our \(\Omega^*\)-aligned canonicalization, Table~\ref{tab:ablation_canon} compares several projection choices under reward perturbations. First, we include a random $\Phi$ variant, which injects an arbitrary PBRS shaping term into the reward. This variant does not yield meaningful robustness improvements, indicating that the gains of our method do not come from simply adding an extra shaping-like component, but from learning a canonicalization that removes PBRS redundancy in a principled way. We also consider an $L^2$ projection that performs orthogonal projection regardless of the uncertainty-set geometry. This variant still removes PBRS-equivalent directions and thus improves robustness relative to the baseline, confirming that redundancy removal alone is beneficial. However, the $\Omega^*$-aligned projection performs better in most cases, suggesting that matching the canonicalization to the robustness geometry provides additional gains under reward perturbations beyond what a generic $L^2$ projection achieves.


\newpage
\section*{NeurIPS Paper Checklist}

\begin{enumerate}

\item {\bf Claims}
    \item[] Question: Do the main claims made in the abstract and introduction accurately reflect the paper's contributions and scope?
    \item[] Answer: \answerYes{}
    \item[] Justification: The abstract and introduction state the core contributions, which is shaping-aware reward-robust RL framework.
    \item[] Guidelines:
    \begin{itemize}
        \item The answer \answerNA{} means that the abstract and introduction do not include the claims made in the paper.
        \item The abstract and/or introduction should clearly state the claims made, including the contributions made in the paper and important assumptions and limitations. A \answerNo{} or \answerNA{} answer to this question will not be perceived well by the reviewers.
        \item The claims made should match theoretical and experimental results, and reflect how much the results can be expected to generalize to other settings.
        \item It is fine to include aspirational goals as motivation as long as it is clear that these goals are not attained by the paper.
    \end{itemize}

\item {\bf Limitations}
    \item[] Question: Does the paper discuss the limitations of the work performed by the authors?
    \item[] Answer: \answerYes{}
    \item[] Justification: Limitations of the work is discussed in Appendix~\ref{appendix:limitations}.
    \item[] Guidelines:
    \begin{itemize}
        \item The answer \answerNA{} means that the paper has no limitation while the answer \answerNo{} means that the paper has limitations, but those are not discussed in the paper.
        \item The authors are encouraged to create a separate ``Limitations'' section in their paper.
        \item The paper should point out any strong assumptions and how robust the results are to violations of these assumptions (e.g., independence assumptions, noiseless settings, model well-specification, asymptotic approximations only holding locally). The authors should reflect on how these assumptions might be violated in practice and what the implications would be.
        \item The authors should reflect on the scope of the claims made, e.g., if the approach was only tested on a few datasets or with a few runs. In general, empirical results often depend on implicit assumptions, which should be articulated.
        \item The authors should reflect on the factors that influence the performance of the approach. For example, a facial recognition algorithm may perform poorly when image resolution is low or images are taken in low lighting. Or a speech-to-text system might not be used reliably to provide closed captions for online lectures because it fails to handle technical jargon.
        \item The authors should discuss the computational efficiency of the proposed algorithms and how they scale with dataset size.
        \item If applicable, the authors should discuss possible limitations of their approach to address problems of privacy and fairness.
        \item While the authors might fear that complete honesty about limitations might be used by reviewers as grounds for rejection, a worse outcome might be that reviewers discover limitations that aren't acknowledged in the paper. The authors should use their best judgment and recognize that individual actions in favor of transparency play an important role in developing norms that preserve the integrity of the community. Reviewers will be specifically instructed to not penalize honesty concerning limitations.
    \end{itemize}

\item {\bf Theory assumptions and proofs}
    \item[] Question: For each theoretical result, does the paper provide the full set of assumptions and a complete (and correct) proof?
    \item[] Answer: \answerYes{}
    \item[] Justification: All the assumptions and proof can be found in the Appendix. 
    \item[] Guidelines:
    \begin{itemize}
        \item The answer \answerNA{} means that the paper does not include theoretical results.
        \item All the theorems, formulas, and proofs in the paper should be numbered and cross-referenced.
        \item All assumptions should be clearly stated or referenced in the statement of any theorems.
        \item The proofs can either appear in the main paper or the supplemental material, but if they appear in the supplemental material, the authors are encouraged to provide a short proof sketch to provide intuition.
        \item Inversely, any informal proof provided in the core of the paper should be complemented by formal proofs provided in appendix or supplemental material.
        \item Theorems and Lemmas that the proof relies upon should be properly referenced.
    \end{itemize}

    \item {\bf Experimental result reproducibility}
    \item[] Question: Does the paper fully disclose all the information needed to reproduce the main experimental results of the paper to the extent that it affects the main claims and/or conclusions of the paper (regardless of whether the code and data are provided or not)?
    \item[] Answer: \answerYes{}
    \item[] Justification: All training hyperparameters are specified in Appendix~\ref{appendix:implementation}.
    \item[] Guidelines:
    \begin{itemize}
        \item The answer \answerNA{} means that the paper does not include experiments.
        \item If the paper includes experiments, a \answerNo{} answer to this question will not be perceived well by the reviewers: Making the paper reproducible is important, regardless of whether the code and data are provided or not.
        \item If the contribution is a dataset and\slash or model, the authors should describe the steps taken to make their results reproducible or verifiable.
        \item Depending on the contribution, reproducibility can be accomplished in various ways. For example, if the contribution is a novel architecture, describing the architecture fully might suffice, or if the contribution is a specific model and empirical evaluation, it may be necessary to either make it possible for others to replicate the model with the same dataset, or provide access to the model. In general. releasing code and data is often one good way to accomplish this, but reproducibility can also be provided via detailed instructions for how to replicate the results, access to a hosted model (e.g., in the case of a large language model), releasing of a model checkpoint, or other means that are appropriate to the research performed.
        \item While NeurIPS does not require releasing code, the conference does require all submissions to provide some reasonable avenue for reproducibility, which may depend on the nature of the contribution. For example
        \begin{enumerate}
            \item If the contribution is primarily a new algorithm, the paper should make it clear how to reproduce that algorithm.
            \item If the contribution is primarily a new model architecture, the paper should describe the architecture clearly and fully.
            \item If the contribution is a new model (e.g., a large language model), then there should either be a way to access this model for reproducing the results or a way to reproduce the model (e.g., with an open-source dataset or instructions for how to construct the dataset).
            \item We recognize that reproducibility may be tricky in some cases, in which case authors are welcome to describe the particular way they provide for reproducibility. In the case of closed-source models, it may be that access to the model is limited in some way (e.g., to registered users), but it should be possible for other researchers to have some path to reproducing or verifying the results.
        \end{enumerate}
    \end{itemize}

\item {\bf Open access to data and code}
    \item[] Question: Does the paper provide open access to the data and code, with sufficient instructions to faithfully reproduce the main experimental results, as described in supplemental material?
    \item[] Answer: \answerYes{}
    \item[] Justification: The code is included in the supplementary. All data is open accessible. 
    \item[] Guidelines:
    \begin{itemize}
        \item The answer \answerNA{} means that paper does not include experiments requiring code.
        \item Please see the NeurIPS code and data submission guidelines (\url{https://neurips.cc/public/guides/CodeSubmissionPolicy}) for more details.
        \item While we encourage the release of code and data, we understand that this might not be possible, so \answerNo{} is an acceptable answer. Papers cannot be rejected simply for not including code, unless this is central to the contribution (e.g., for a new open-source benchmark).
        \item The instructions should contain the exact command and environment needed to run to reproduce the results. See the NeurIPS code and data submission guidelines (\url{https://neurips.cc/public/guides/CodeSubmissionPolicy}) for more details.
        \item The authors should provide instructions on data access and preparation, including how to access the raw data, preprocessed data, intermediate data, and generated data, etc.
        \item The authors should provide scripts to reproduce all experimental results for the new proposed method and baselines. If only a subset of experiments are reproducible, they should state which ones are omitted from the script and why.
        \item At submission time, to preserve anonymity, the authors should release anonymized versions (if applicable).
        \item Providing as much information as possible in supplemental material (appended to the paper) is recommended, but including URLs to data and code is permitted.
    \end{itemize}

\item {\bf Experimental setting/details}
    \item[] Question: Does the paper specify all the training and test details (e.g., data splits, hyperparameters, how they were chosen, type of optimizer) necessary to understand the results?
    \item[] Answer: \answerYes{}
    \item[] Justification: All details can be found in Appendix~\ref{appendix:implementation}.
    \item[] Guidelines:
    \begin{itemize}
        \item The answer \answerNA{} means that the paper does not include experiments.
        \item The experimental setting should be presented in the core of the paper to a level of detail that is necessary to appreciate the results and make sense of them.
        \item The full details can be provided either with the code, in appendix, or as supplemental material.
    \end{itemize}

\item {\bf Experiment statistical significance}
    \item[] Question: Does the paper report error bars suitably and correctly defined or other appropriate information about the statistical significance of the experiments?
    \item[] Answer: \answerYes{}
    \item[] Justification: We report mean $\pm$ 95\% confidence interval over 5 random seeds.
    \item[] Guidelines:
    \begin{itemize}
        \item The answer \answerNA{} means that the paper does not include experiments.
        \item The authors should answer \answerYes{} if the results are accompanied by error bars, confidence intervals, or statistical significance tests, at least for the experiments that support the main claims of the paper.
        \item The factors of variability that the error bars are capturing should be clearly stated (for example, train/test split, initialization, random drawing of some parameter, or overall run with given experimental conditions).
        \item The method for calculating the error bars should be explained (closed form formula, call to a library function, bootstrap, etc.)
        \item The assumptions made should be given (e.g., Normally distributed errors).
        \item It should be clear whether the error bar is the standard deviation or the standard error of the mean.
        \item It is OK to report 1-sigma error bars, but one should state it. The authors should preferably report a 2-sigma error bar than state that they have a 96\% CI, if the hypothesis of Normality of errors is not verified.
        \item For asymmetric distributions, the authors should be careful not to show in tables or figures symmetric error bars that would yield results that are out of range (e.g., negative error rates).
        \item If error bars are reported in tables or plots, the authors should explain in the text how they were calculated and reference the corresponding figures or tables in the text.
    \end{itemize}

\item {\bf Experiments compute resources}
    \item[] Question: For each experiment, does the paper provide sufficient information on the computer resources (type of compute workers, memory, time of execution) needed to reproduce the experiments?
    \item[] Answer: \answerYes{}
    \item[] Justification: All details can be found in Appendix~\ref{appendix:implementation}.
    \item[] Guidelines:
    \begin{itemize}
        \item The answer \answerNA{} means that the paper does not include experiments.
        \item The paper should indicate the type of compute workers CPU or GPU, internal cluster, or cloud provider, including relevant memory and storage.
        \item The paper should provide the amount of compute required for each of the individual experimental runs as well as estimate the total compute.
        \item The paper should disclose whether the full research project required more compute than the experiments reported in the paper (e.g., preliminary or failed experiments that didn't make it into the paper).
    \end{itemize}

\item {\bf Code of ethics}
    \item[] Question: Does the research conducted in the paper conform, in every respect, with the NeurIPS Code of Ethics \url{https://neurips.cc/public/EthicsGuidelines}?
    \item[] Answer: \answerYes{}
    \item[] Justification: The research conforms with the NeurIPS Code of Ethics.
    \item[] Guidelines:
    \begin{itemize}
        \item The answer \answerNA{} means that the authors have not reviewed the NeurIPS Code of Ethics.
        \item If the authors answer \answerNo, they should explain the special circumstances that require a deviation from the Code of Ethics.
        \item The authors should make sure to preserve anonymity (e.g., if there is a special consideration due to laws or regulations in their jurisdiction).
    \end{itemize}

\item {\bf Broader impacts}
    \item[] Question: Does the paper discuss both potential positive societal impacts and negative societal impacts of the work performed?
    \item[] Answer: \answerNA{}
    \item[] Justification: This paper presents work whose goal is to advance the field
of Machine Learning. There are many potential societal
consequences of our work, none which we feel must be
specifically highlighted.
    \item[] Guidelines:
    \begin{itemize}
        \item The answer \answerNA{} means that there is no societal impact of the work performed.
        \item If the authors answer \answerNA{} or \answerNo, they should explain why their work has no societal impact or why the paper does not address societal impact.
        \item Examples of negative societal impacts include potential malicious or unintended uses (e.g., disinformation, generating fake profiles, surveillance), fairness considerations (e.g., deployment of technologies that could make decisions that unfairly impact specific groups), privacy considerations, and security considerations.
        \item The conference expects that many papers will be foundational research and not tied to particular applications, let alone deployments. However, if there is a direct path to any negative applications, the authors should point it out. For example, it is legitimate to point out that an improvement in the quality of generative models could be used to generate Deepfakes for disinformation. On the other hand, it is not needed to point out that a generic algorithm for optimizing neural networks could enable people to train models that generate Deepfakes faster.
        \item The authors should consider possible harms that could arise when the technology is being used as intended and functioning correctly, harms that could arise when the technology is being used as intended but gives incorrect results, and harms following from (intentional or unintentional) misuse of the technology.
        \item If there are negative societal impacts, the authors could also discuss possible mitigation strategies (e.g., gated release of models, providing defenses in addition to attacks, mechanisms for monitoring misuse, mechanisms to monitor how a system learns from feedback over time, improving the efficiency and accessibility of ML).
    \end{itemize}

\item {\bf Safeguards}
    \item[] Question: Does the paper describe safeguards that have been put in place for responsible release of data or models that have a high risk for misuse (e.g., pre-trained language models, image generators, or scraped datasets)?
    \item[] Answer: \answerNA{}
    \item[] Justification: The paper poses no such risks.
    \item[] Guidelines:
    \begin{itemize}
        \item The answer \answerNA{} means that the paper poses no such risks.
        \item Released models that have a high risk for misuse or dual-use should be released with necessary safeguards to allow for controlled use of the model, for example by requiring that users adhere to usage guidelines or restrictions to access the model or implementing safety filters.
        \item Datasets that have been scraped from the Internet could pose safety risks. The authors should describe how they avoided releasing unsafe images.
        \item We recognize that providing effective safeguards is challenging, and many papers do not require this, but we encourage authors to take this into account and make a best faith effort.
    \end{itemize}

\item {\bf Licenses for existing assets}
    \item[] Question: Are the creators or original owners of assets (e.g., code, data, models), used in the paper, properly credited and are the license and terms of use explicitly mentioned and properly respected?
    \item[] Answer: \answerYes{}
    \item[] Justification: All assets have been cited.
    \item[] Guidelines:
    \begin{itemize}
        \item The answer \answerNA{} means that the paper does not use existing assets.
        \item The authors should cite the original paper that produced the code package or dataset.
        \item The authors should state which version of the asset is used and, if possible, include a URL.
        \item The name of the license (e.g., CC-BY 4.0) should be included for each asset.
        \item For scraped data from a particular source (e.g., website), the copyright and terms of service of that source should be provided.
        \item If assets are released, the license, copyright information, and terms of use in the package should be provided. For popular datasets, \url{paperswithcode.com/datasets} has curated licenses for some datasets. Their licensing guide can help determine the license of a dataset.
        \item For existing datasets that are re-packaged, both the original license and the license of the derived asset (if it has changed) should be provided.
        \item If this information is not available online, the authors are encouraged to reach out to the asset's creators.
    \end{itemize}

\item {\bf New assets}
    \item[] Question: Are new assets introduced in the paper well documented and is the documentation provided alongside the assets?
    \item[] Answer: \answerNA{}
    \item[] Justification: The paper does not release new assets.
    \item[] Guidelines:
    \begin{itemize}
        \item The answer \answerNA{} means that the paper does not release new assets.
        \item Researchers should communicate the details of the dataset\slash code\slash model as part of their submissions via structured templates. This includes details about training, license, limitations, etc.
        \item The paper should discuss whether and how consent was obtained from people whose asset is used.
        \item At submission time, remember to anonymize your assets (if applicable). You can either create an anonymized URL or include an anonymized zip file.
    \end{itemize}

\item {\bf Crowdsourcing and research with human subjects}
    \item[] Question: For crowdsourcing experiments and research with human subjects, does the paper include the full text of instructions given to participants and screenshots, if applicable, as well as details about compensation (if any)?
    \item[] Answer: \answerNA{}
    \item[] Justification: This work does not involve crowdsourcing or research with human subjects.
    \item[] Guidelines:
    \begin{itemize}
        \item The answer \answerNA{} means that the paper does not involve crowdsourcing nor research with human subjects.
        \item Including this information in the supplemental material is fine, but if the main contribution of the paper involves human subjects, then as much detail as possible should be included in the main paper.
        \item According to the NeurIPS Code of Ethics, workers involved in data collection, curation, or other labor should be paid at least the minimum wage in the country of the data collector.
    \end{itemize}

\item {\bf Institutional review board (IRB) approvals or equivalent for research with human subjects}
    \item[] Question: Does the paper describe potential risks incurred by study participants, whether such risks were disclosed to the subjects, and whether Institutional Review Board (IRB) approvals (or an equivalent approval/review based on the requirements of your country or institution) were obtained?
    \item[] Answer: \answerNA{}
    \item[] Justification: This work does not involve human subjects research.
    \item[] Guidelines:
    \begin{itemize}
        \item The answer \answerNA{} means that the paper does not involve crowdsourcing nor research with human subjects.
        \item Depending on the country in which research is conducted, IRB approval (or equivalent) may be required for any human subjects research. If you obtained IRB approval, you should clearly state this in the paper.
        \item We recognize that the procedures for this may vary significantly between institutions and locations, and we expect authors to adhere to the NeurIPS Code of Ethics and the guidelines for their institution.
        \item For initial submissions, do not include any information that would break anonymity (if applicable), such as the institution conducting the review.
    \end{itemize}

\item {\bf Declaration of LLM usage}
    \item[] Question: Does the paper describe the usage of LLMs if it is an important, original, or non-standard component of the core methods in this research? Note that if the LLM is used only for writing, editing, or formatting purposes and does \emph{not} impact the core methodology, scientific rigor, or originality of the research, declaration is not required.
    \item[] Answer: \answerNA{}
    \item[] Justification: The core method development in this research does not involve LLMs as any important, original, or non-standard components.
    \item[] Guidelines:
    \begin{itemize}
        \item The answer \answerNA{} means that the core method development in this research does not involve LLMs as any important, original, or non-standard components.
        \item Please refer to our LLM policy in the NeurIPS handbook for what should or should not be described.
    \end{itemize}

\end{enumerate}

\end{document}